\documentclass{article} 
\usepackage{iclr2016_conference,times}
\usepackage{amsmath,amssymb}
\usepackage[pdftex]{graphicx}
\usepackage{hyperref}
\usepackage{url}
\usepackage{color}
\usepackage{multirow}
\usepackage{xspace}
\usepackage{subfig}

\makeatletter
\DeclareRobustCommand\onedot{\futurelet\@let@token\@onedot}
\def\@onedot{\ifx\@let@token.\else.\null\fi\xspace}
\makeatother
\def\stateoftheart{state-of-the-art}

\def\groundtruth{ground truth}

\def\eg{\emph{e.g}\onedot}

 \def\vs{\emph{vs}\onedot}
 
\def\etal{\emph{et al}\onedot}

\newcommand{\refsec}[1]{Section\onedot~\ref{#1}}
\newcommand{\reffig}[1]{Figure\onedot~\ref{#1}}
\newcommand{\refeq}[1]{Equation\onedot~\ref{#1}}
\newcommand{\reftab}[1]{Table\onedot~\ref{#1}}

\newcommand{\secref}[1]{Sec\onedot~\ref{#1}}

\newcommand{\wf}{\ww}

\newcommand{\trset}{\mathcal{X}}
\newcommand{\param}{\mathbf{w}}
\newcommand{\cnnparams}{\mathbf{w}^{\mathrm{C}}}
\newcommand{\cnn}{\mathrm{CNN}}
\newcommand{\rbmparams}{\mathbf{w}^{\mathrm{R}}}
\newcommand{\rbm}{\mathrm{RBM}}
\newcommand{\bi}{\begin{itemize}}
\newcommand{\ei}{\end{itemize}}
\newcommand{\ba}{\begin{eqnarray}}
\newcommand{\ea}{\end{eqnarray}}

\newcommand{\yy}{\mathbf{y}}
\newcommand{\zz}{\mathbf{z}}
\newcommand{\hh}{\mathbf{h}}
\newcommand{\vv}{\mathbf{v}}
\newcommand{\ww}{\mathbf{W}}
\newcommand{\params}{W}

\newcommand{\bg}{\begin{gather}}
\renewcommand{\eg}{\end{gather}}

\newcommand{\mycomment}[1]{{}}

\renewcommand{\vec}[1]{\mathbf{#1}}

\def\argmin{\operatornamewithlimits{arg\,min}}

\graphicspath{{./figures/}{./figures/teaser/}{./figures/horse/}{./figures/cow/}{./figures/car/}{./figures/faces/}{./figures/cub200/}{./figures/pedestrian/}{./figures/pedestrian/cnn-results/}{./figures/pedestrian/Color/}{./figures/pedestrian/Fowlkes/}{./figures/pedestrian/GroundTruth/}{./figures/rbms/}{./figures/multi_scale}}
\DeclareGraphicsExtensions{.pdf,.jpeg,.png,.jpg}

\title{Deep Learning for Semantic Part Segmentation with High-Level Guidance}

\author{Stavros Tsogkas \& Iasonas Kokkinos\\
Center for Visual Computing\\
Centrale-Supelec\\
\texttt{\{stavros.tsogkas,iasonas.kokkinos\}@centralesupelec.fr} \\
\And
George Papandreou\\
Google Inc. \\
\texttt{gapapan@google.com} \\
\AND
Andrea Vedaldi \\
Visual Geometry Group\\
University of Oxford\\
\texttt{vedaldi@robots.ox.ac.uk}
}

\begin{document}

\maketitle

\begin{abstract}
In this work we address the task of segmenting an object into its parts, or semantic part segmentation. We start by adapting a \stateoftheart{} semantic segmentation system to this task, and show that a combination of a fully-convolutional Deep CNN system coupled with Dense CRF labelling  provides excellent results for a broad range of object categories. Still, this approach remains agnostic to high-level  constraints between object parts. We introduce such prior information by means of the Restricted Boltzmann Machine, adapted to our task and train our model in an discriminative fashion, as a hidden CRF, demonstrating that prior information can yield additional improvements. We also investigate the performance of our approach ``in the wild'', without information concerning the objects' bounding boxes, using an object detector to guide a multi-scale segmentation scheme. 

We evaluate the performance of our approach on the Penn-Fudan and LFW datasets for the tasks of pedestrian parsing and face labelling respectively. We show superior performance with respect to competitive methods that have been extensively engineered on these benchmarks, as well as realistic qualitative results on part segmentation, even for occluded or deformable objects. We also provide quantitative and extensive qualitative results on three classes from the PASCAL Parts dataset. Finally, we show that our multi-scale segmentation scheme can boost accuracy, recovering segmentations for finer parts.
\end{abstract}

\section{Introduction}\label{sec:introduction}

Recently Deep Convolutional Neural Networks (DCNNs) have delivered excellent results in a broad range of  computer vision problems, including but not limited to image classification \cite{krizhevsky2012imagenet,sermanet2014overfeat, simonyan2014very,szegedy2014going,papandreou2014untangling}, semantic segmentation \cite{chen2014semantic,long2014fully},  object detection \cite{girshick2014rich} and fine-grained categorization \cite{zhang2014part}. Given this broad success, DCNNs seem to have become the method of choice for any image understanding task where the input-output mapping can be described as a classification problem and the output space is a set of discrete labels.

In this work we investigate the use of DCNNs to address the problem of semantic part segmentation, namely segmenting an object into its constituent parts. Part segmentation is an important subproblem for tasks such as recognition, pose estimation, tracking, or applications that require accurate segmentation of complex shapes, such as a host of medical applications. For example, \stateoftheart{} methods for fine-grained categorization rely on the localization and/or segmentation of object parts~\cite{zhang2014part}.

Part segmentation is also interesting from the modeling perspective, as the configurations of parts are, for most objects, highly structured. Incorporating prior knowledge about the parts of an object lends itself naturally to structured prediction, which aims at training a map whose output space has a well defined structure.  The key question addressed in this paper is how DCNN can be combined with structured output prediction to effectively parse object parts. In this manner, one can combine the discriminative power of CNNs to identify part positions and prior information about object layout, to recover from possible failures of the CNN.  Integrating DCNNs with structured prediction is not novel. For example, early work by~\citep{LeCun98} used Graph Transformer Networks for parsing 1D lines into digits. However, the combination of DCNN with models of shape, such as the Shape Boltzmann Machines, or of object parts, such as Deformable Part Models, are recent~\cite{Bregler14,Schwing15,WEF14}. 

This work makes several contributions. First, we show that, by adapting the semantic segmentation system of~\cite{chen2014semantic} (\refsec{sec:dcnn}), it is possible to obtain excellent results in part segmentation. This system uses a dense Conditional Random Field (CRF) applied on top of the output of a DCNN. This simple and non-specialized combination often outperforms specialized approaches to part segmentation and localization by a substantial margin. In \refsec{sec:rbm} we turn to the problem of augmenting this system with a statistical model of the shape of the object and its parts. A key challenge is that the shape of parts is subject to substantial geometric variations, including potentially a variable number of parts per instance, caused by variations in the object pose. We model this variability using Restricted Boltzmann Machines (RBMs). These implicitly incorporate rich distributed mixture models in a representation that is particularly effective at capturing complex localized variations in shape.

In order to use RBMs with DCNNs in a structured-output prediction formulation, we modify RBMs in several ways: first, we use hidden CRF training to estimate the RBM parameters in a discriminative manner, aiming at maximizing the posterior likelihood of the ground-truth part masks given the DCNN scores as input. We demonstrate that this can yield an improvement over the raw DCNN scores by injecting high-level knowledge about the desired object layout.

Extensive experimental results in \refsec{sec:experiments}, confirm the merit of our approach on four different datasets, while in \refsec{sec:multi} we propose a simple scheme to segment objects in the image domain, without knowing their bounding boxes. We conclude in \refsec{sec:discussion} with a summary of our findings.

\section{Related Work}\label{sec:related_work}

The layout of object parts (shape, for short) obeys statistical constraints that can be both strict (e.g. head attached to torso) and diverse (e.g. for hair). As such, accounting for these constraints  requires statistical models that can accommodate multi-modal distributions. Statistical shape models traditionally used in vision, such as Active Appearance Models \cite{CET01} or Deformable Part Models \cite{FGMR10} need to determine in advance a small, fixed number of mixtures (e.g. 3 or 6), which may not be sufficient to encompass the variability of shapes due to viewpoint, rotation, and object deformations. 

A common approch in previous works has been combining appearance features with a shape model to enforce a valid spatial part structure. In~\cite{bo2011shape}, the authors compute appearance and shape features on oversegmentations of cropped pedestrian images from the Penn-Fudan pedestrian dataset~\cite{wang2007object,bo2011shape}. They use color and texture histograms to model appearance and spatial histograms of segment edge orientations as the shape features. The label of each superpixel is estimated by comparing appearance and shape features to a library of exemplar segments. Small segments are sequentially merged into larger ones and simple constraints, (such as that ``head'' appears above ``upper body'' and that ``hair'' appears above ``head'') enforce a consistent layout of parts in resulting segmentations. 

Multi-modal distributions can be naturally captured through distributed representations~\cite{hintonpdp}, which represent data through an assembly of complementary patterns that can be combined in all possible ways. Restricted Boltzmann Machines (RBMs)~\cite{smolensky,hintonrbm} provide a probabilistic distributed representation that can be understood as a discrete counterpart to Factor Analysis (or PCA), while their restricted, bipartite graph topology makes sampling efficient. Stacking together multiple RBMs into a Deep Boltzmann Machine (DBM) architecture allows us to build increasingly powerful probabilistic models of data, as demonstrated for a host of diverse modalities e.g. in~\cite{SalakhutdinovH09}.

In  \cite{eslami2014shape} RBMs are thoroughly studied and assessed as models of shape. The authors additionally introduce the Shape Boltzmann Machine (SBM),  a two-layer network that combines ideas from part-based modelling and DBMs, and show that it is substantially more flexible and expressive than single-layer RBMs. The same approach was extended to deal with multiple region labels (parts) in~\cite{eslami2012generative} and coupled with a model for part appearances. The layered architecture of the model allows it to capture both local and global statistics of the part shapes and part-based object segmentations, while parameter sharing during training helps avoid overfitting despite the small size of the training datasets.


The discriminative training of RBMs has been pursued in shape modelling by  \cite{kae2013augmenting} in a probabilistic setting and by \cite{YangSY14} in a max-margin setting. We pursue a probabilistic setting and detail our approach below. Despite small theoretical differences, the major practical difference between our method and the aforementioned ones is that we do not use any superpixels, pooled features, or boundary signals, as \cite{kae2013augmenting,YangSY14} do, but we rather entirely rely on the CNN scores. 


\section{DCNNs for semantic part segmentation}\label{sec:dcnn}

Deep Convolutional Neural Networks have proven to be particulary successful in ``Semantic Image Segmentation'', the task of pixel-wise labeling of images \cite{sermanet2014overfeat,long2014fully,chen2014semantic}. In this section
we adapt the recently introduced, state-of-art DeepLab system~\cite{chen2014semantic} to our task of semantic part segmentation.

Following~\cite{chen2014semantic}, we adopt the architecture of the state-of-art 16-layer classification network of \cite{simonyan2014very} (VGG-16). We employ it in a fully-convolutional manner, turning it into  a dense feature extractor for semantic image segmentation, as in \cite{sermanet2014overfeat,Oquab13,long2014fully}, treating the last fully-connected layers of the DCNN as $1\times1$ spatial convolution kernels. 
Similarly to \cite{chen2014semantic}, we employ linear interpolation to upsample by a (factor of 8) the class scores of the final network layer to the original image resolution. We learn the DCNN network parameters using training images annotated with semantic object parts at the pixel-level, minimizing the cross-entropy loss averaged over all image positions with Stochastic Gradient Descent (SGD), initializing network parameters from the Imagenet-pretrained VGG-16 model of~\cite{simonyan2014very}.

The model's ability to capture low-level information related to region boundaries is enhanced by employing the fully-connected Conditional Random Field (CRF) of \cite{krahenbuhl2011efficient},
exploiting its ability to combine fine edge details with long-range dependencies. This particularly simple combination is both efficient and effective:  the DCNN evaluation runs at 8 frames per second for a
${321}\times{321}$ image on a GPU and  CRF inference requires $0.5$ seconds on a CPU. Similarly to \cite{chen2014semantic}, we set the dense CRF hyperparameters by cross-validation, performing grid
search to find the values that perform best on a small held-out validation set for each task.

In order to simplify the evaluation of the learned networks we fine-tune one network per object category. The system is thoroughly evaluated in \refsec{sec:experiments}; qualitative results show that it is surprisingly effective in segmenting parts even for objects such as horses that exhibit complicated, articulated deformations. 
While this DCNN + CRF model is very powerful, it can still make gross errors in some cases. Such errors could be corrected by introducing knowledge of the layout of objects, allowing for a better, more principled use of global information. Integrating this information is the goal of \refsec{sec:rbm}.

\section{Conditional Boltzmann Machines}\label{sec:rbm}

The aim of this section is to construct a probabilistic model of image segmentations that can capture prior information on the layout of an object category. The goal of this model is to complement and correct information extracted bottom-up from an image by the DCNN as explained in the previous section.
 
In order to construct this model, we introduce three types of variables: (i) the output $\vv$ of the densely-computed DCNN that is visible during both training and testing; (ii) the binary latent variables $\hh$ that are hidden during both training and testing; and (iii) the ground-truth segmentation labels $\yy$ that are observed during training and inferred during testing. The latter is a one-hot vector for each pixel, with $\yy_{i,k}=1$ indicating that pixel $i$ takes label $k$ out of a set of $K$ possible choices (the parts plus background).

The conditional probability $P(\yy,\hh|\vv;\params)$ of the labels and hidden variables given the observed DCNN features is the Boltzmann-Gibbs distribution
\ba
P(\yy,\hh|\vv;\params) 
= 
\frac
{\exp(-E(\yy,\hh,\vv;\params))}
{\sum_{\yy,\hh}\exp(-E(\yy,\hh,\vv;\params))}
\ea
where $E(\yy,\hh,\vv;\params)$ is an energy function described below. The posterior probability of the labelling is obtained by marginalizing the latent variables:
\ba
P(\yy|\vv;\params) = \sum_{\hh} P(\yy,\hh|\vv;\params).
\ea
The goal is to estimate the parameters $\params$ of the energy function during training and to use $P(\yy|\vv;\params)$ during testing to drive inference towards more probable segmentations.

Before describing the energy function $E(\yy,\hh|\vv;\params)$ in detail note that (i) the DCNN-based quantities $\vv$ are always observed and the model does not describe their distribution; in other words, we construct a \emph{conditional} model of $\yy$ \cite{LaffertyMP01,HeZC04}; (ii) unlike common CRFs, there are also hidden variables $\hh$, which results in a Hidden Conditional Random Fields (HCRFs)~\cite{Quattoni07,Murphy}; (iii) however, unlike the loopy graphs used in generic HCRFs, the factor graph in this model is bipartite, which makes block Gibbs sampling possible (\secref{sec:param-est}). 

Consider now the relationship between the DCNN output $\vv$ and the pixel label $\yy$ and recall that $\vv$ are obtained from the last layer of the DCNN. The DCNN is trained so that, for a given pixel $i$, $\vv_i$ contains the class posteriors up to the softmax operation: 
\ba
P(\yy_{i,k}=1|\vv) = \frac{\exp(\vv_{i,k})}{\sum_{k'=1}^{K} \exp(\vv_{i,k'})}.
\ea
This suggests that $\vv_{i,k}$ can be used as a bias term for $\yy_{i,k}$ in the energy model, such that a larger value of $\vv_{i,k}$ rewards the assignment $\yy_{i,k}=1$. The raw values of $\vv$  are rescaled using a set of learnable parameters which allows auto-calibration during training. The contribution to the energy term is then:
\ba
E_{\cnn}(\yy,\vv;\params) = \sum_{i,k,k'} \cnnparams_{k,k'} \yy_{i,k} \vv_{i,k'},
\ea
where the CNN calibration parameters, $\cnnparams_{\cdot,\cdot}$ are contained in the overall model parameters $\params$.  Note also that this formulation allows to learn interactions between classes as class $k'$ as predicted by the DCNN can vote through weight $\cnnparams_{k,k'}$ for class $k$ in the energy.

We can now write the term linking output and hidden variables, which takes the form of an RBM:
\ba
E_{\rbm}(\yy,\hh;\params) =  \sum_{i,j,k}  \yy_{i,k} \rbmparams_{i,j,k} \hh_{j} +  \sum_{i,k}  \yy_{i,k} \rbmparams_{i,k}.
\ea
Note that this does not include any `lateral' connection between the observed variables, or between the hidden variables (this would correspond to terms in which pairs of the same type of variables are multiplied). Instead, there are two types of terms. The first type has biases $\rbmparams_{i,k}$ for each pixel location $i$ and label $k$, favoring certain labels based on their spatial location only. The second type expresses the interaction between  labels and latent variables through the interaction weights $\rbmparams_{i,j,k}$. These weights determine the effect that activating the hidden node $\hh_j$ has on labelling position $i$ as part $k$. Activating $\hh_j$ will favor or discourage simultaneously the activation of labels at different locations according to the pattern encoded by the weights $\rbmparams_{\cdot,j,\cdot}$ -- intuitively latent variables can in this manner encode segmentation fragments.

The overall energy is obtained as the sum of these two terms:
\ba
E(\yy,\hh,\vv;\params) = E_{\cnn}(\yy,\vv;\params)  + E_{\rbm}(\yy,\hh;\params)
\ea
By aggregating the output variables $\yy$,  the hidden variables $\hh$, and the observable variables $\vv$ into a single vector $\zz$,  the energy above can be rewritten in the form:
\ba 
E(\zz;\params) = \zz^T \wf \zz \label{eq:joint}
\ea
where $\params$ is a matrix of interactions. 

\subsection{Parameter Estimation for Conditional RBMs}\label{sec:param-est}
Given a set of $M$ training examples $\trset = \{(\yy^1,\vv^1),\ldots,(\yy^M,\vv^M)\}$, parameter estimation aims at maximizing the conditional log-likelihood of the ground-truth labels:
\begin{gather}
S(\wf) = \sum_{m=1}^{M} \log P(\yy^m|\vv^m;\wf) =  \sum_{m=1}^{M} \log \sum_{\hh} \frac{\exp(-E(\yy^m,\hh,\vv;\wf))}{Z(\vv^m)}, \label{eq:log}\\ 
\mathrm{where}~~Z(\vv^m) = \sum_{\yy,\hh}{\exp(-E(\yy,\hh,\vv;\wf))}.
\end{gather}

\renewcommand{\param}{\mathbf{w}}
Using the notation of \refeq{eq:joint}, a parameter $\wf_{k,m}$ connects nodes $\zz_k$ and $\zz_m$ that can be either hidden or visible. 
The partial derivative of the conditional log-likelihood with respect to $\wf_{k,m}$ is given by \cite{Murphy}:
\ba
\frac{\partial S}{\partial \wf_{k,m}} = 
\sum_{m=1}^{M} \left< \zz_k \zz_m \right>_{P(\hh|\yy^m,\vv^m;\wf)} - 
\left< \zz_k \zz_m \right>_{P(\hh,\yy|\vv^m;\wf)} \nonumber
\ea
where $\left< \cdot,\cdot \right>$ denotes expectation.

In order to compute the first term, the $\yy$ and $\vv$ components  of the $\zz$ vector are given and one has to average over the posterior on $\hh$ to compute the expectation of $\zz_k\zz_m$. To do so, one starts with the CNN scores ($\vv$) and the ground-truth segmentation maps ($\yy$) and computes the posterior over the hidden variables $\hh$, which can be obtained analytically.  Then he computes the expectation of the product of any pair of interacting nodes, also in closed form.

In order to compute the second term one needs to consider the joint expectation over segmentations $\yy$ and hidden variables $\hh$ when presented with the CNN scores $\vv$. The exact computation of this term is intractable, and is instead computed through Monte Carlo approximation using Contrastive Divergence \cite{hintonreport}. Namely we initialize the state $\yy$ to $\yy^m$, perform $C =10$ iterations of Block-Gibbs sampling over $\yy$ and $\hh$, and use the resulting state as a sample from $P(\hh,\yy|\vv^m;\wf)$. 

This training algorithm is  identical to RBM training with the difference that the partition function is image-dependent, resulting in minor algorithmic modifications.

%
%


\section{Experiments}\label{sec:experiments}
We evaluate our method on four datasets (LFW,Penn-Fudan,CUB and PASCAL-parts) and report qualitative and quantitative results. 
We compare the accuracy of our pipeline before and after refining part boundaries using the fully-connected CRF, and  also report on the improvements delivered by the combination of RBMs with CNNs on three categories (faces, cows, horses). 
While using  the exact same settings for the network and parameter values described in~\refsec{sec:dcnn}, we obtain state-of-the-art results when comparing to carefully engineered approaches for the individual problems. 

\subsubsection*{Penn-Fudan pedestrian datataset}\label{sec:penn-fudan}
The Penn-Fudan dataset~\cite{bo2011shape} provides manual segmentations of 170 pedestrians into head, hair, clothes, arms, legs and shoes/feet. This dataset does not come with a train/test split, so we had to train our networks on a different dataset. We finetune our network on the Pascal~\emph{person} category, using all images and corresponding part annotations from~\cite{chen2014detect}. 

A complication is that in PASCAL-Parts clothing is not taken into account when segmenting people - the only regions are ``torso'', ``arms'', ``legs'' and ``feet''; whereas in Penn-Fudan the semantic parts used are ``hair'', ``face'', ``upper clothes'', ``arms'', ``lower clothes'', ``legs'' and ``shoes/feet''. To facilitate comparison of the methods, we merge ``torso'' and ``arms'' from PASCAL and ``upper clothes'' and ``arms'' from Penn-Fudan into ``upper body''; similarly we merge  ``legs'' and ``feet'' from PASCAL and ``lower clothes'', ``legs'' and ``feet'' from Penn-Fudan into ``lower body''. 
Other methods also report results on these two superregions, making comparison possible.   
Detailed numbers for Intersection-over-Union (IOU) for each part are included in Tab.~\ref{tab:pedestrian}.

\begin{table*}
\subfloat[Segmentation accuracies on Penn-Fudan.]{
\resizebox{0.65\linewidth}{!}{
    \begin{tabular}{|c|c|c|c|c|c|c|}
        \hline
        Method                              & head           & upper body       & lower body     & FG             & BG         & Average          \\\hline
        SBP~\cite{bo2011shape}              & 51.6            & 72.6             & 71.6           & 73.3            & 81.0      & 70.3             \\\hline
        DDN~\cite{luo2013pedestrian}     & 60.2              & 75.7             & 73.1           & 78.4                & 85.0      & 74.5            \\\hline
        DL~\cite{luo2013pedestrian}         & 60.0              & 76.3             & 75.6               & 78.7            & 86.30        & 75.4          \\\hline
        Ours (CNN)                        & \bf{67.8}           & 77.0             & 76.0            & 83.0             & 85.4       & 77.8            \\\hline
        Ours (CNN+CRF)                     & {64.2}        & \bf{81.5}       & \bf{80.9}     & \bf{84.4}        & \bf{87.3}    & \bf{79.7}     \\\hline
    \end{tabular}
    \label{tab:pedestrian}
    }}
    \subfloat[LFW (superpixel accuracies).]{
    \resizebox{0.3\linewidth}{!}{
    \begin{tabular}{|c|c|c|}
        \hline
        Method                              & Accuracy (SP)    \\\hline
        GLOC~\cite{kae2013augmenting}     & 94.95\%         \\\hline
        Ours (CNN)                           & 96.54\%          \\\hline
        Ours (CNN+RBM)                       & 96.78\%          \\\hline
        Ours (CNN+CRF)                      & 96.76\%           \\\hline
        Ours (CNN+RBM+CRF)                  & \bf{96.97\%}  \\\hline
    \end{tabular}
    \label{tab:lfw}
    }}
    \caption{Segmentation accuracies of our system on the Penn-Fudan and LFW datasets}
\end{table*}

\newcommand{\pennheight}{1.55cm}
\newcommand{\pennwidth}{0.05\textwidth}
\newcommand{\hpascal}{\pennheight}
\newcommand{\wpascal}{0.12}
\begin{figure*}[t!]
\centering
\includegraphics[height=\pennheight, width=\pennwidth]{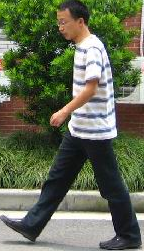}
\includegraphics[height=\pennheight, width=\pennwidth]{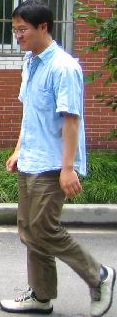}
\includegraphics[height=\pennheight, width=\pennwidth]{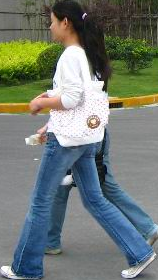}
\includegraphics[height=\pennheight, width=\pennwidth]{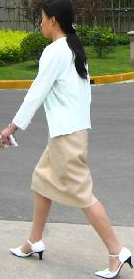}
\includegraphics[height=\pennheight, width=\pennwidth]{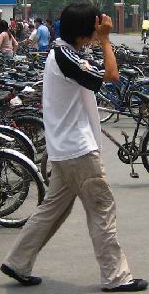}
\includegraphics[height=\pennheight, width=\pennwidth]{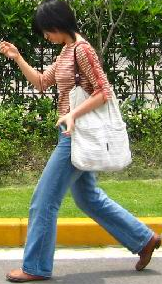}
\includegraphics[height=\pennheight, width=\pennwidth]{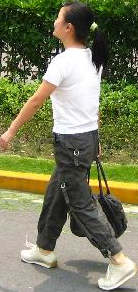}
\includegraphics[height=\pennheight, width=\pennwidth]{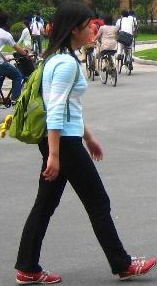}\,
\includegraphics[height=\hpascal, width=\wpascal\textwidth]{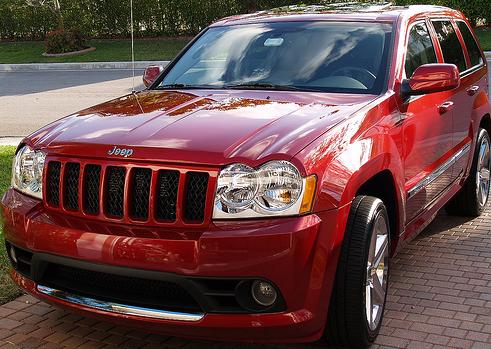}
\includegraphics[height=\hpascal, width=\wpascal\textwidth]{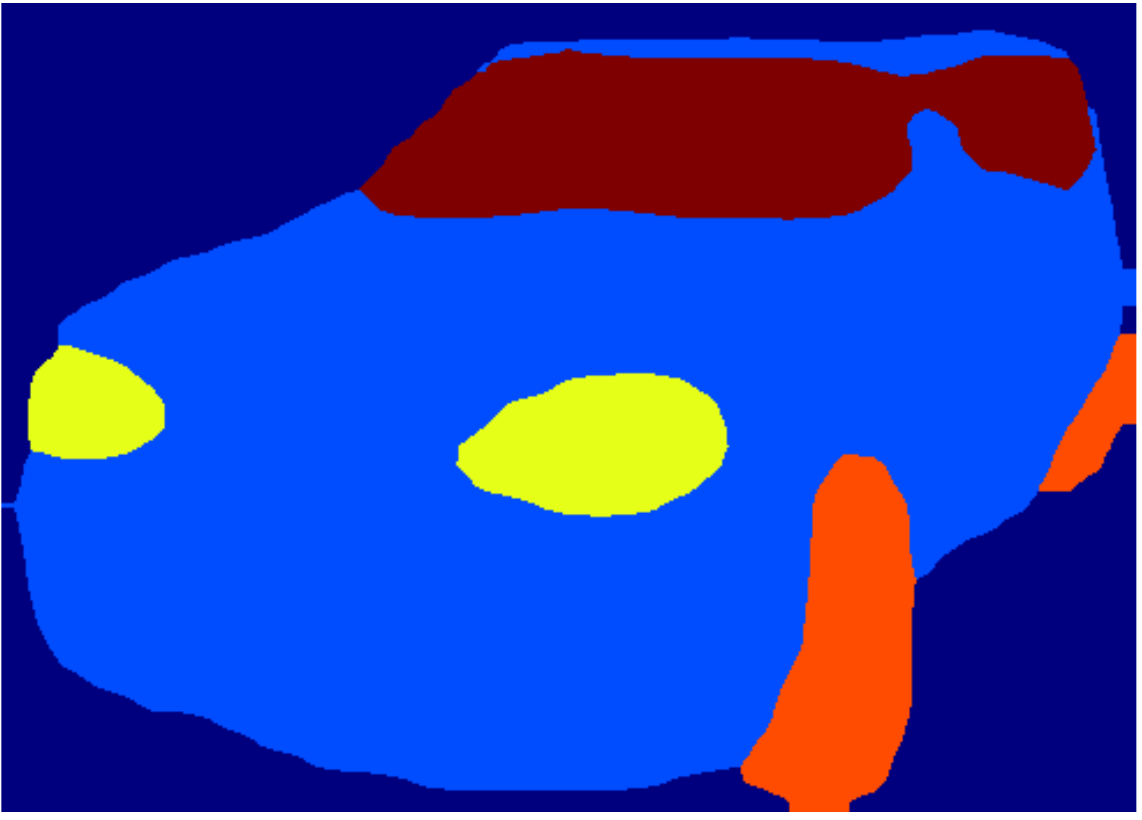}
\includegraphics[height=\hpascal, width=\wpascal\textwidth]{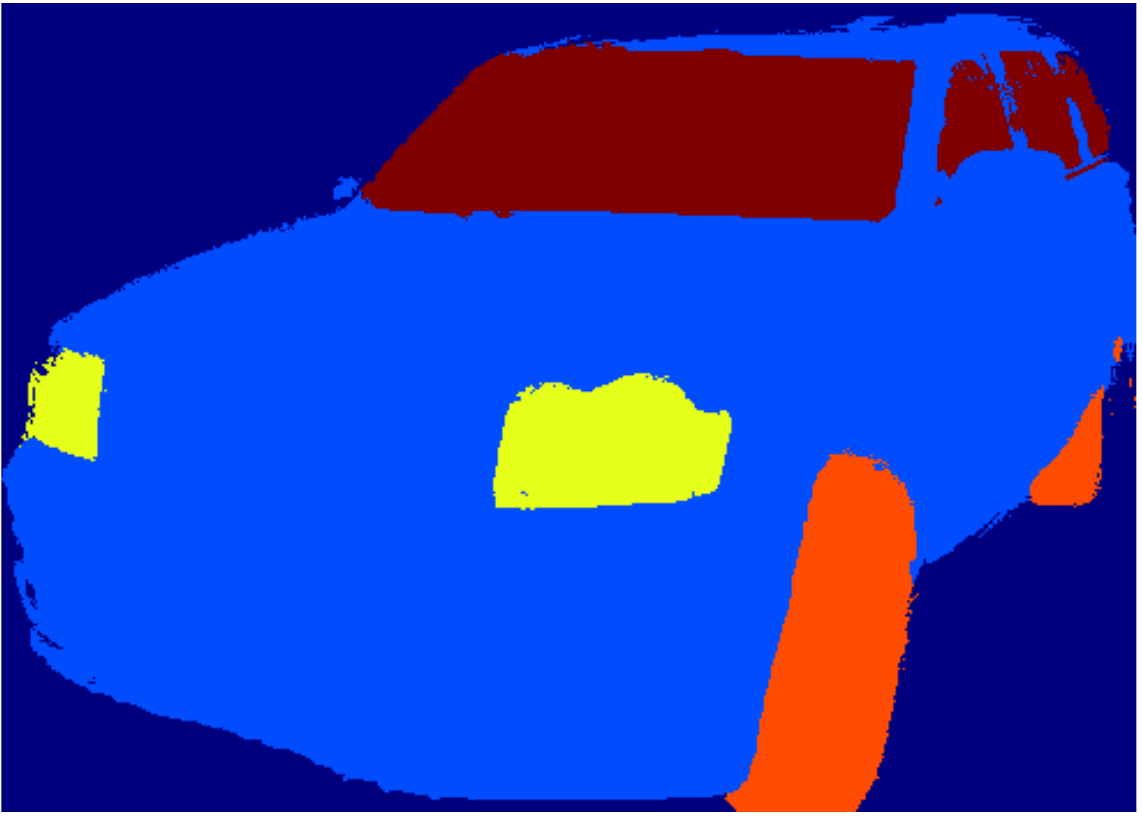}
\includegraphics[height=\hpascal, width=\wpascal\textwidth]{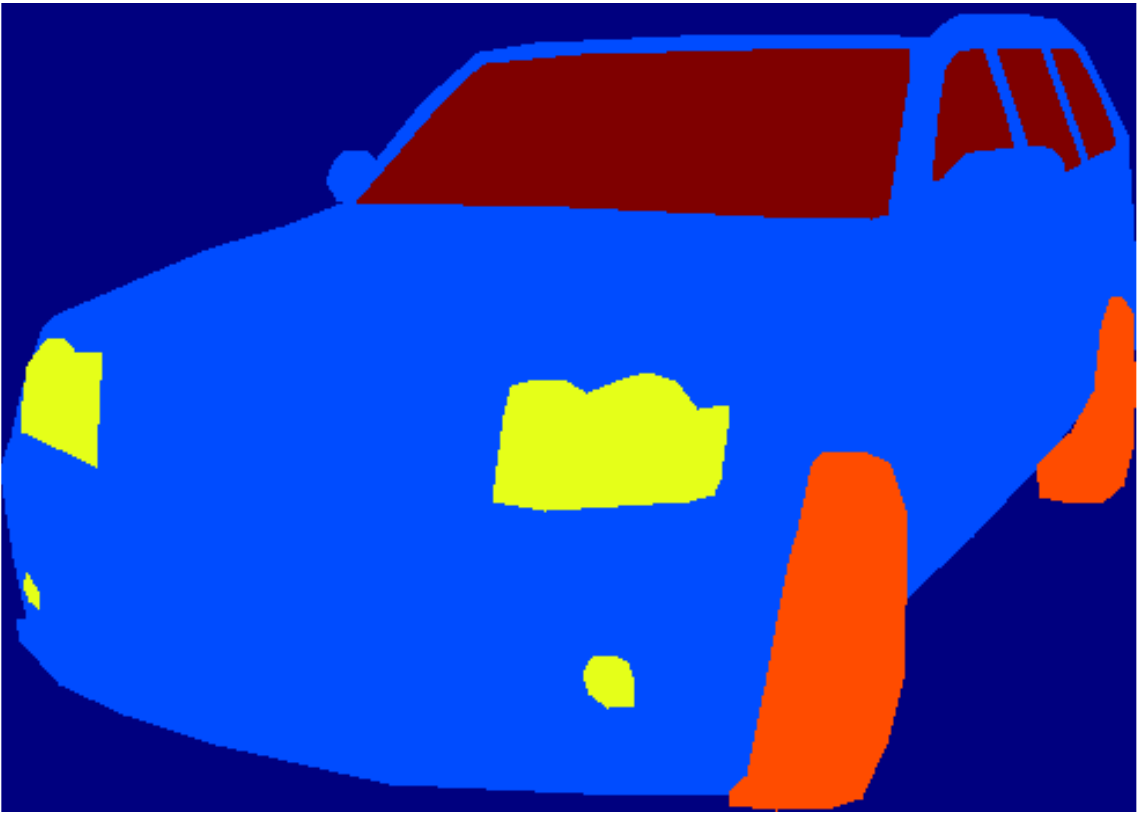}


\includegraphics[height=\pennheight, width=\pennwidth]{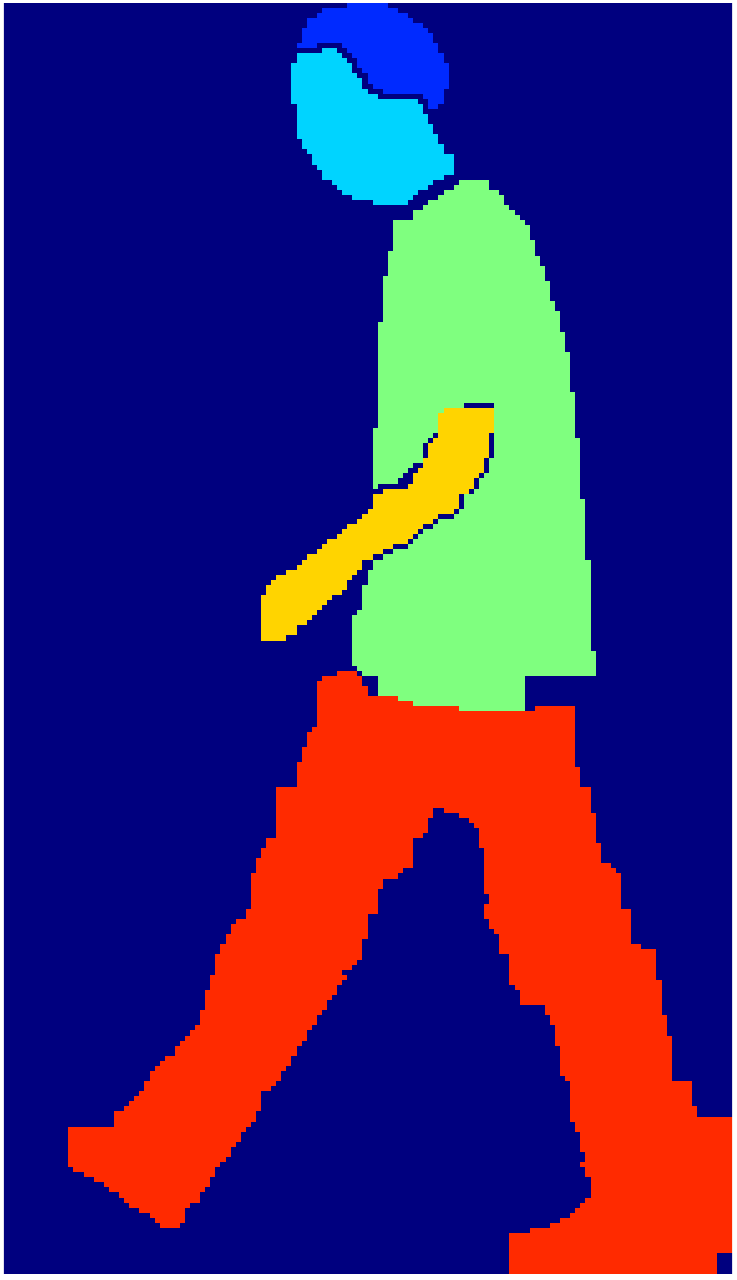}
\includegraphics[height=\pennheight, width=\pennwidth]{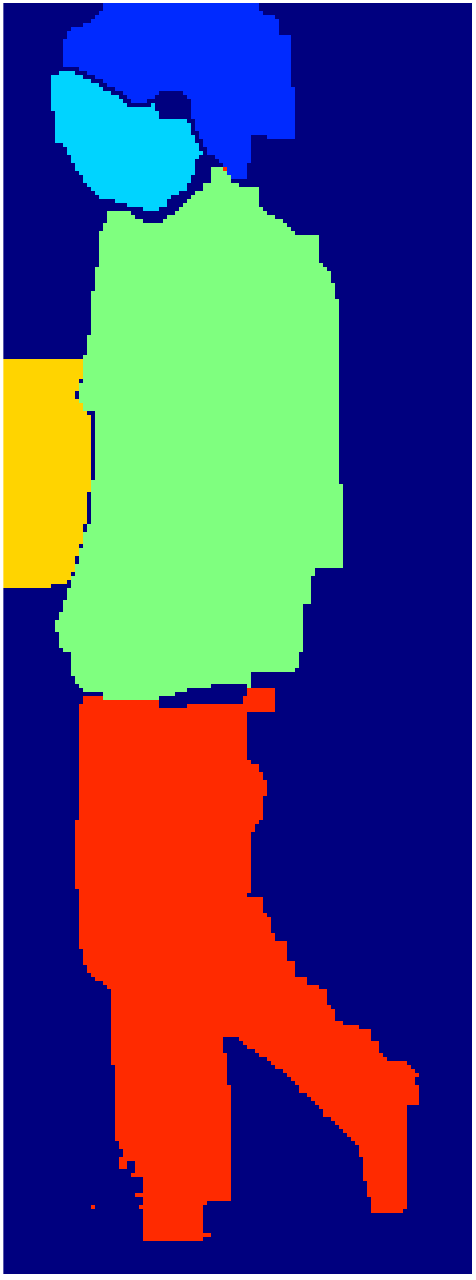}
\includegraphics[height=\pennheight, width=\pennwidth]{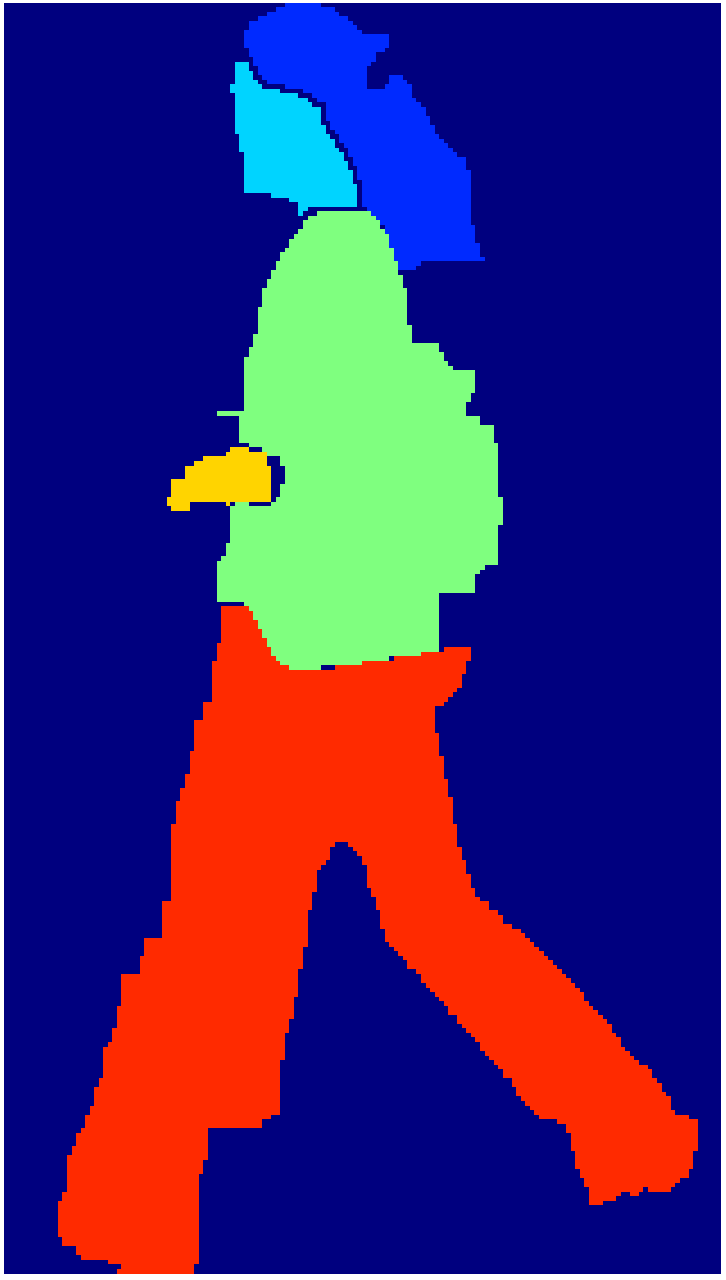}
\includegraphics[height=\pennheight, width=\pennwidth]{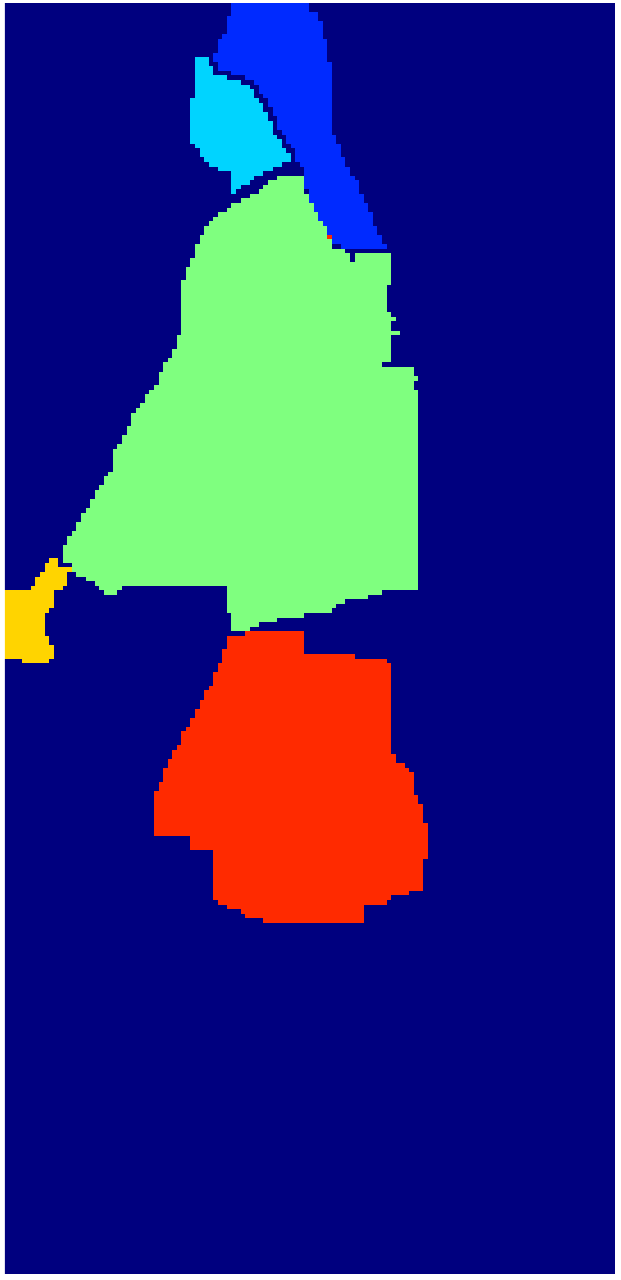}
\includegraphics[height=\pennheight, width=\pennwidth]{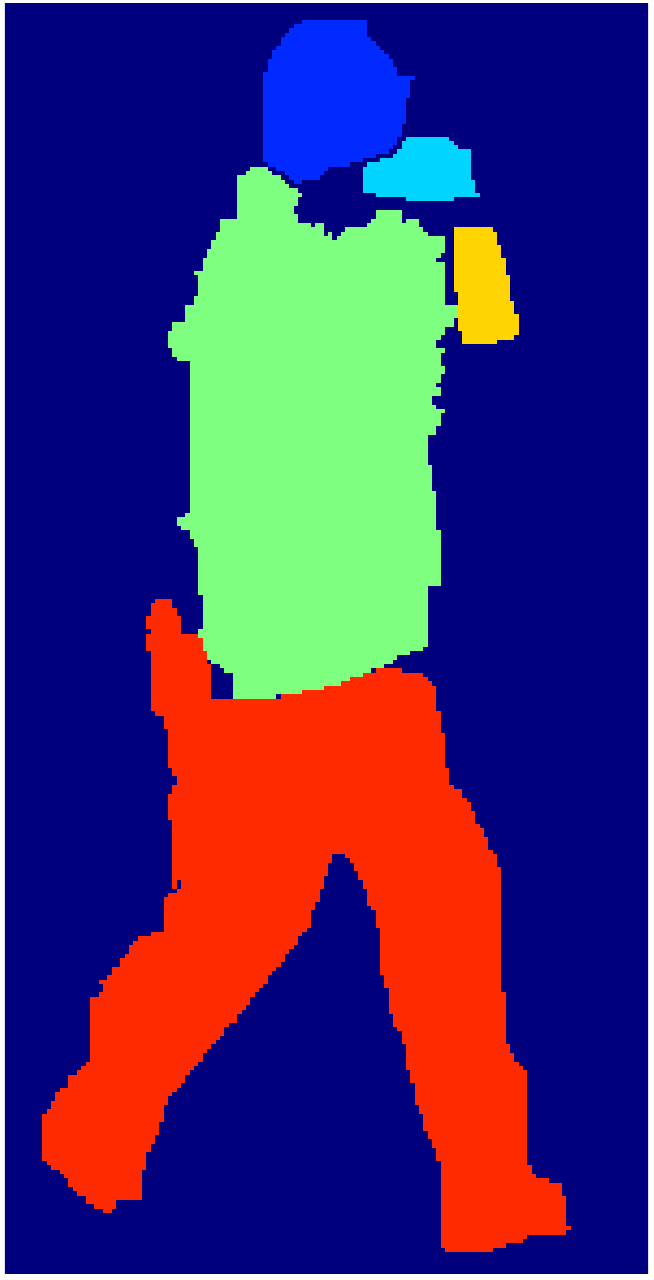}
\includegraphics[height=\pennheight, width=\pennwidth]{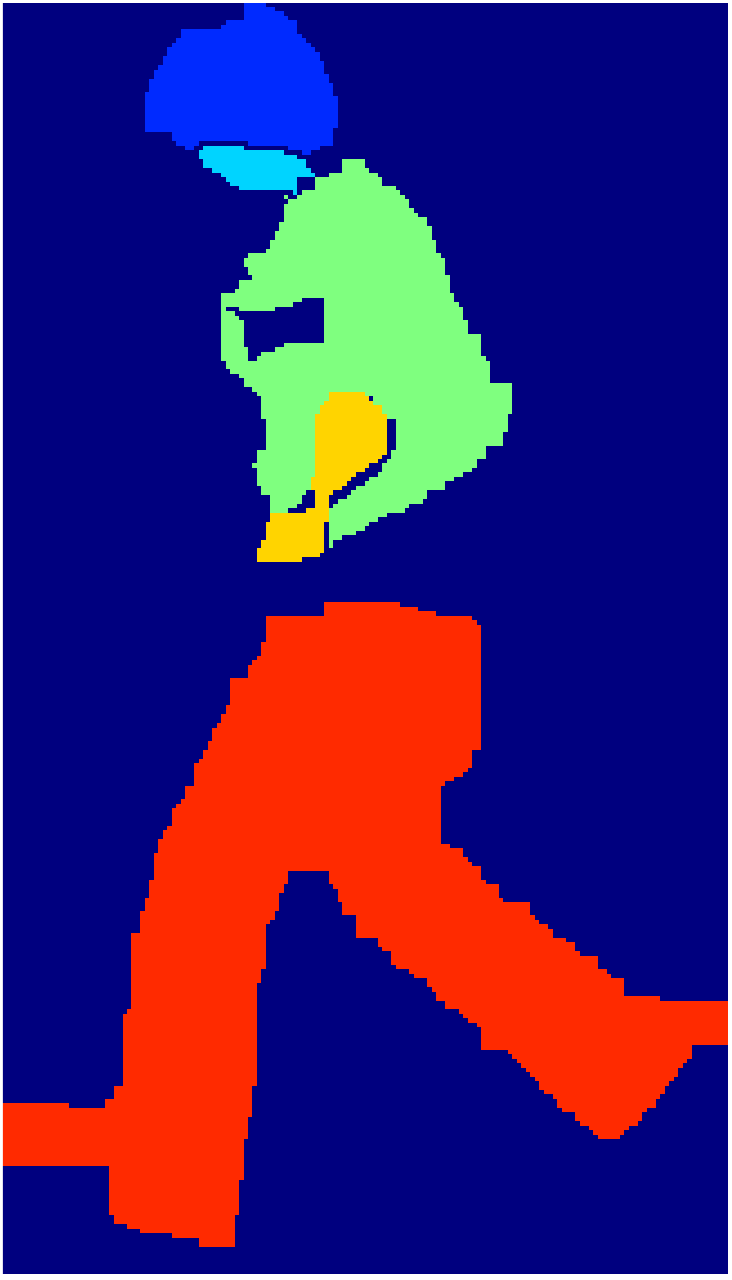}
\includegraphics[height=\pennheight, width=\pennwidth]{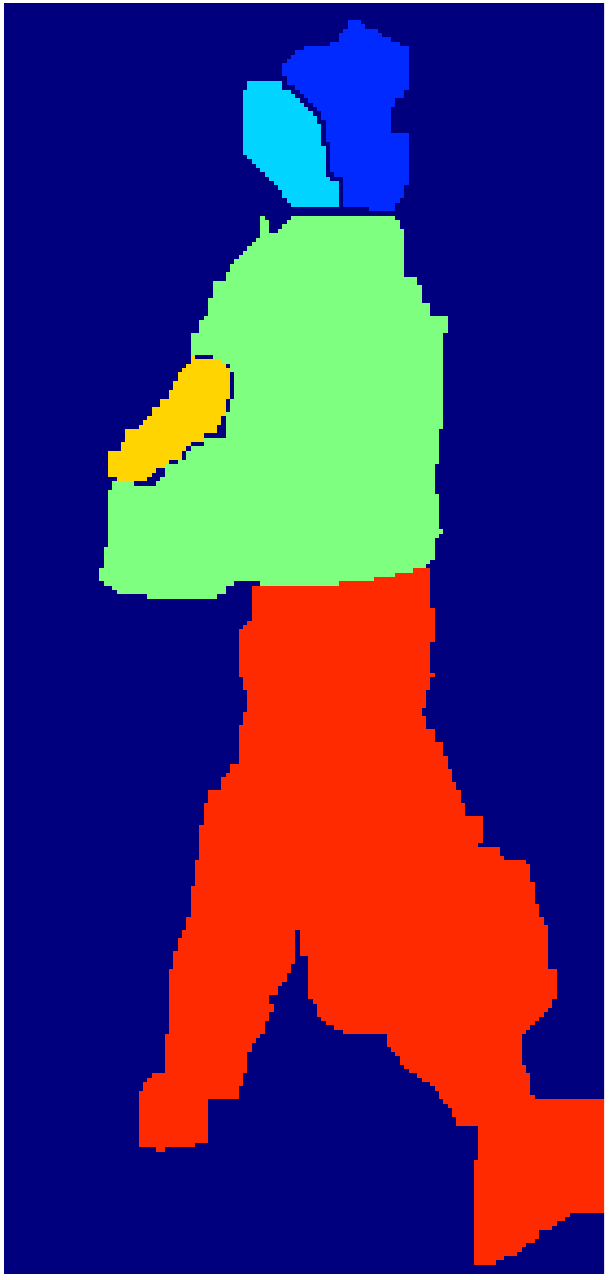}
\includegraphics[height=\pennheight, width=\pennwidth]{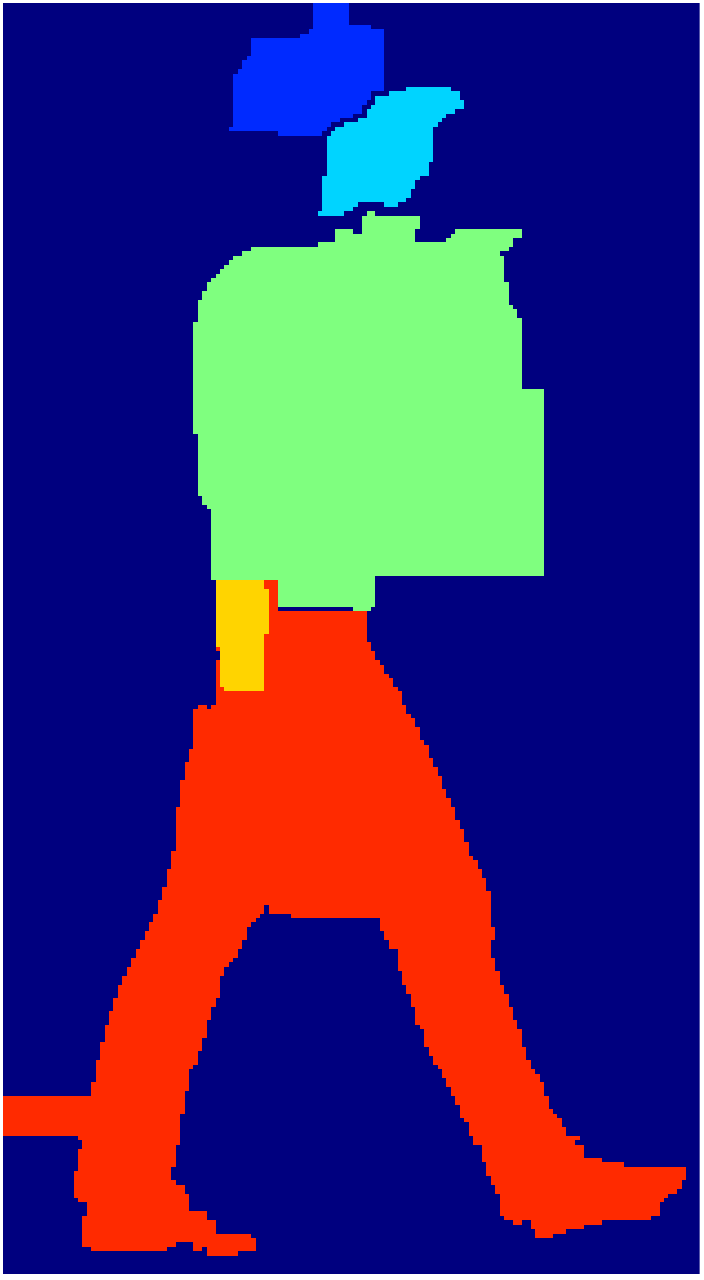}\,
\includegraphics[height=\hpascal, width=\wpascal\textwidth]{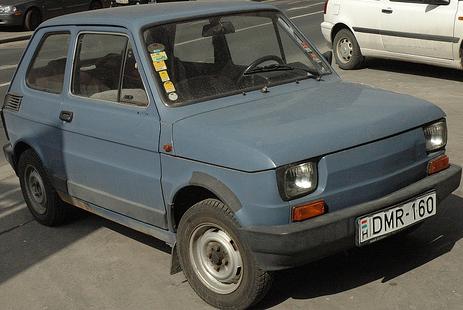}
\includegraphics[height=\hpascal, width=\wpascal\textwidth]{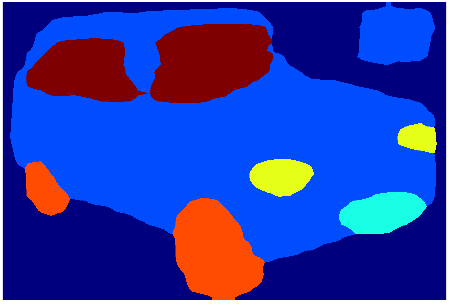}
\includegraphics[height=\hpascal, width=\wpascal\textwidth]{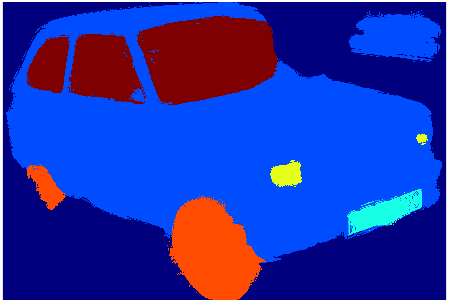}
\includegraphics[height=\hpascal, width=\wpascal\textwidth]{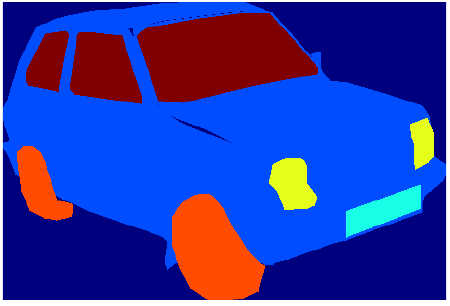}

\includegraphics[height=\pennheight, width=\pennwidth]{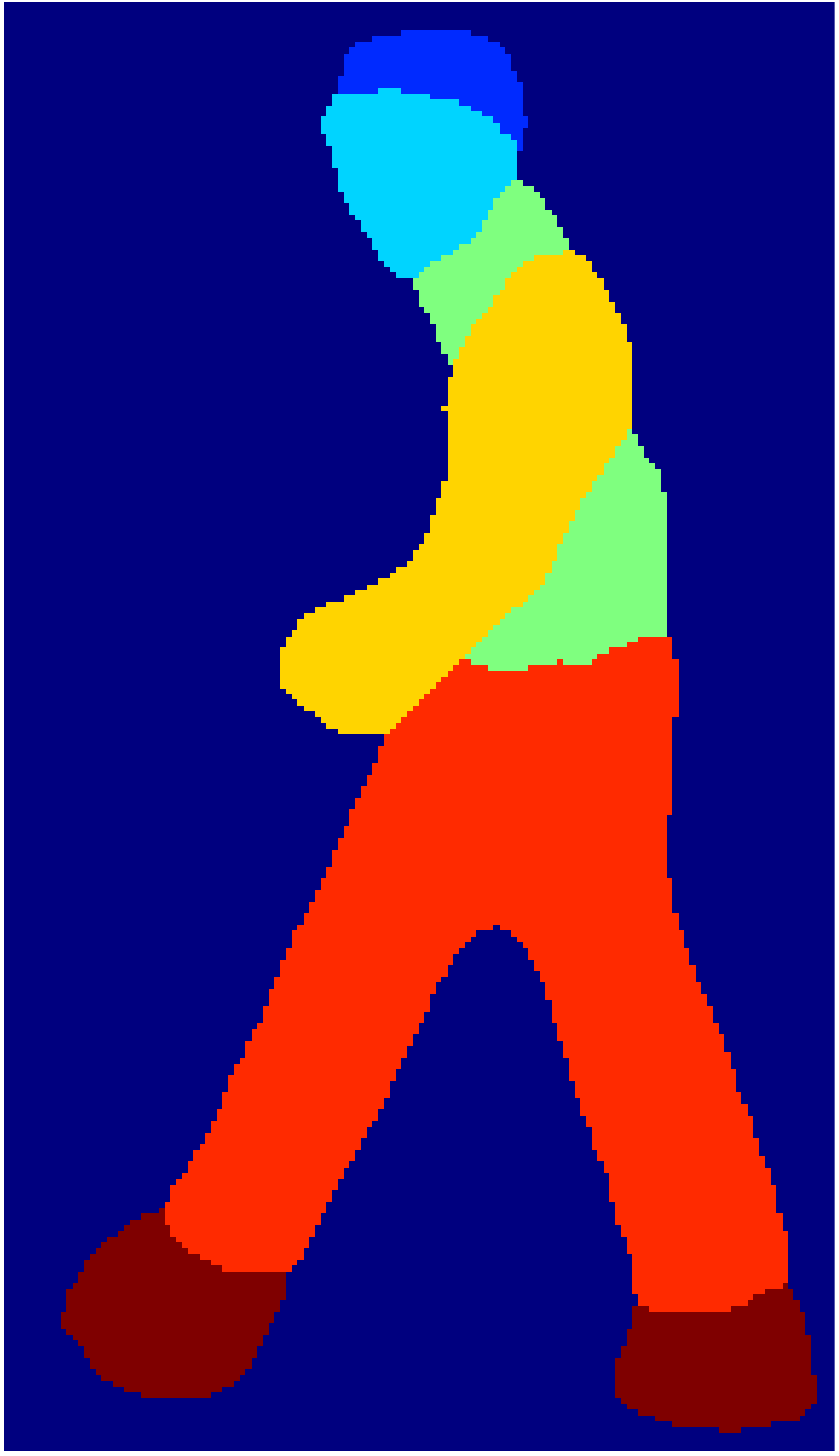}
\includegraphics[height=\pennheight, width=\pennwidth]{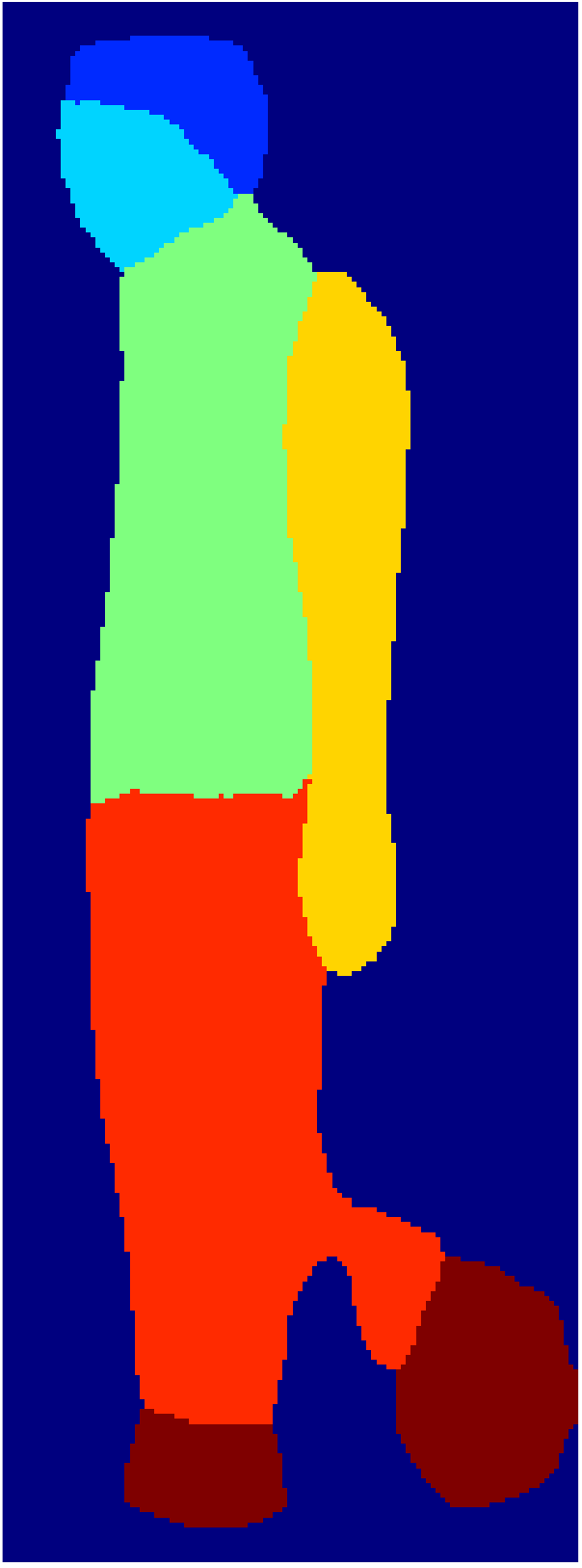}
\includegraphics[height=\pennheight, width=\pennwidth]{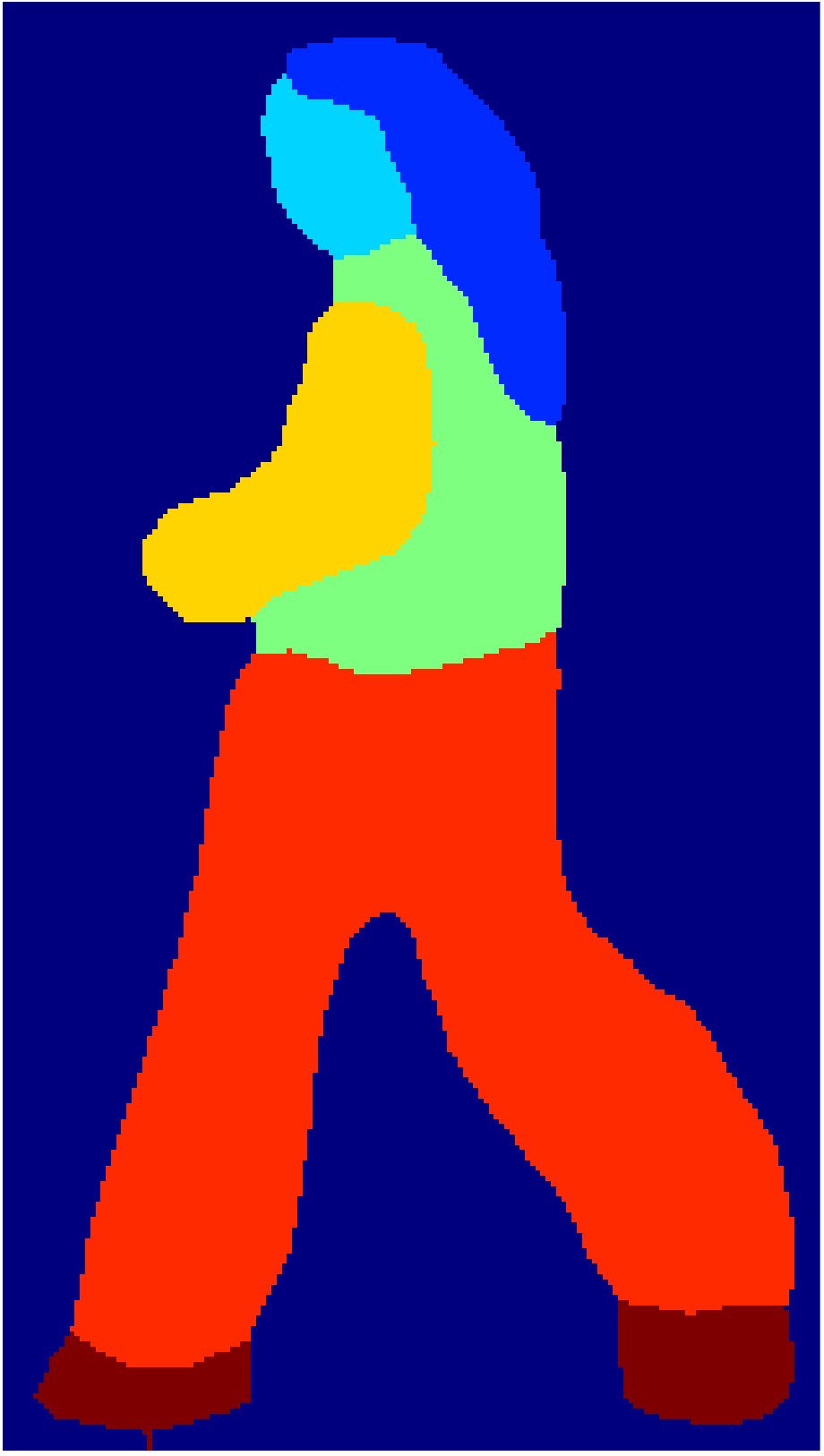}
\includegraphics[height=\pennheight, width=\pennwidth]{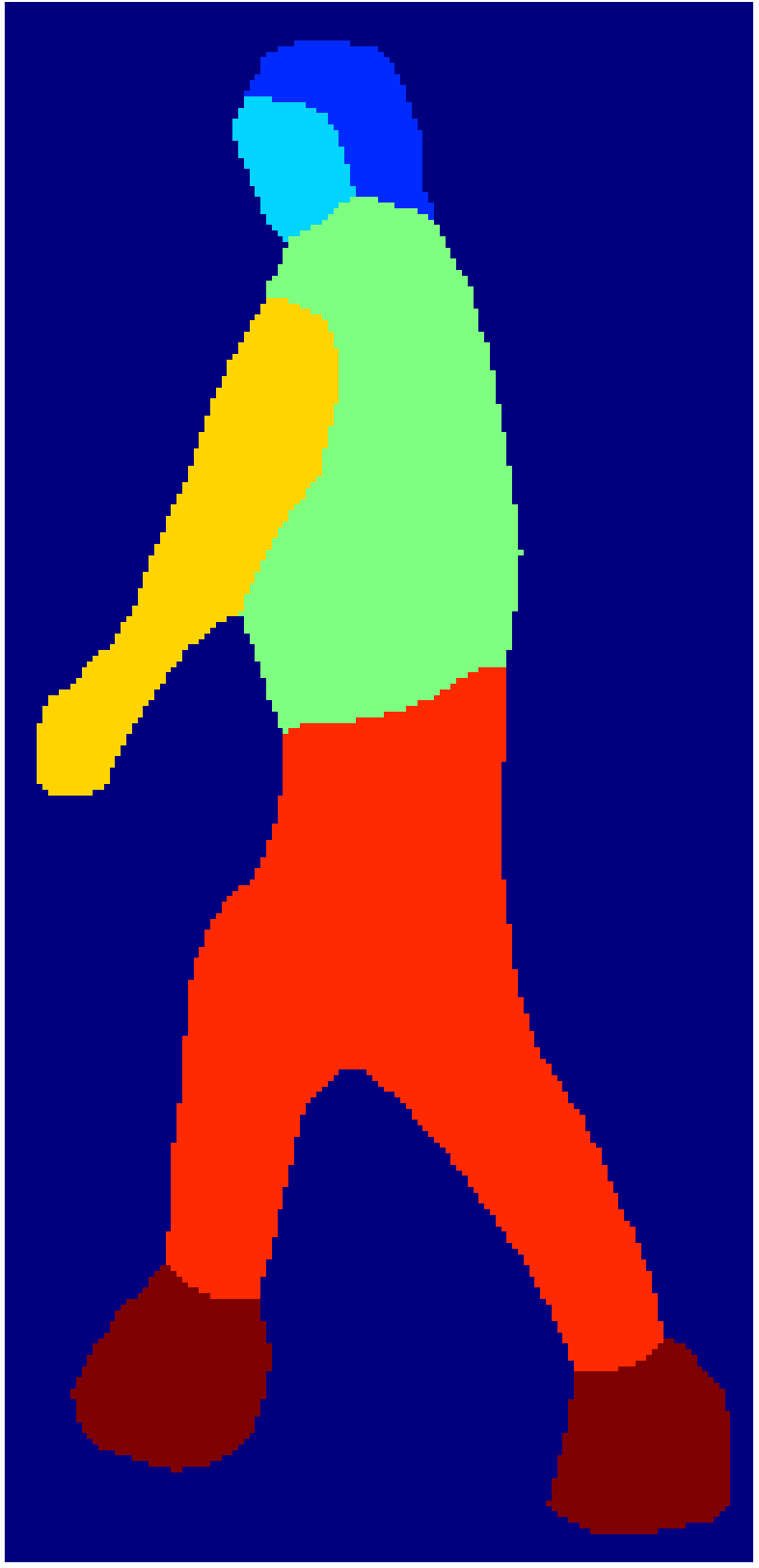}
\includegraphics[height=\pennheight, width=\pennwidth]{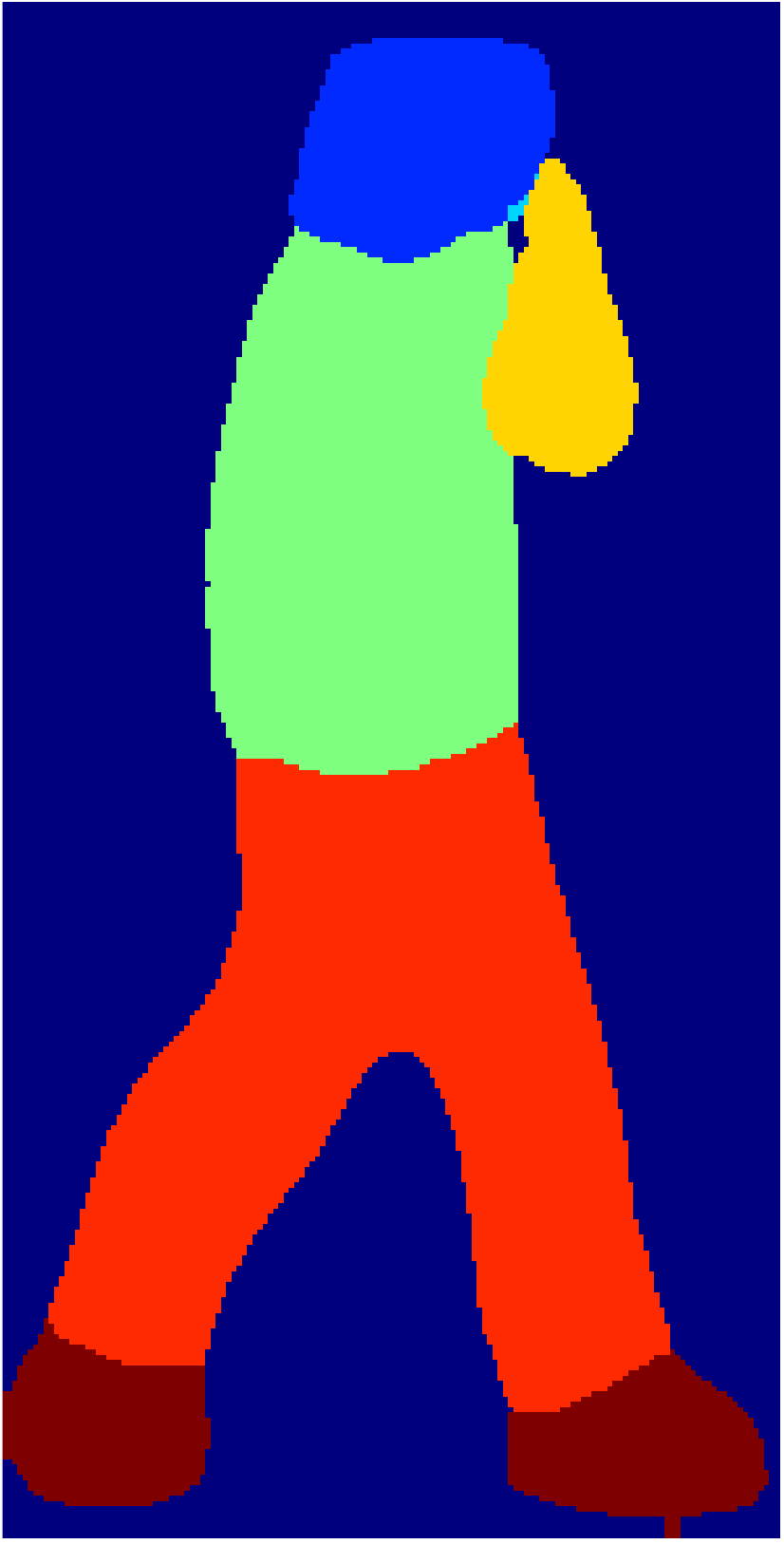}
\includegraphics[height=\pennheight, width=\pennwidth]{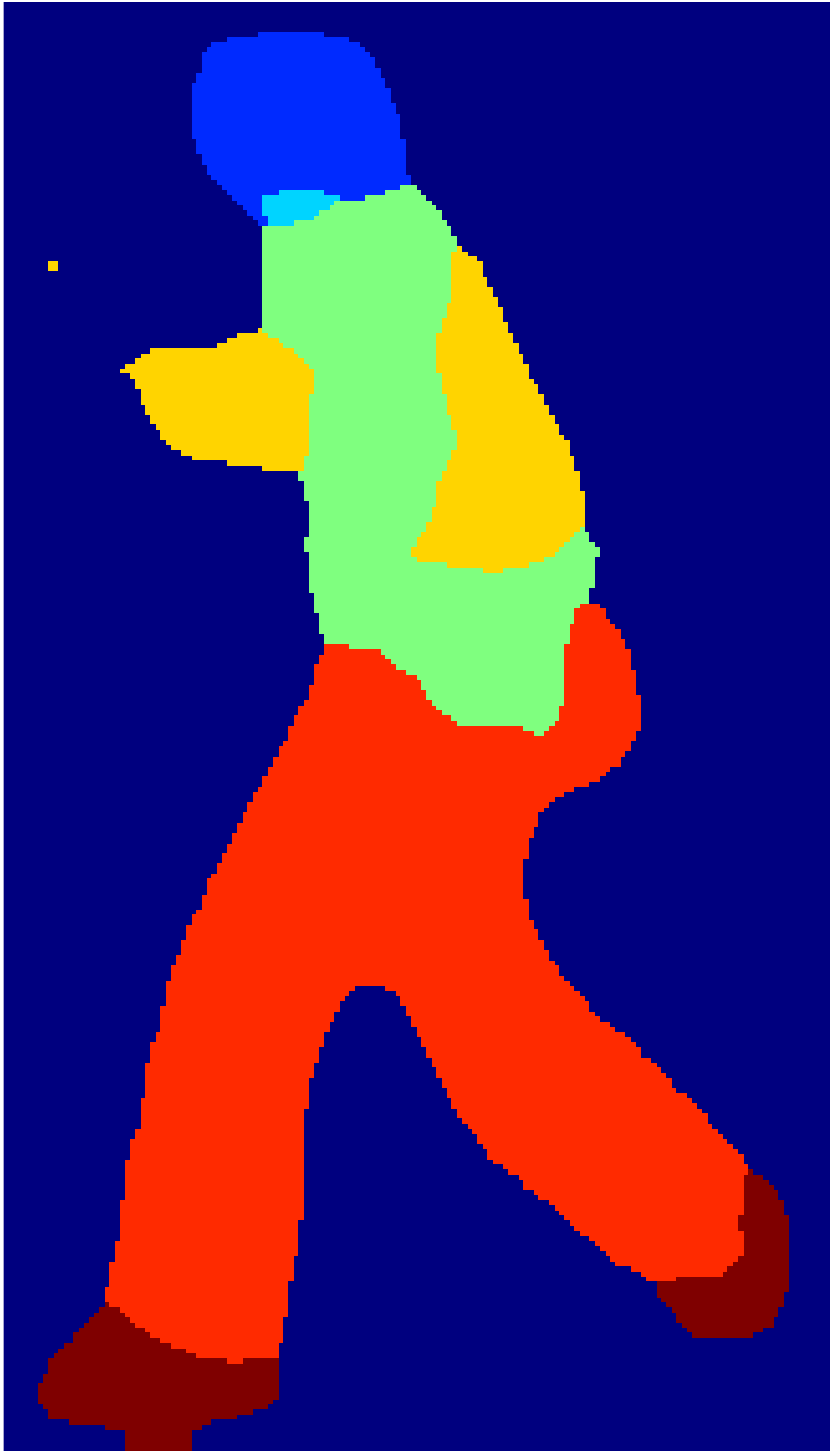}
\includegraphics[height=\pennheight, width=\pennwidth]{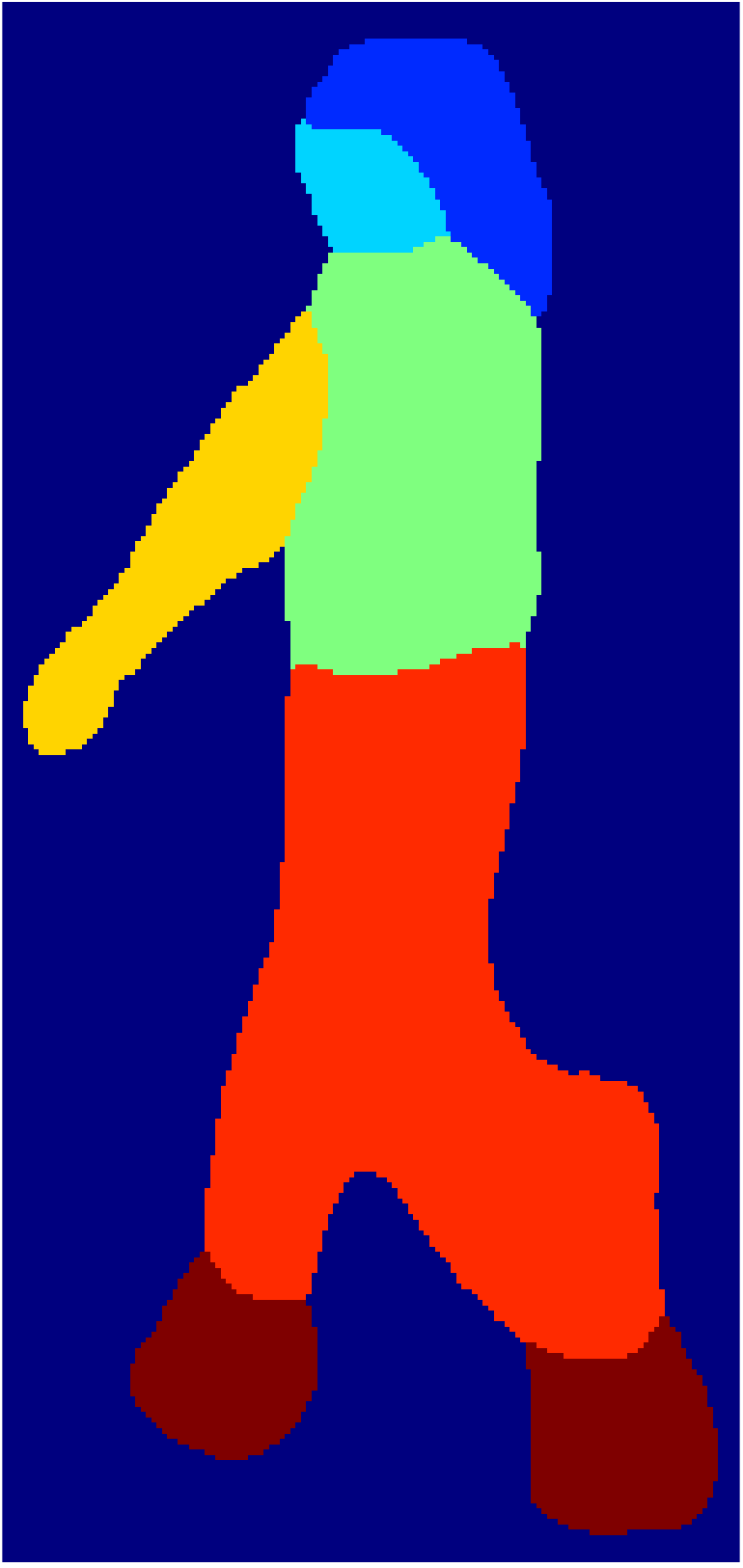}
\includegraphics[height=\pennheight, width=\pennwidth]{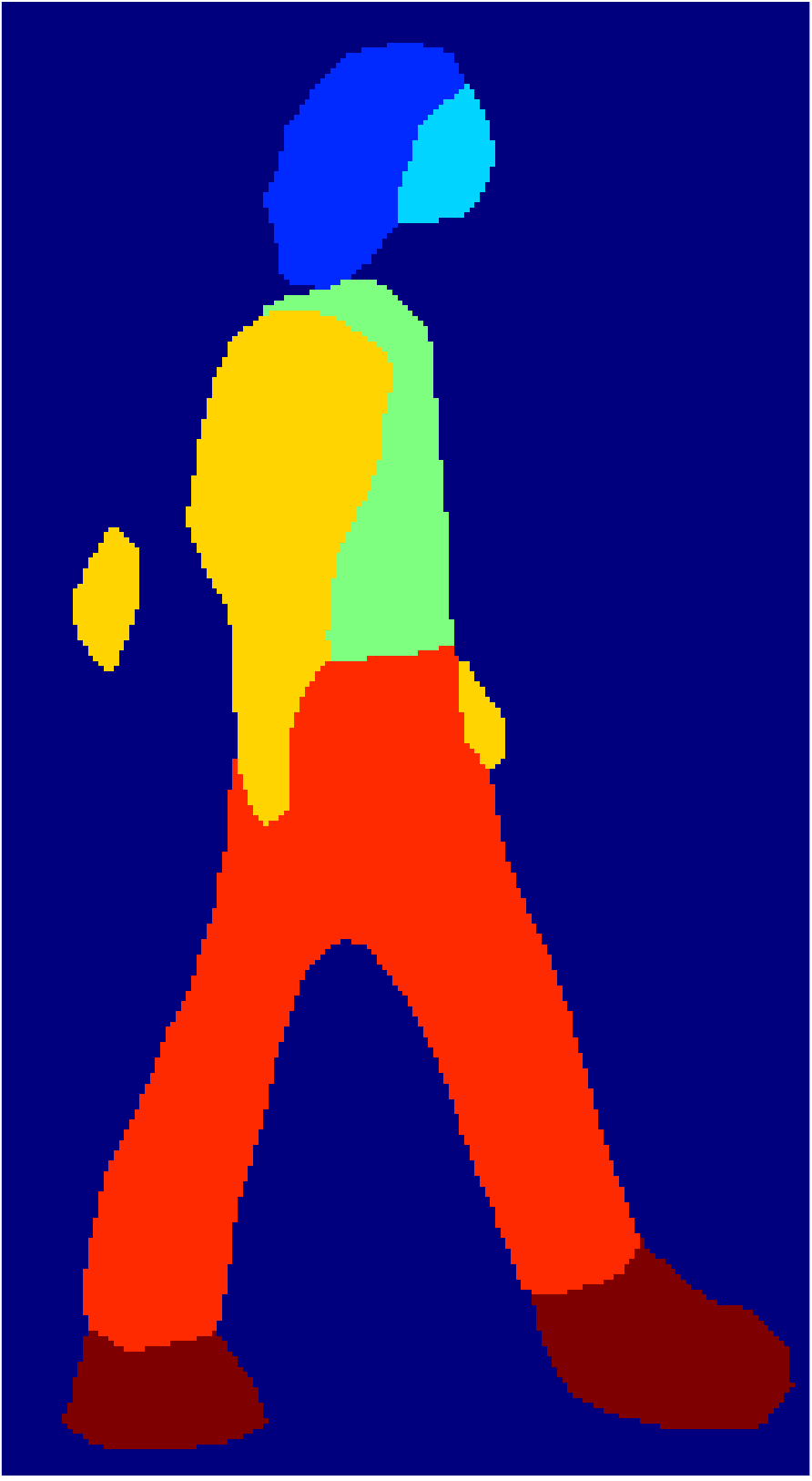}\,
\includegraphics[height=\hpascal, width=\wpascal\textwidth]{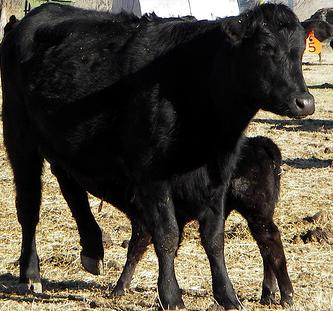}
\includegraphics[height=\hpascal, width=\wpascal\textwidth]{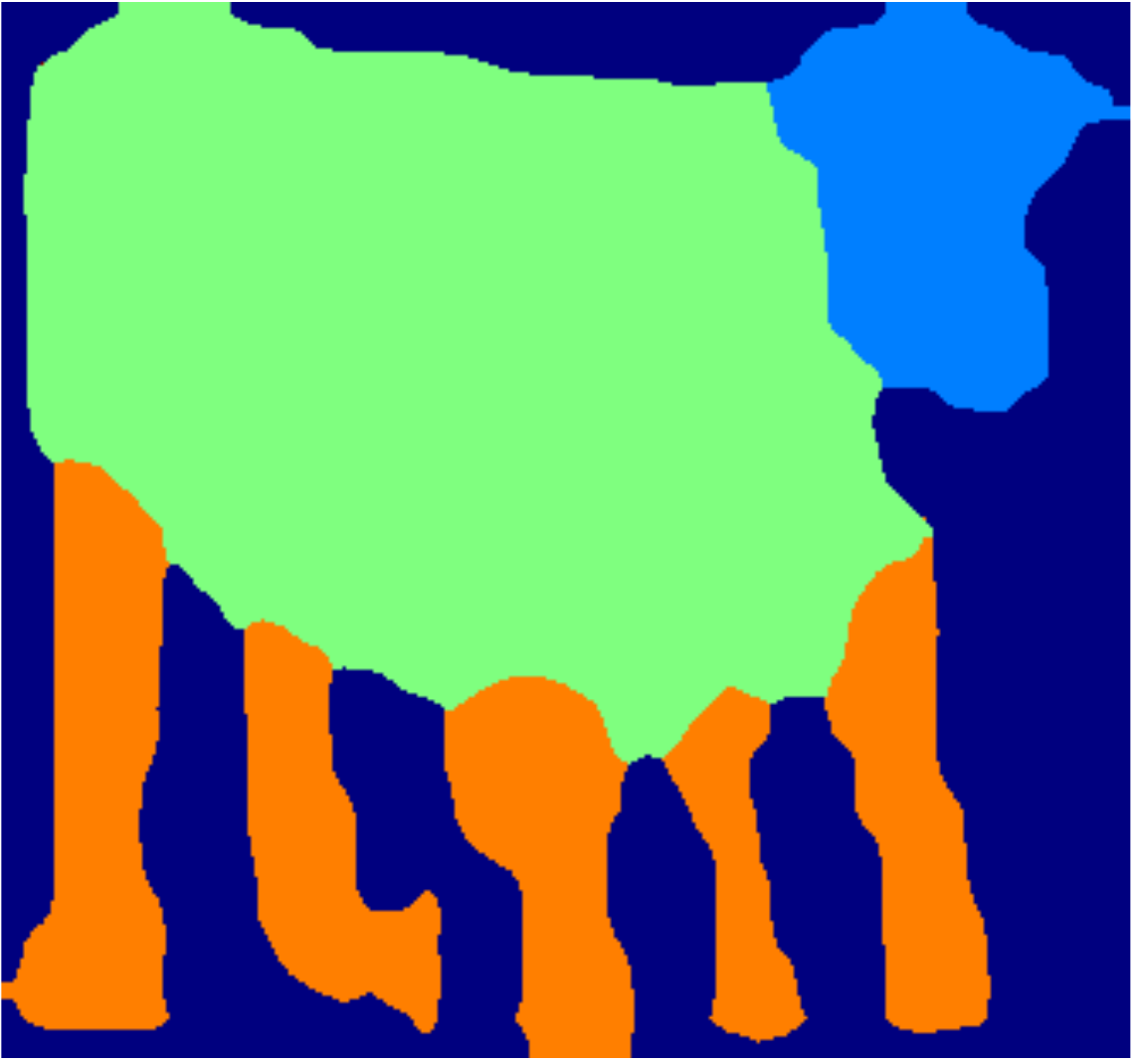}
\includegraphics[height=\hpascal, width=\wpascal\textwidth]{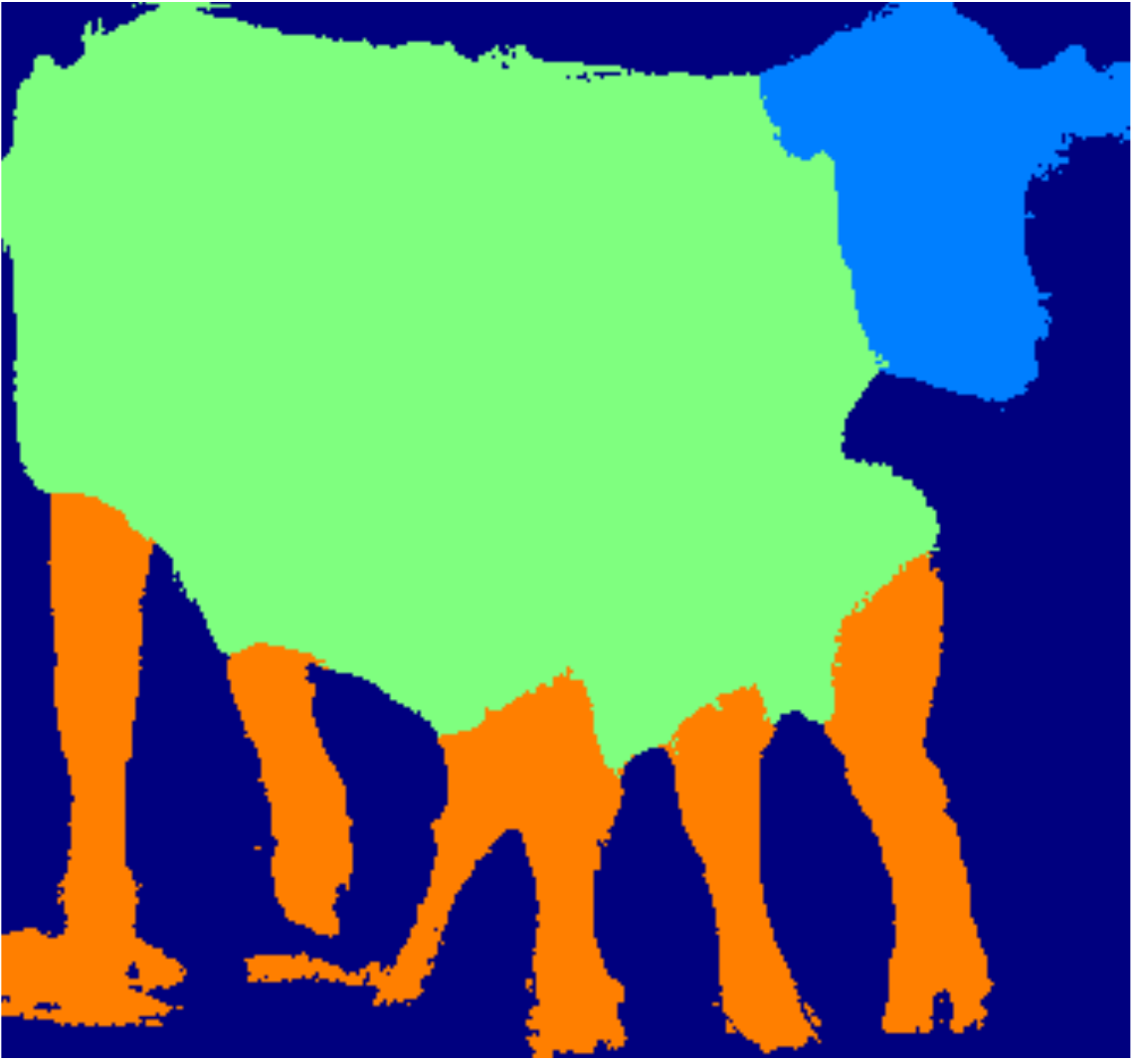}
\includegraphics[height=\hpascal, width=\wpascal\textwidth]{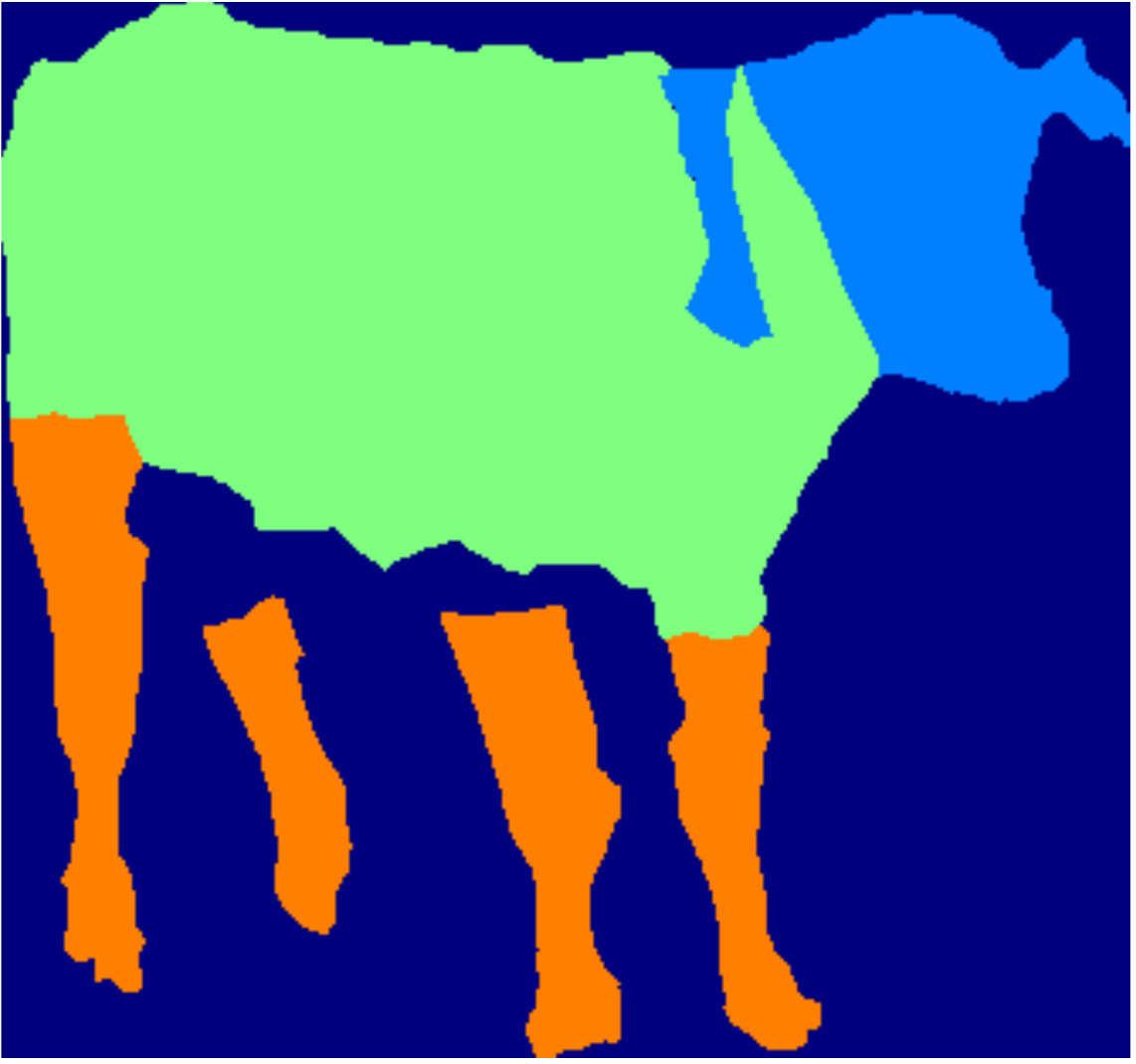}


\includegraphics[height=\pennheight, width=\pennwidth]{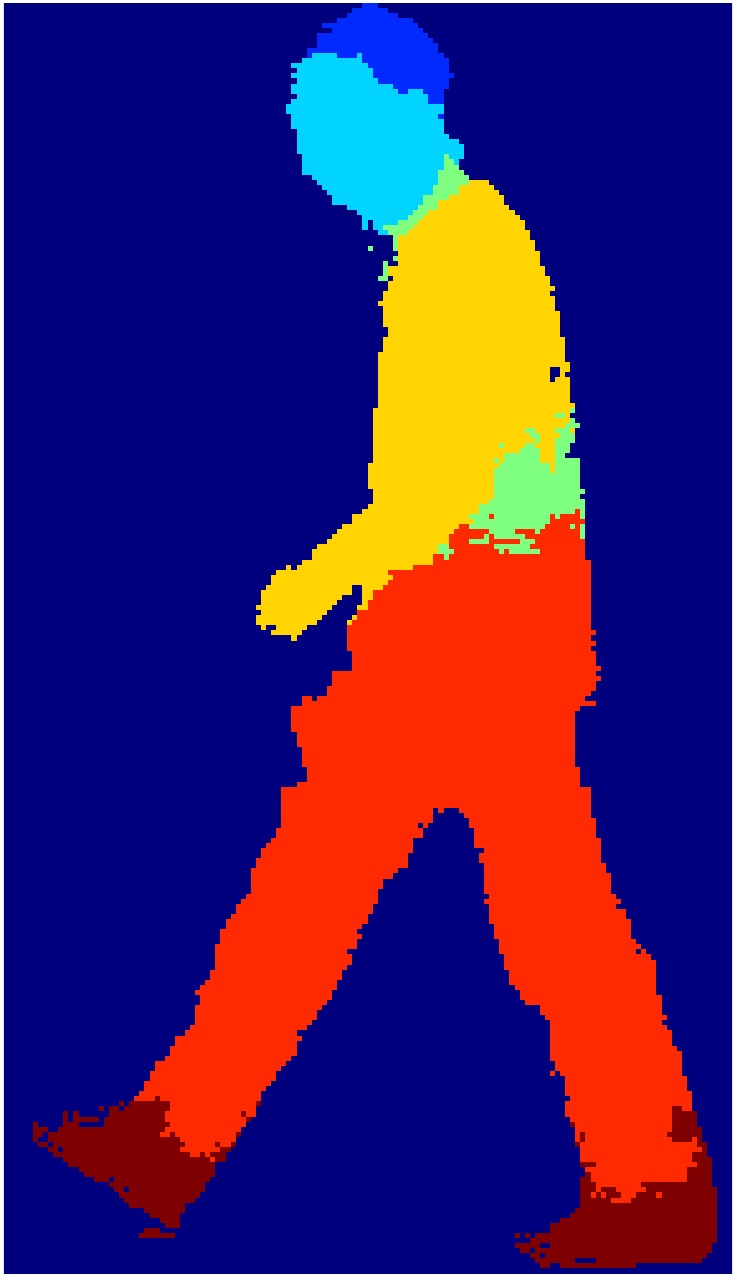}
\includegraphics[height=\pennheight, width=\pennwidth]{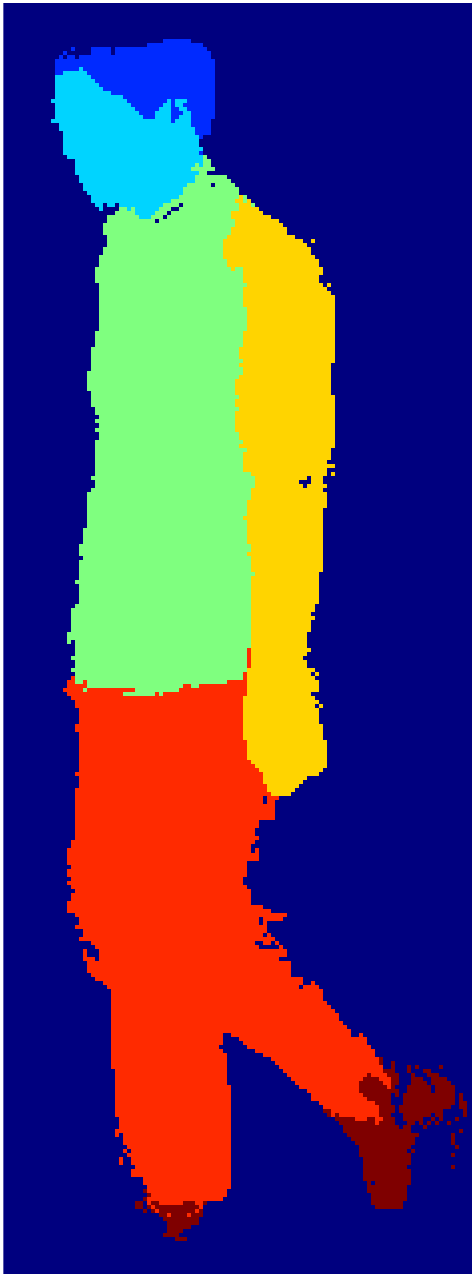}
\includegraphics[height=\pennheight, width=\pennwidth]{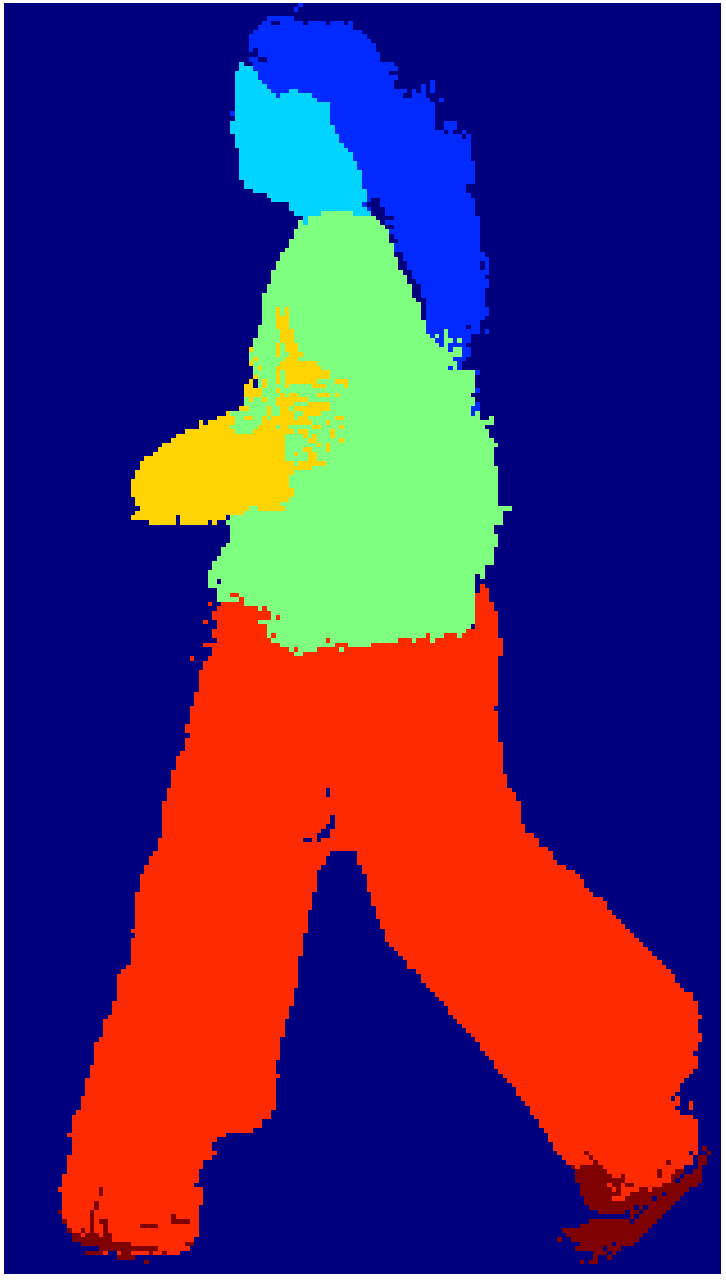}
\includegraphics[height=\pennheight, width=\pennwidth]{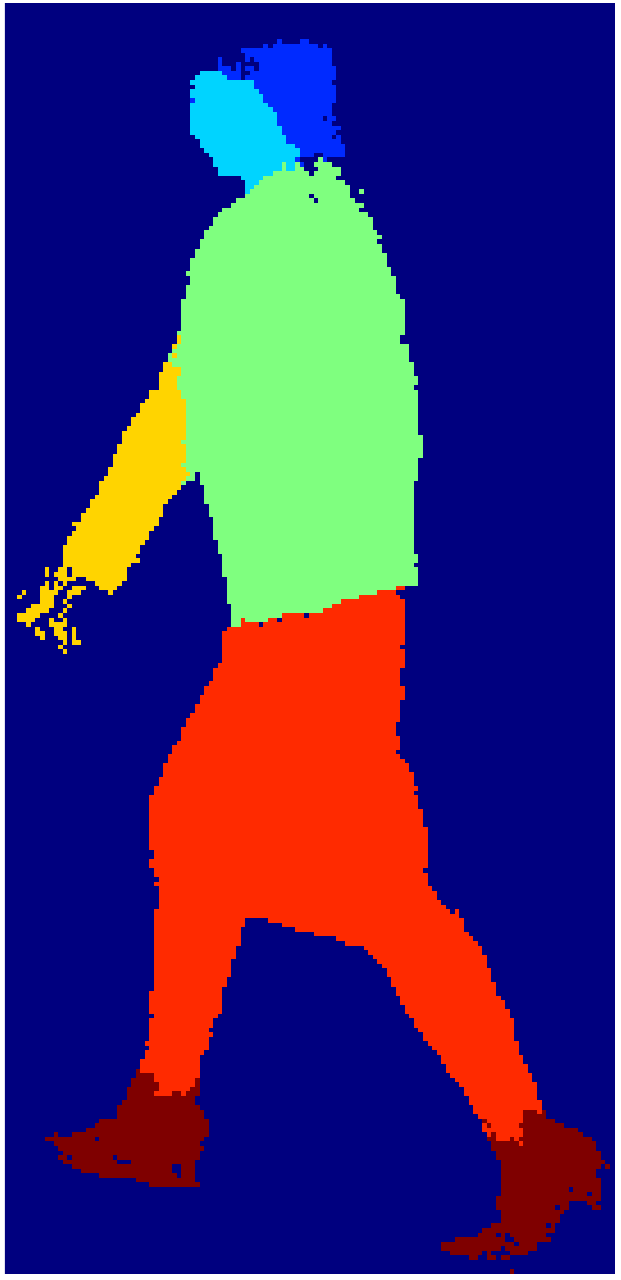}
\includegraphics[height=\pennheight, width=\pennwidth]{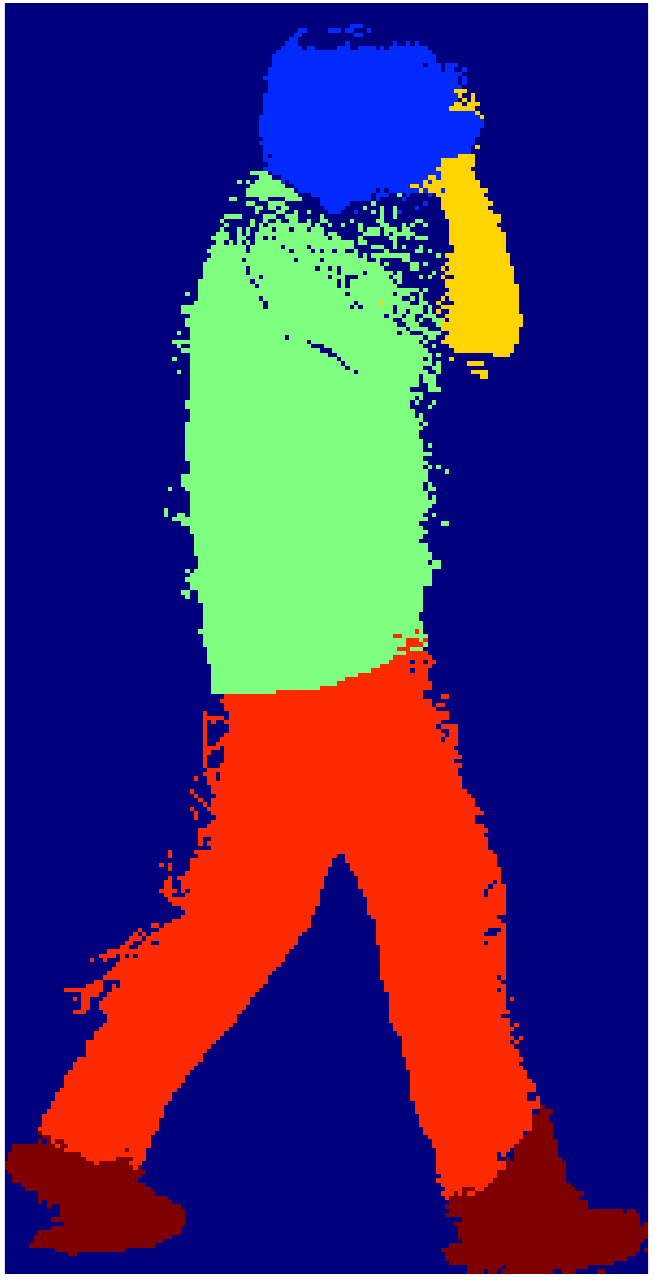}
\includegraphics[height=\pennheight, width=\pennwidth]{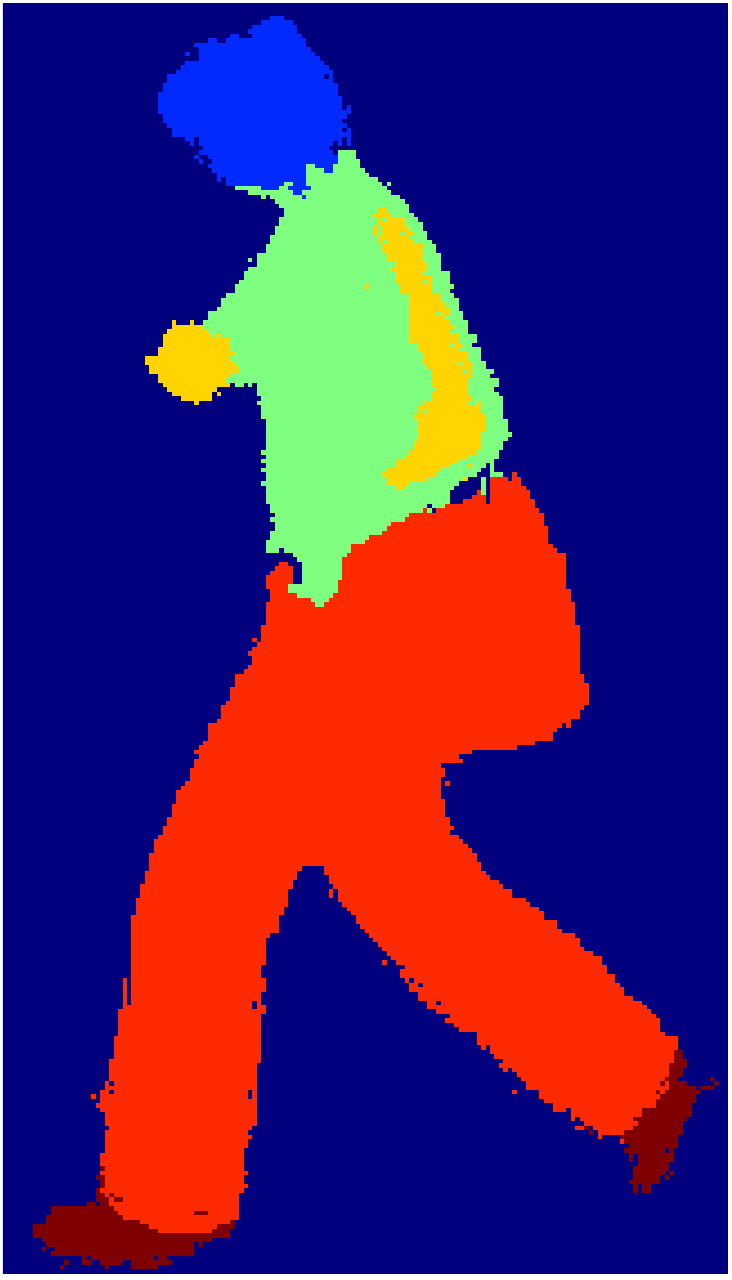}
\includegraphics[height=\pennheight, width=\pennwidth]{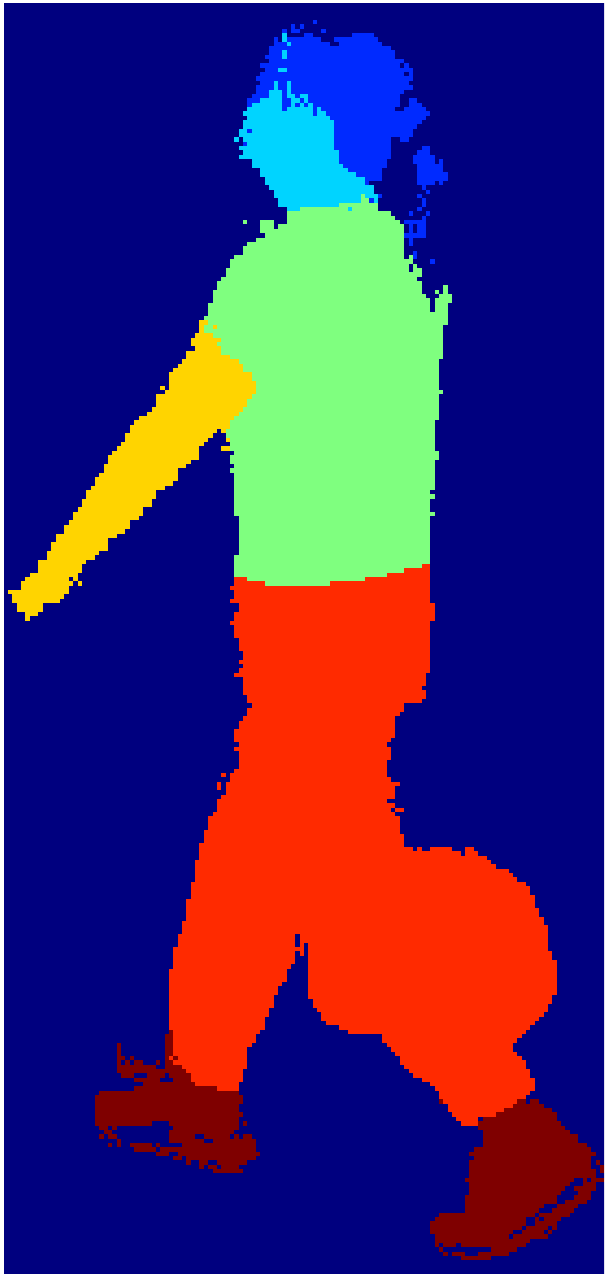}
\includegraphics[height=\pennheight, width=\pennwidth]{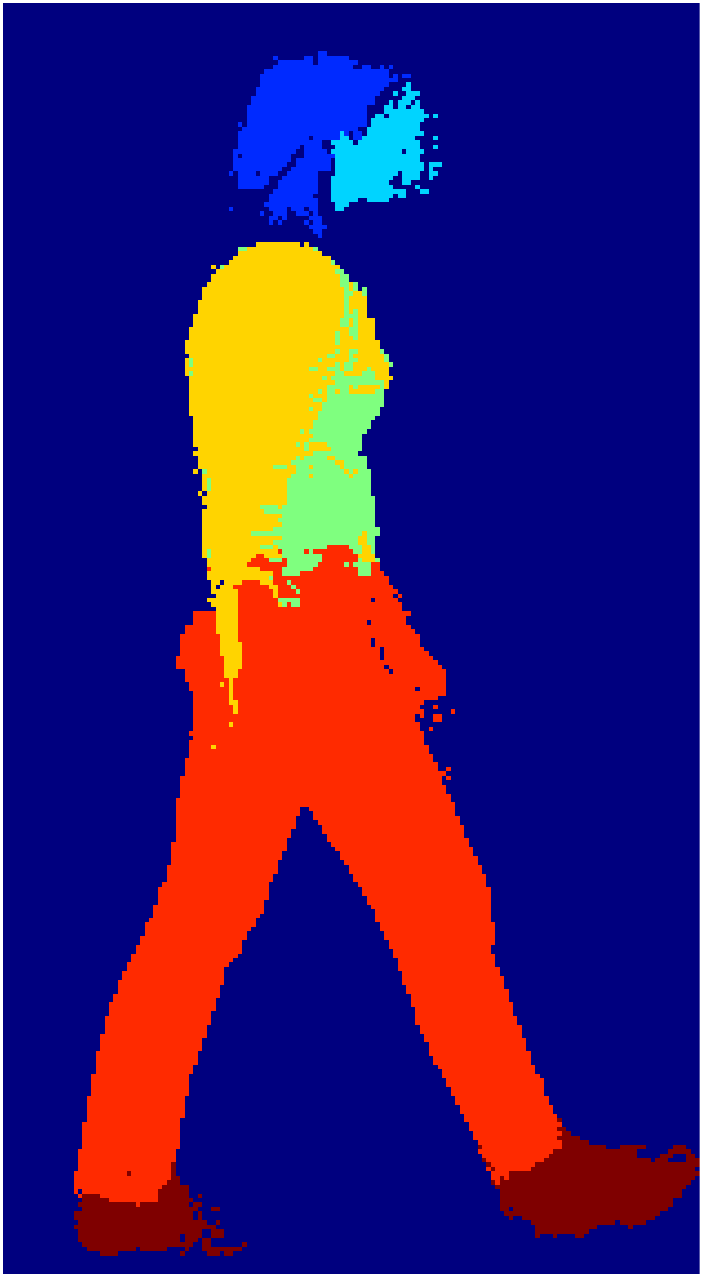}\,
\includegraphics[height=\hpascal, width=\wpascal\textwidth]{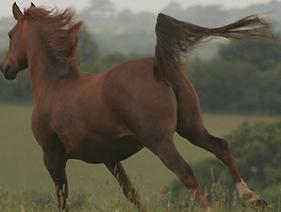}
\includegraphics[height=\hpascal, width=\wpascal\textwidth]{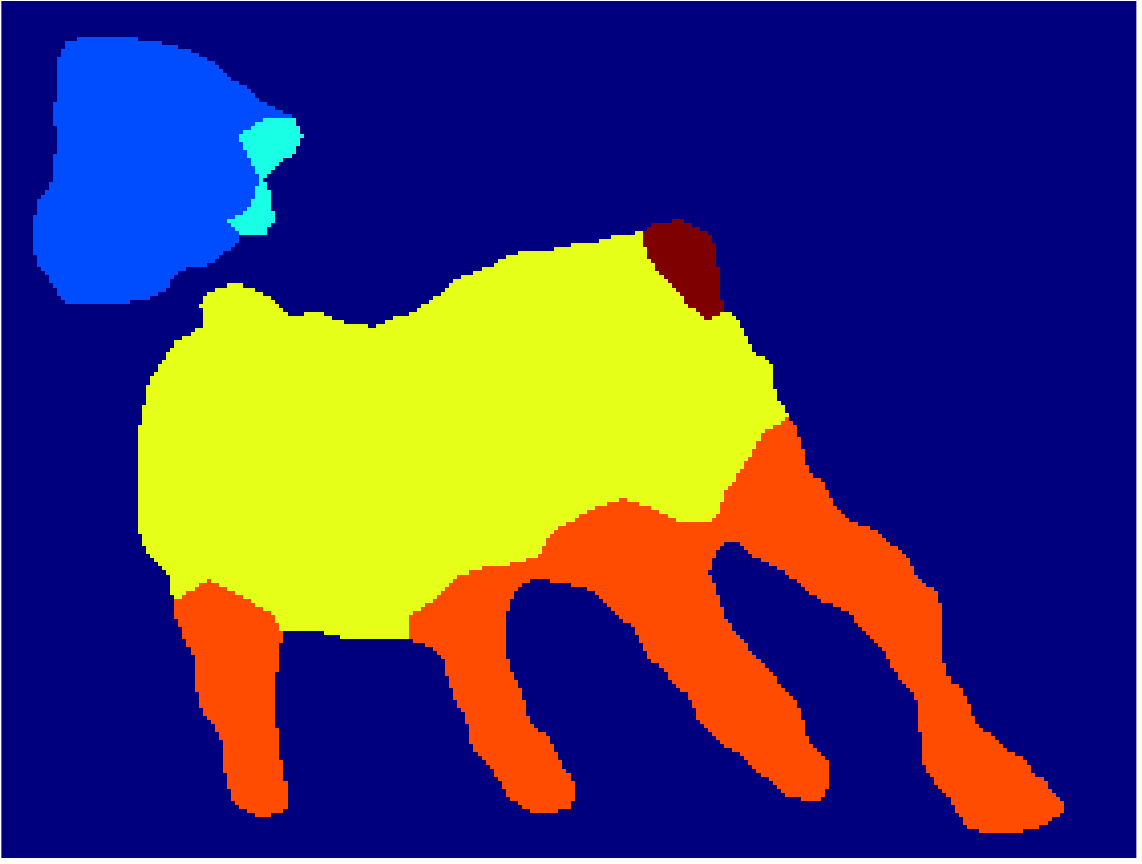}
\includegraphics[height=\hpascal, width=\wpascal\textwidth]{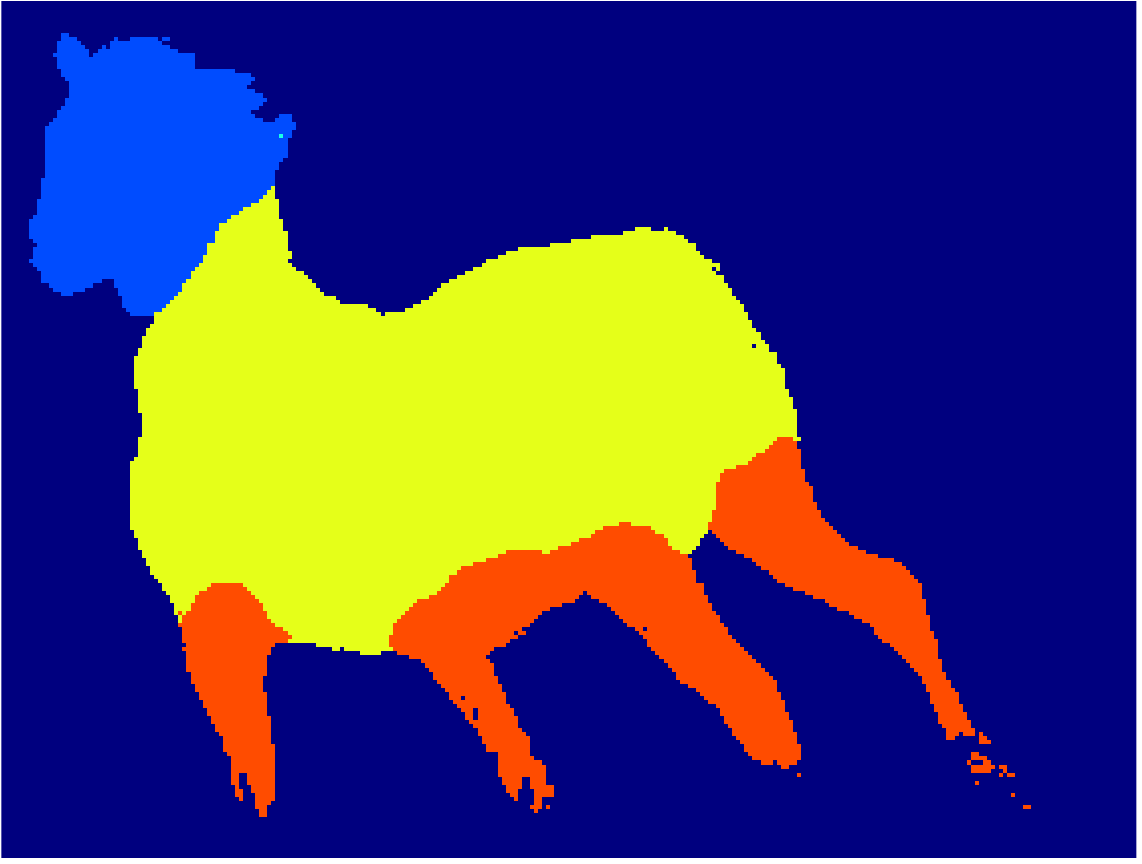}
\includegraphics[height=\hpascal, width=\wpascal\textwidth]{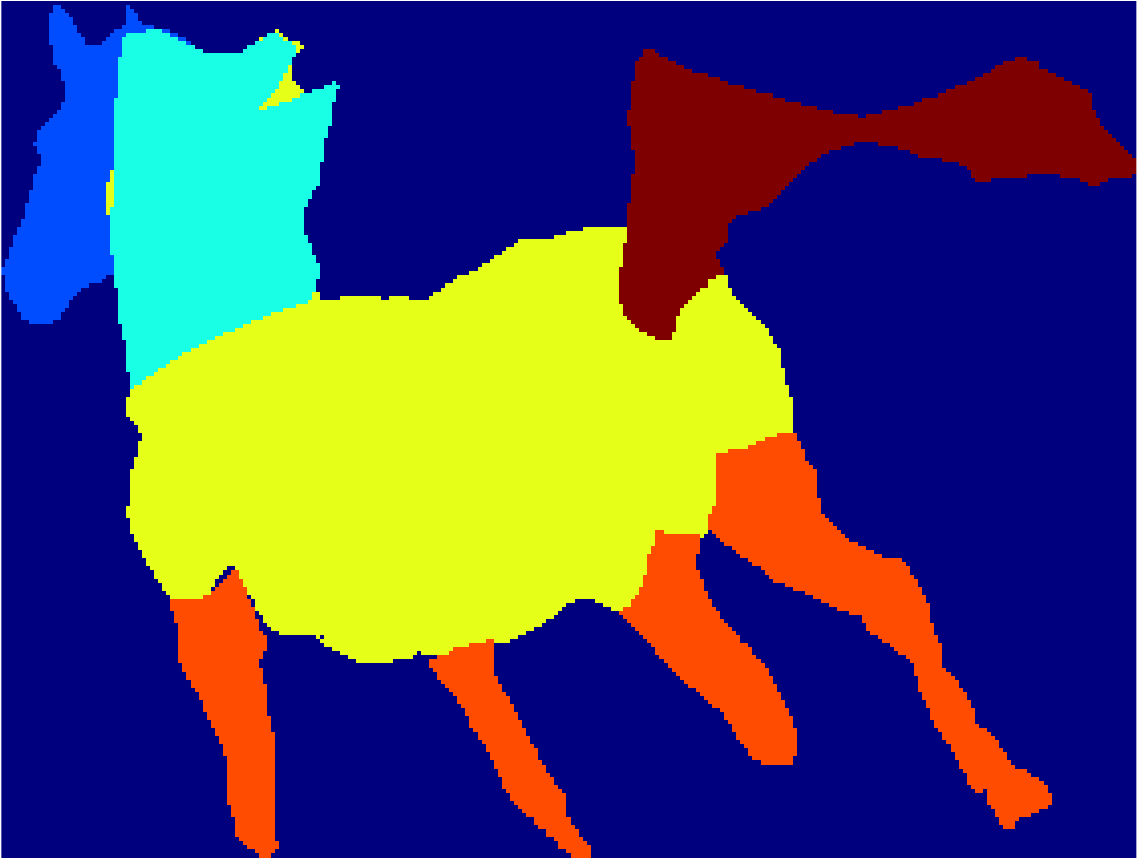}


\includegraphics[height=\pennheight, width=\pennwidth]{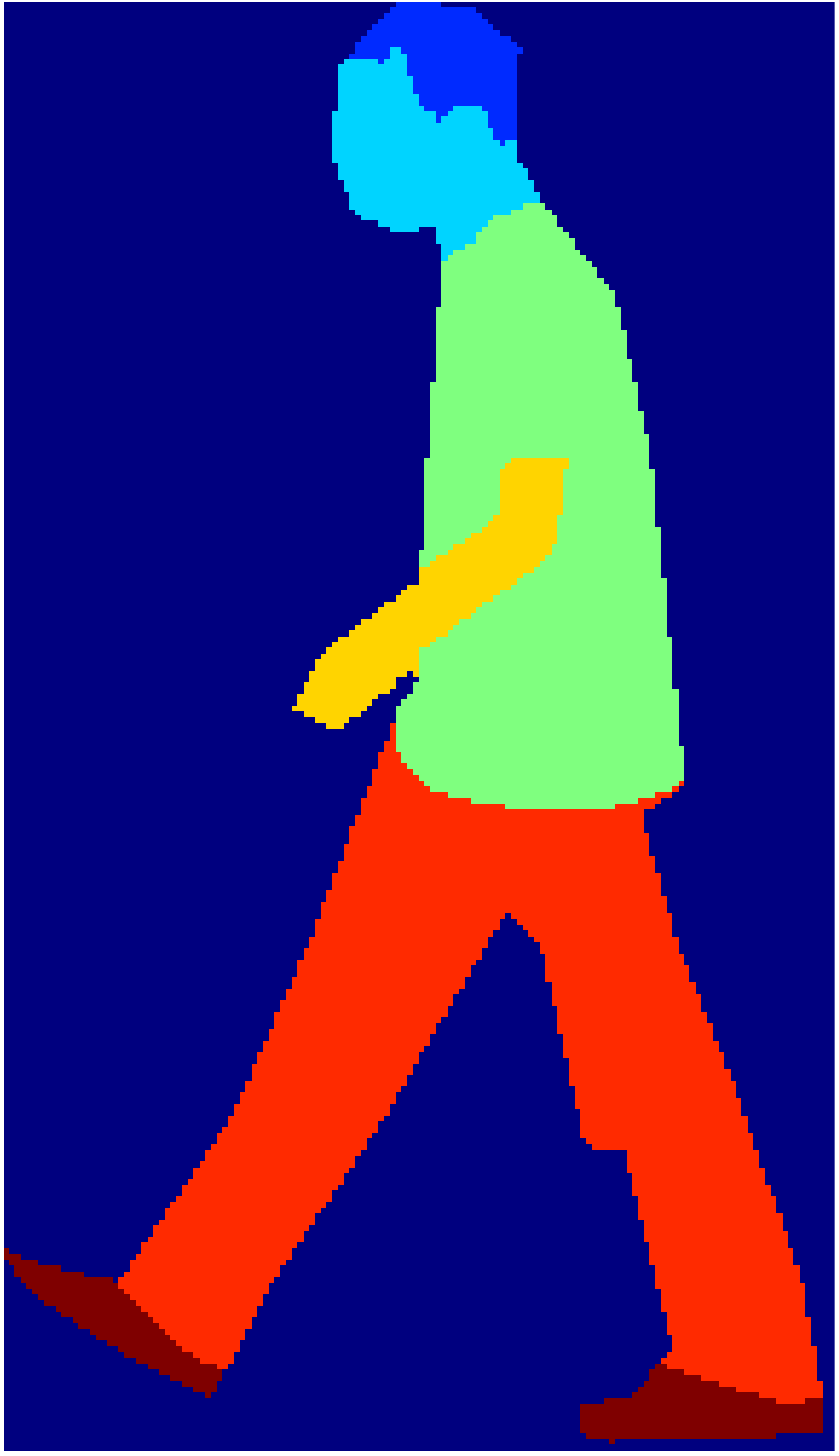}
\includegraphics[height=\pennheight, width=\pennwidth]{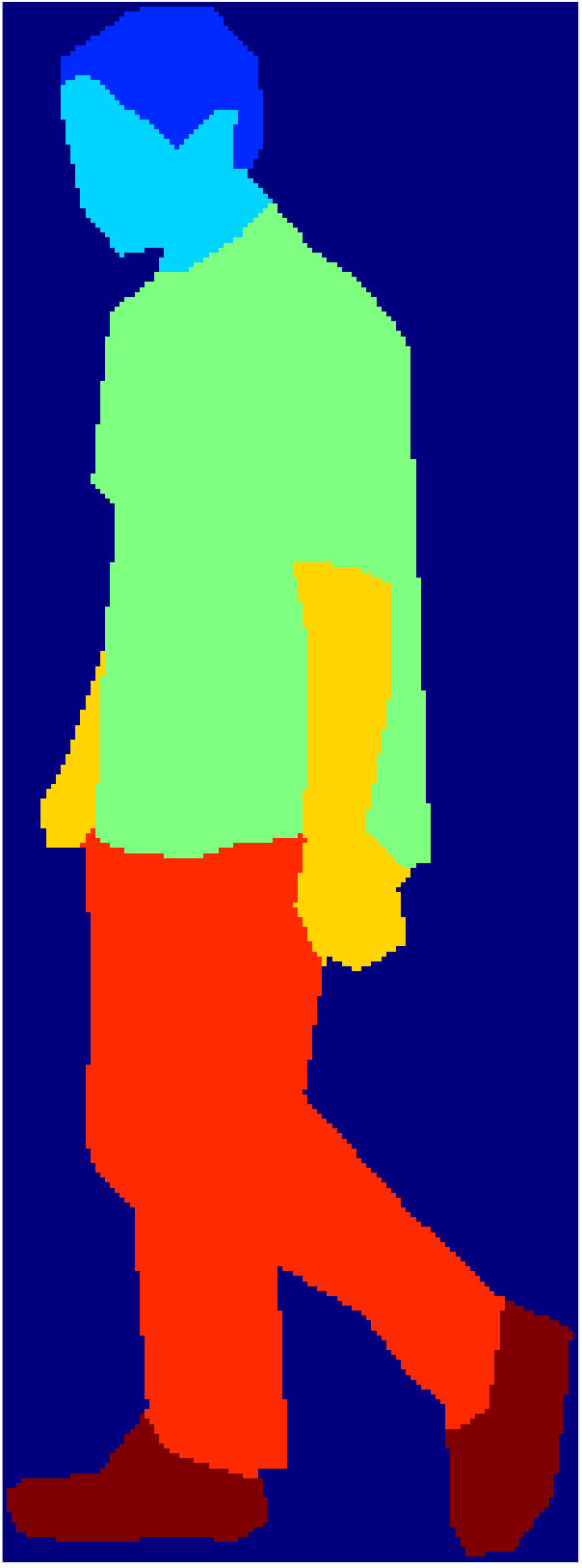}
\includegraphics[height=\pennheight, width=\pennwidth]{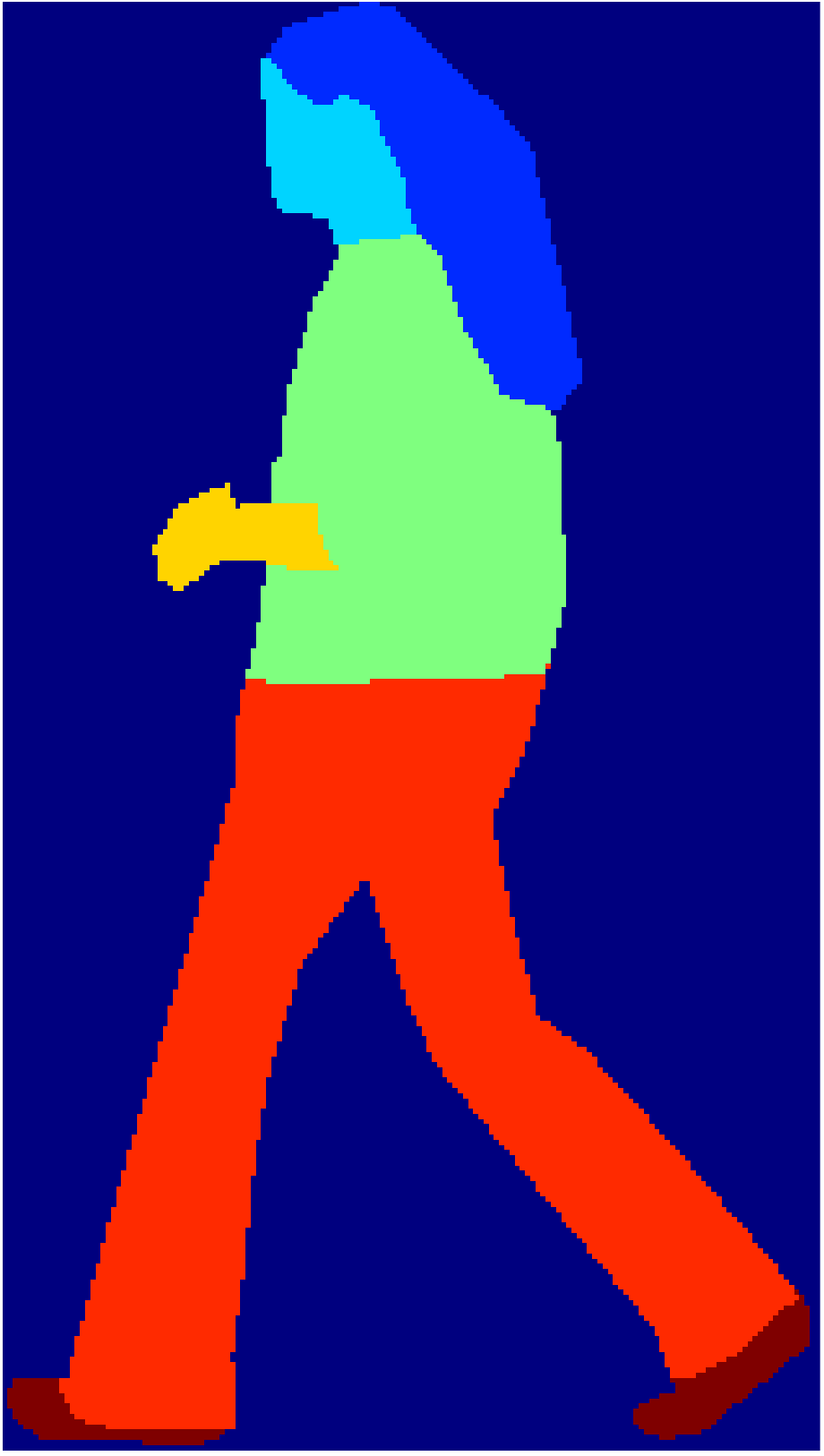}
\includegraphics[height=\pennheight, width=\pennwidth]{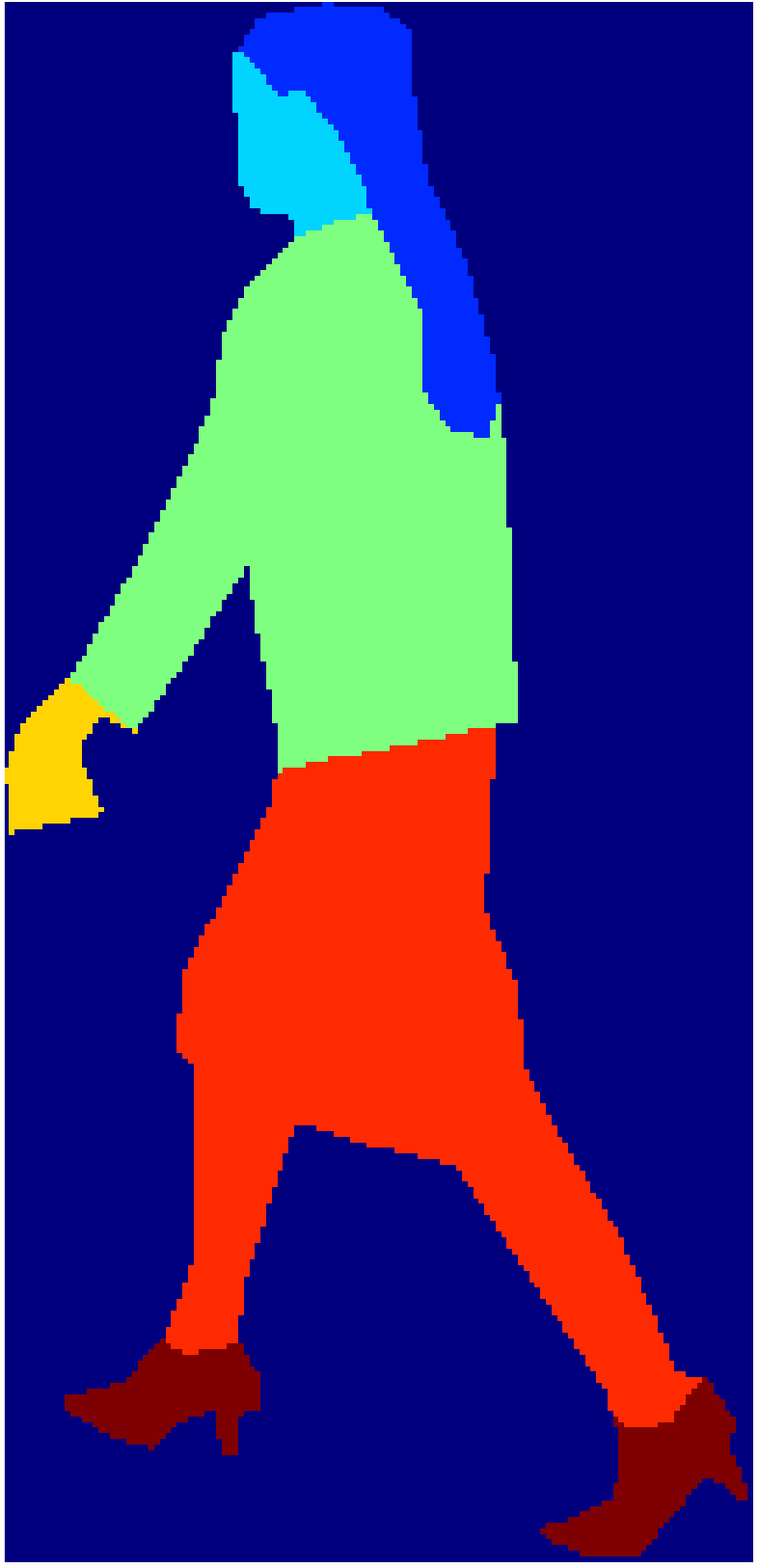}
\includegraphics[height=\pennheight, width=\pennwidth]{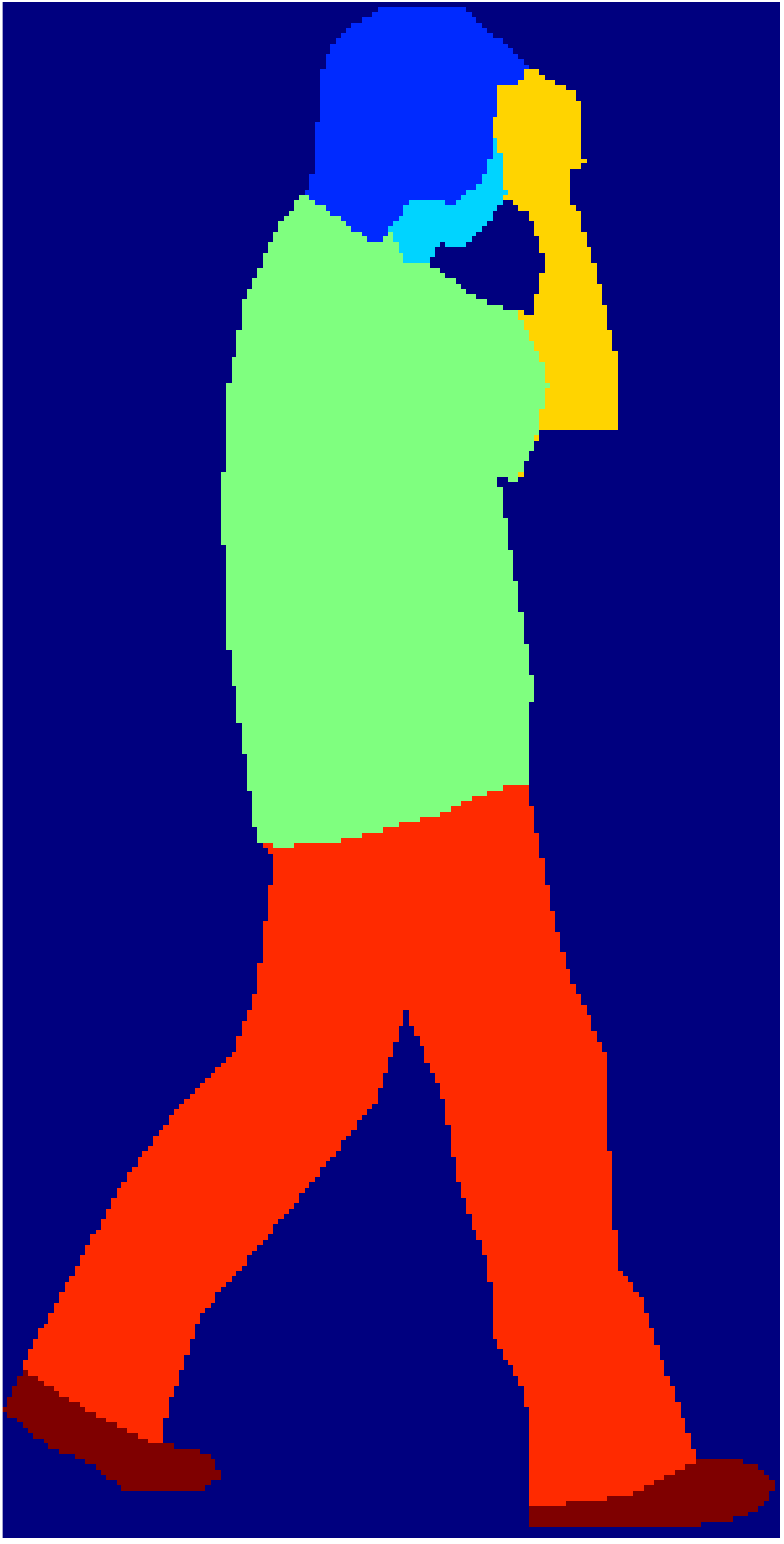}
\includegraphics[height=\pennheight, width=\pennwidth]{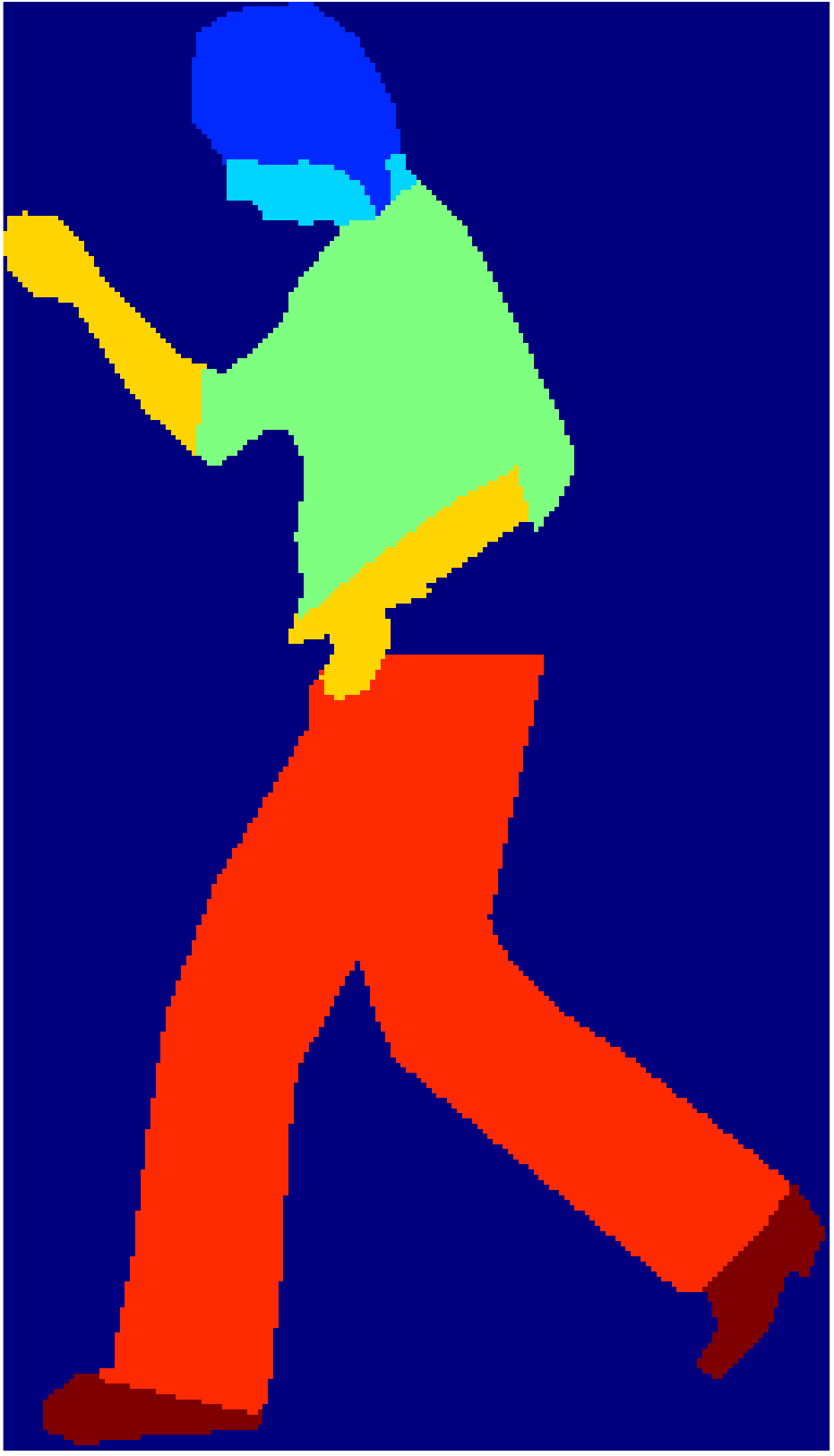}
\includegraphics[height=\pennheight, width=\pennwidth]{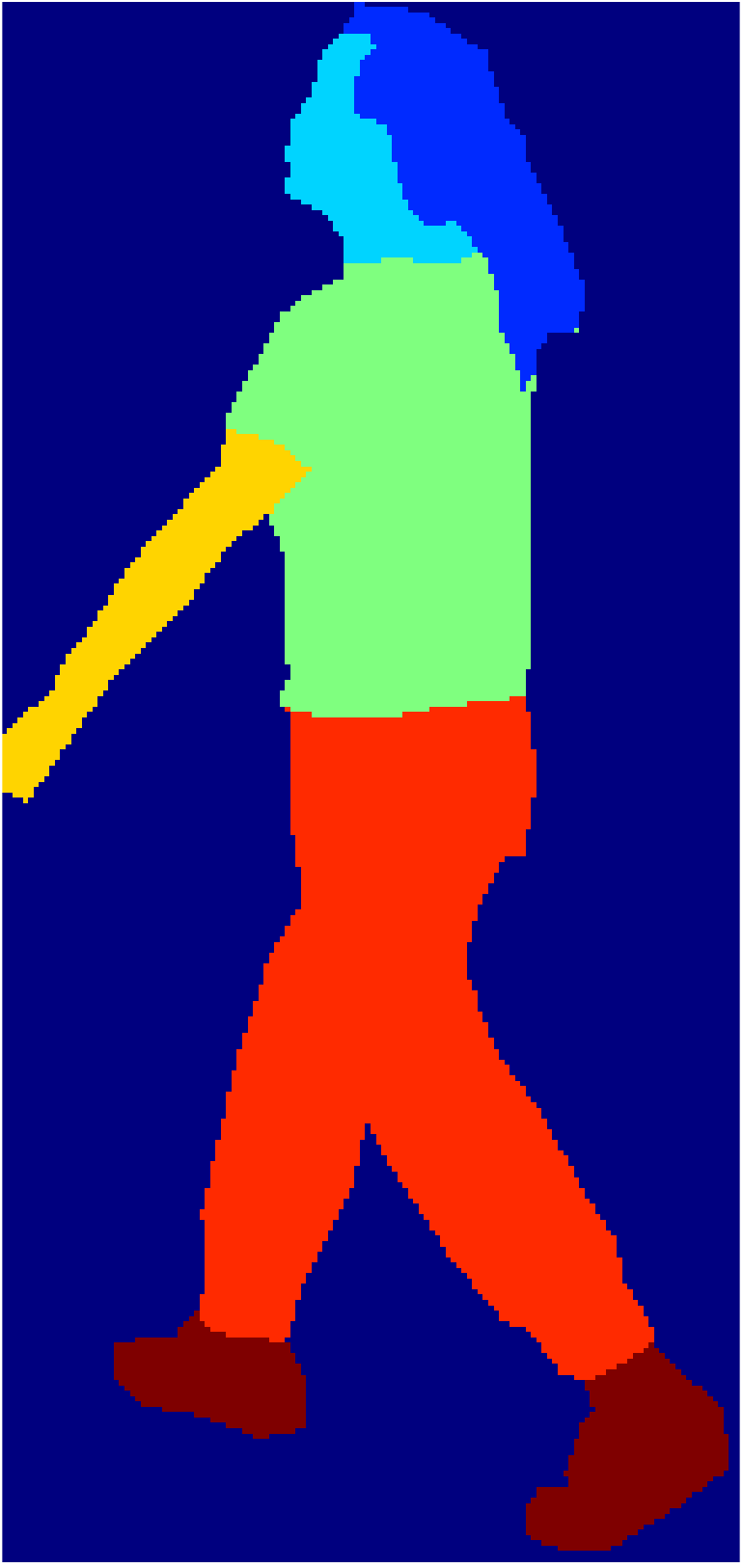}
\includegraphics[height=\pennheight, width=\pennwidth]{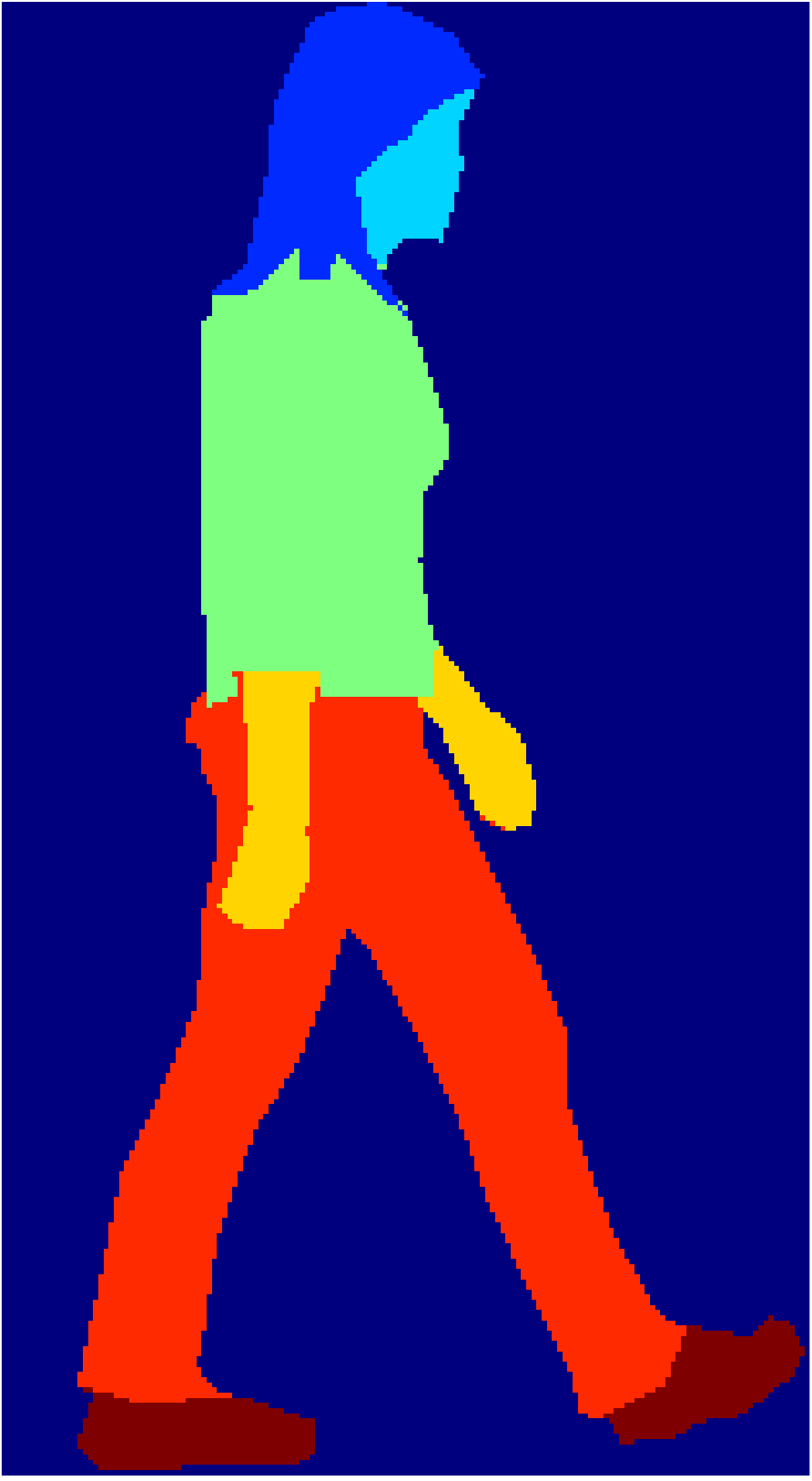}\,
\includegraphics[height=\hpascal, width=\wpascal\textwidth]{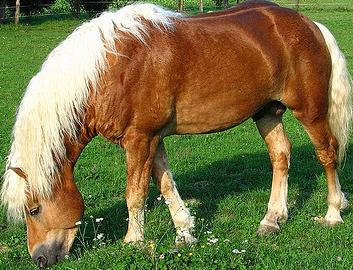}
\includegraphics[height=\hpascal, width=\wpascal\textwidth]{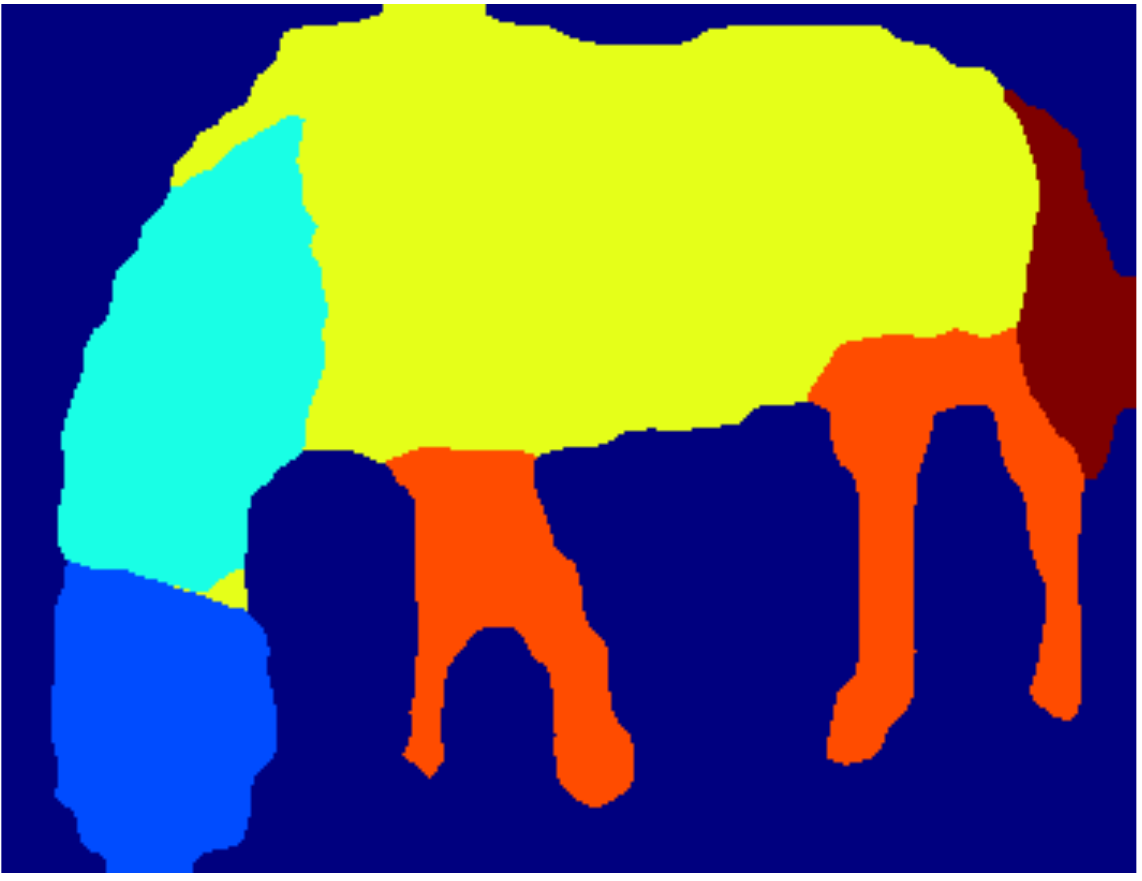}
\includegraphics[height=\hpascal, width=\wpascal\textwidth]{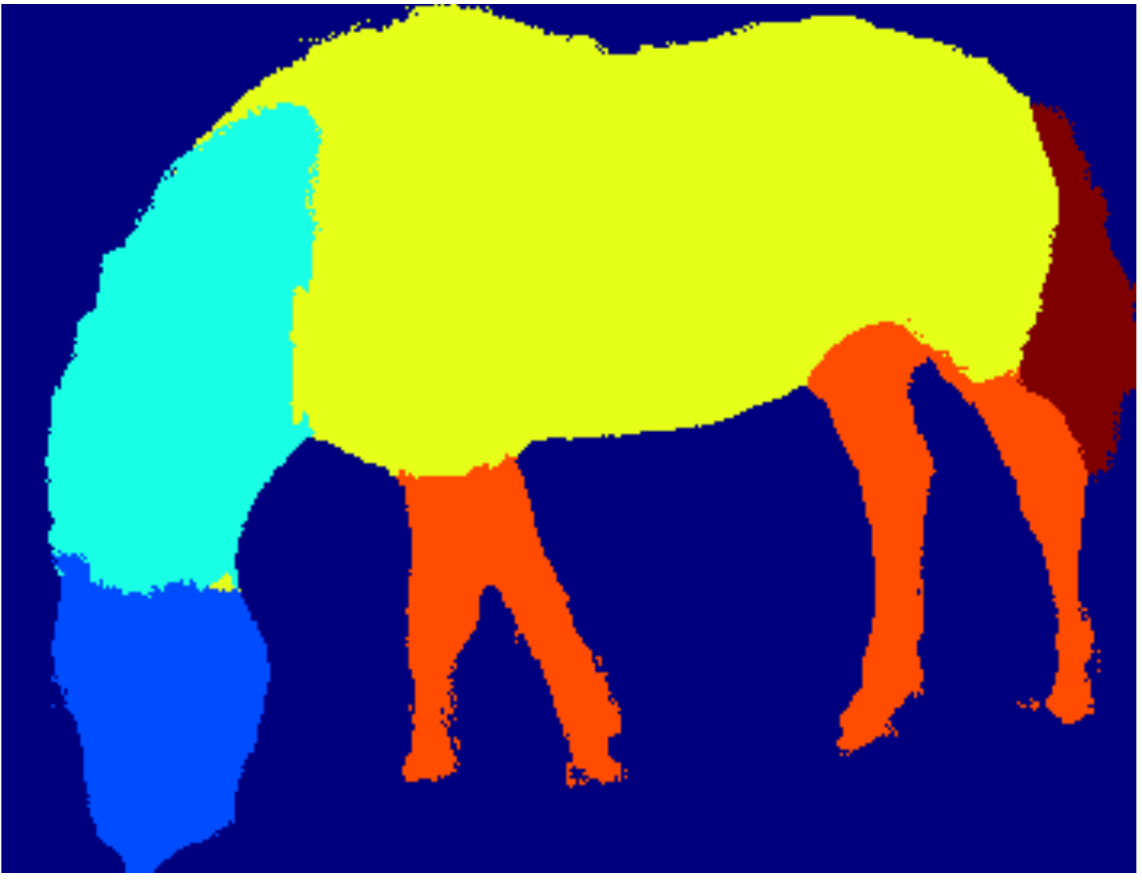}
\includegraphics[height=\hpascal, width=\wpascal\textwidth]{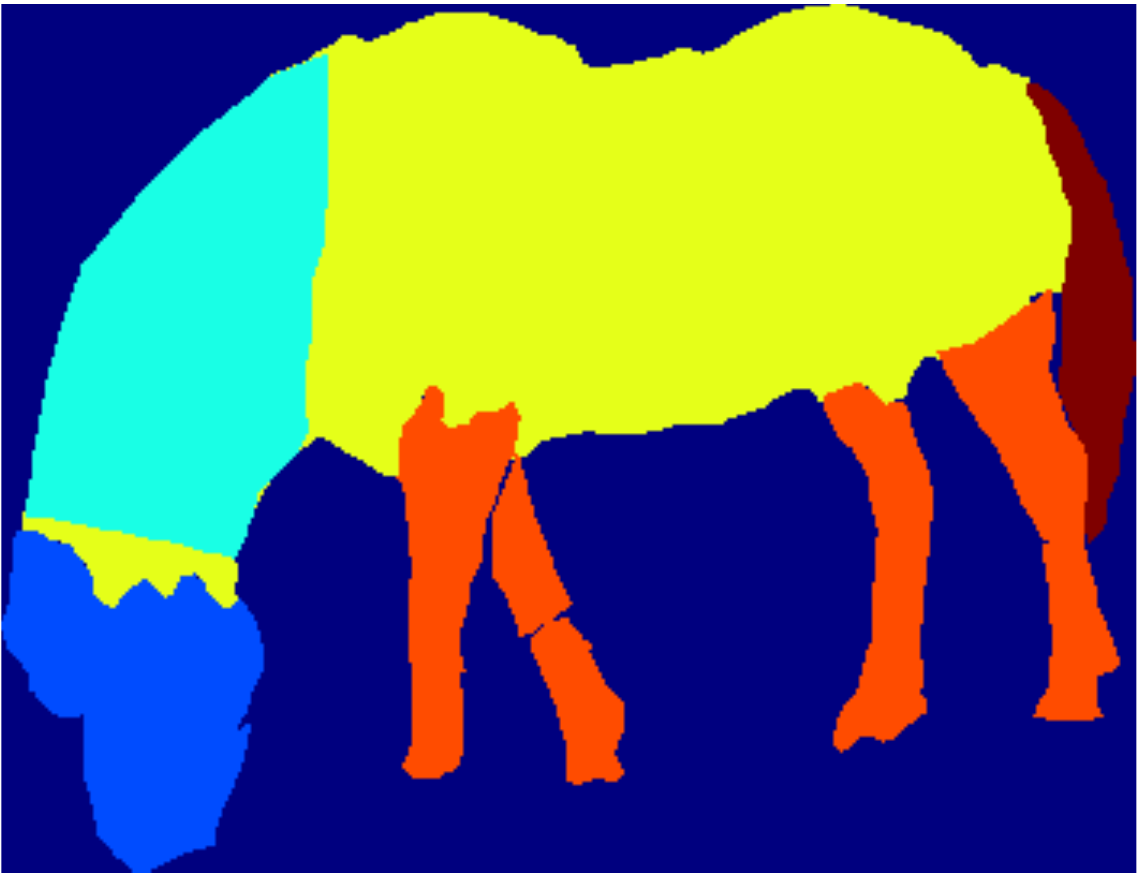}

\caption{\textbf{Left:} Pedestrian parsing results on Penn-Fudan dataset. From top to bottom: a) Input image, b) SBP~\cite{bo2011shape}, c) Raw CNN scores, d) CNN+CRF, e) Groundtruth. \textbf{Right:} Part segmentation results for car, horse and cow on the PASCAL-Parts dataset. From left to right: a) Input image, b) Masks from raw CNN scores, c) Masks from CNN+CRF, d) Groundtruth. Best seen in color. } 

\label{fig:pedestrian}
\end{figure*}

\subsubsection*{Labeled Faces in the Wild}
\label{sec:lfw}
Labeled Faces in the Wild (LFW) is a dataset containing more than 13000 images of faces collected from  the web. For our purposes, we used the ``funneled'' version of the dataset, in which images have been coarsely aligned using a congealing-style
joint alignment approach~\cite{huang2007unsupervised}. This is the subset also used in~\cite{kae2013augmenting} and consists of 1500 train, 500 validation and 927 testing images of faces, and their corresponding superpixel segmentations, with labels for background, hair (including facial hair) and face. We train our DCNN on the 2000 trainval images and evaluate on the 927 test images, using superpixel accuracy as in~\cite{kae2013augmenting} for the purpose of comparison. Since our system returns pixelwise labels for each image, we employ a simple scheme to obtain superpixel labels: for each superpixel we compute a histogram of the pixel labels it contains and choose the most frequent label as the superpixel label. 

\begin{figure*}[!t]
\centering
\includegraphics[width=0.06\textwidth]{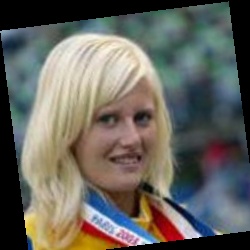}
\includegraphics[width=0.06\textwidth]{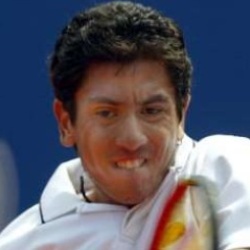}
\includegraphics[width=0.06\textwidth]{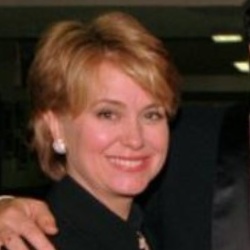}
\includegraphics[width=0.06\textwidth]{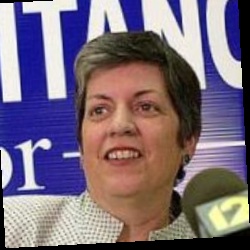}
\includegraphics[width=0.06\textwidth]{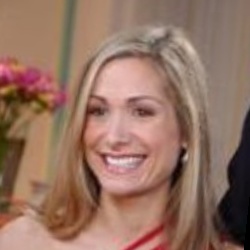}
\includegraphics[width=0.06\textwidth]{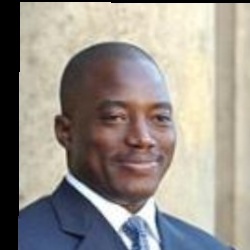}
\includegraphics[width=0.06\textwidth]{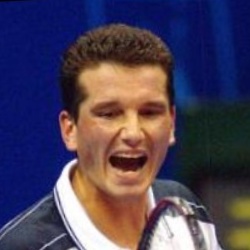}
\includegraphics[width=0.06\textwidth]{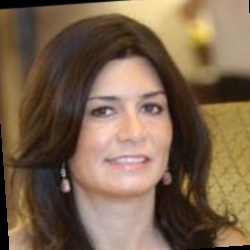}
\includegraphics[width=0.06\textwidth]{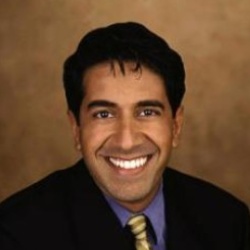}
\includegraphics[width=0.06\textwidth]{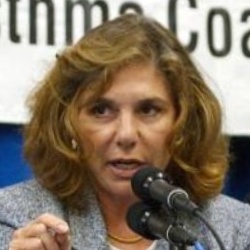}
\includegraphics[width=0.06\textwidth]{Janet_Napolitano_0002}
\includegraphics[width=0.06\textwidth]{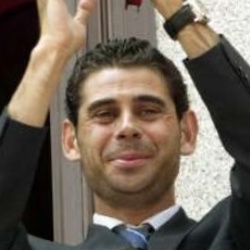}
\includegraphics[width=0.06\textwidth]{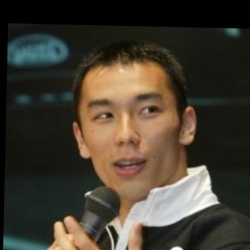}
\includegraphics[width=0.06\textwidth]{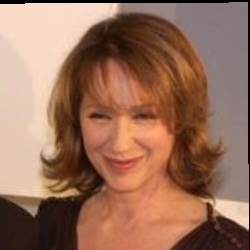}
\includegraphics[width=0.06\textwidth]{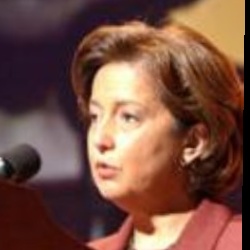}

\includegraphics[width=0.06\textwidth]{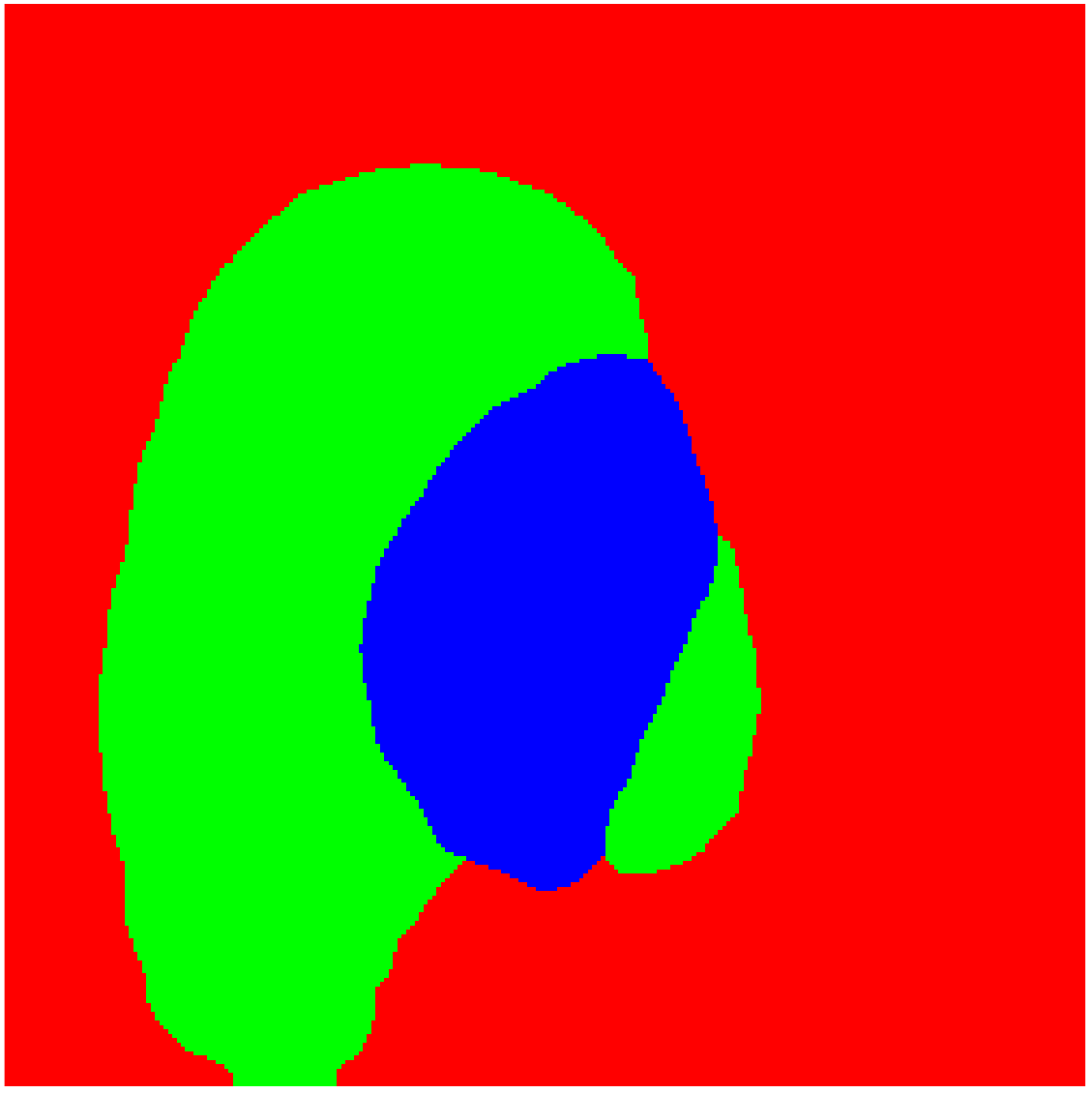}
\includegraphics[width=0.06\textwidth]{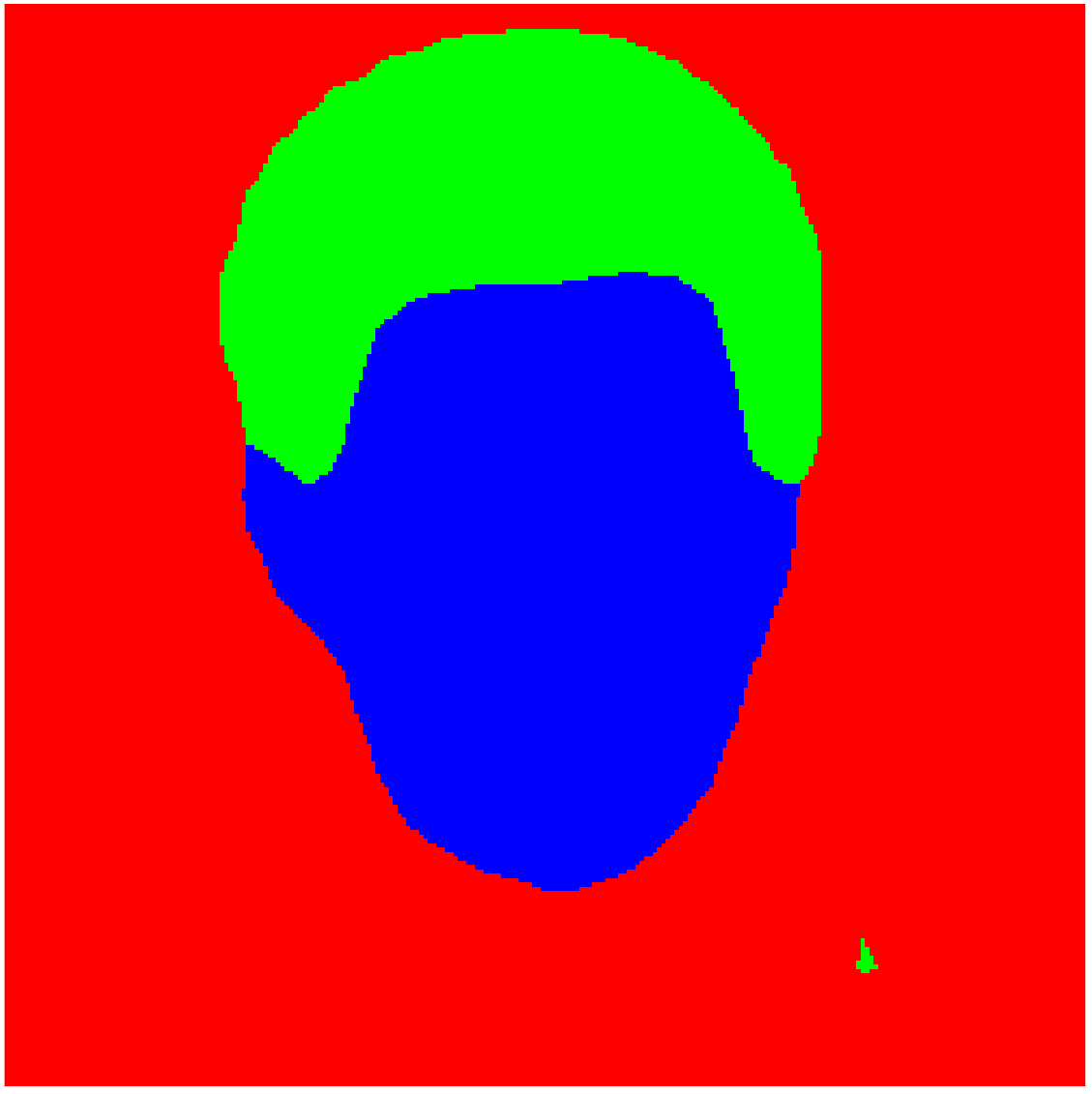}
\includegraphics[width=0.06\textwidth]{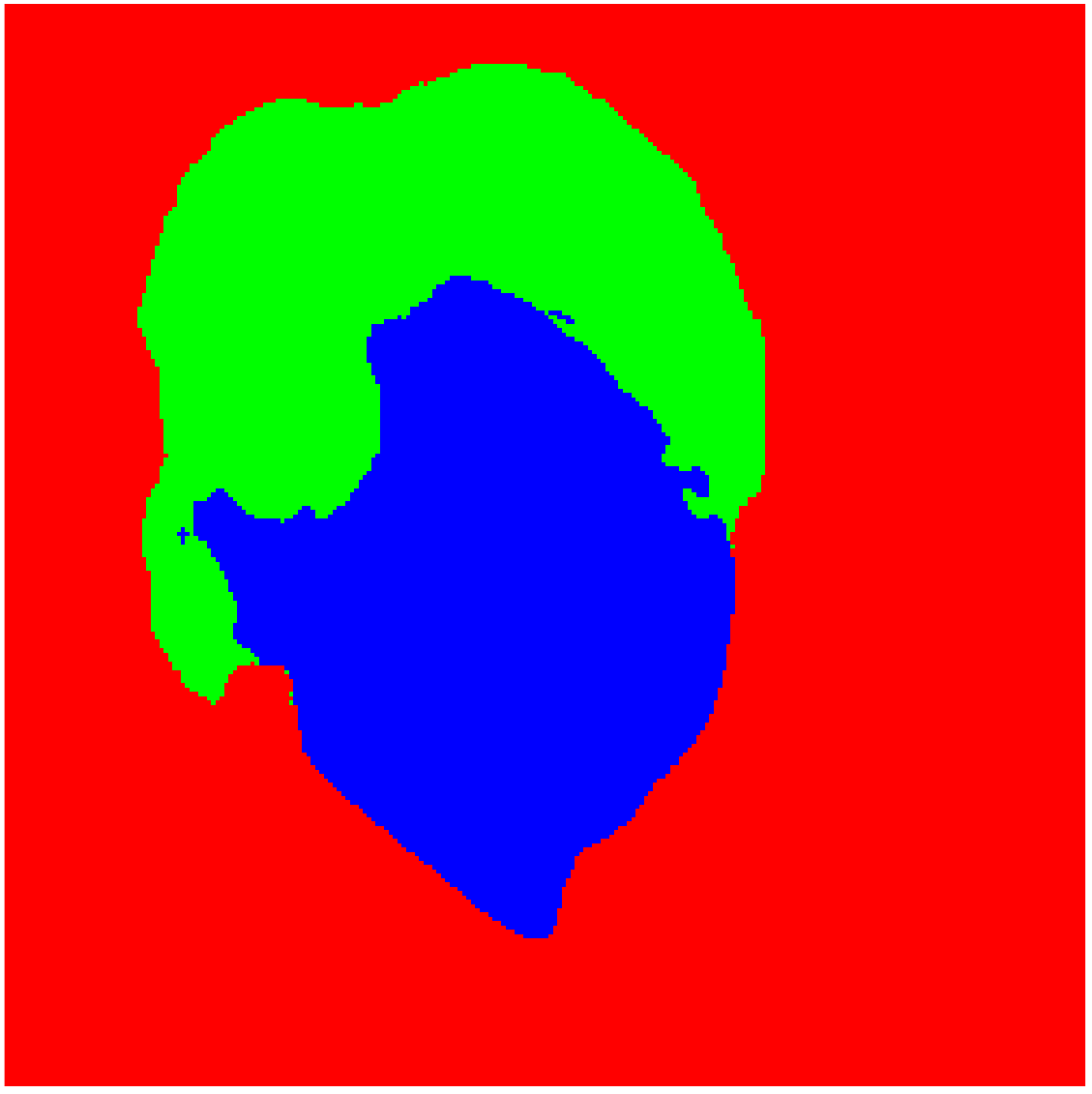}
\includegraphics[width=0.06\textwidth]{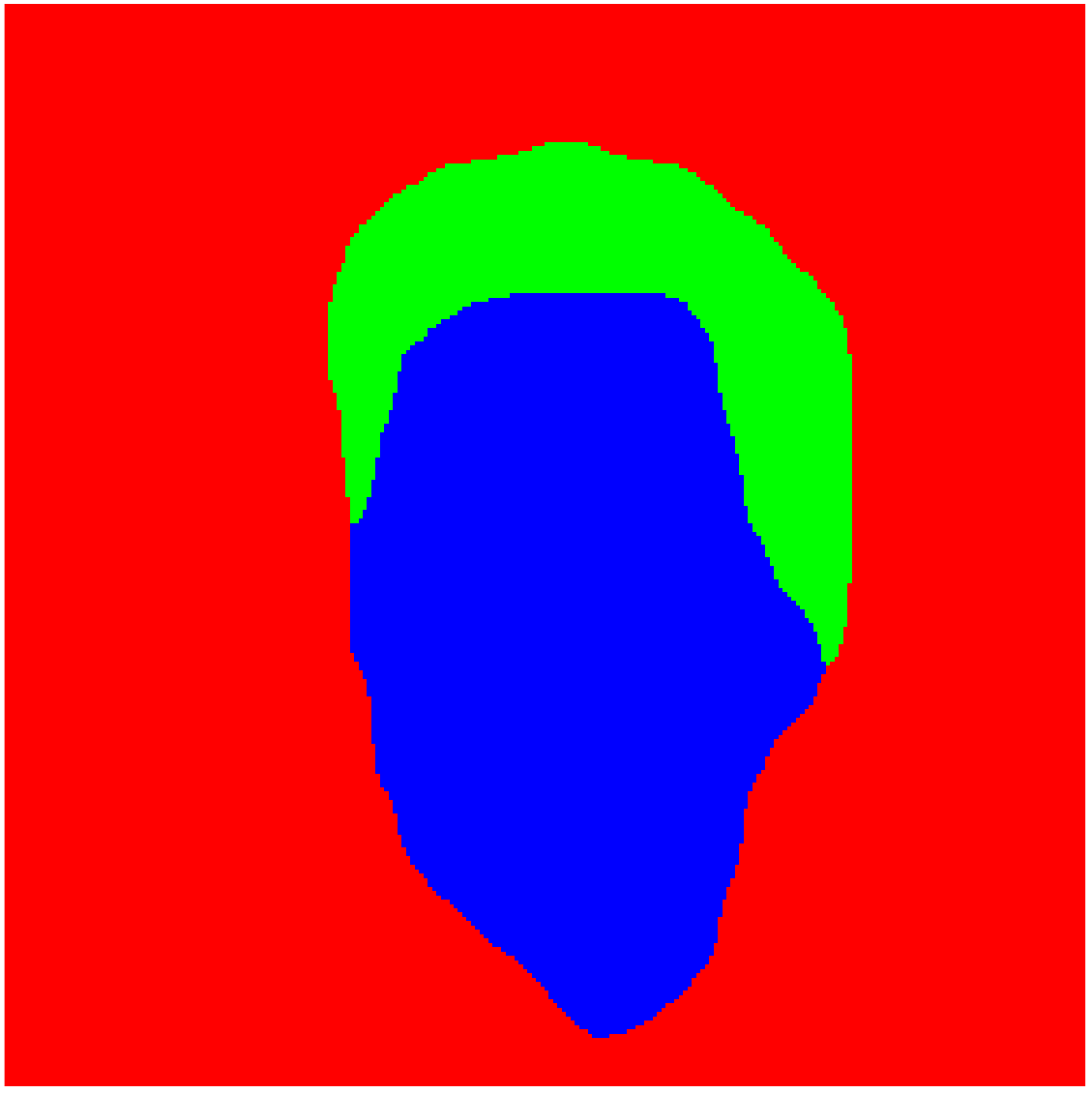}
\includegraphics[width=0.06\textwidth]{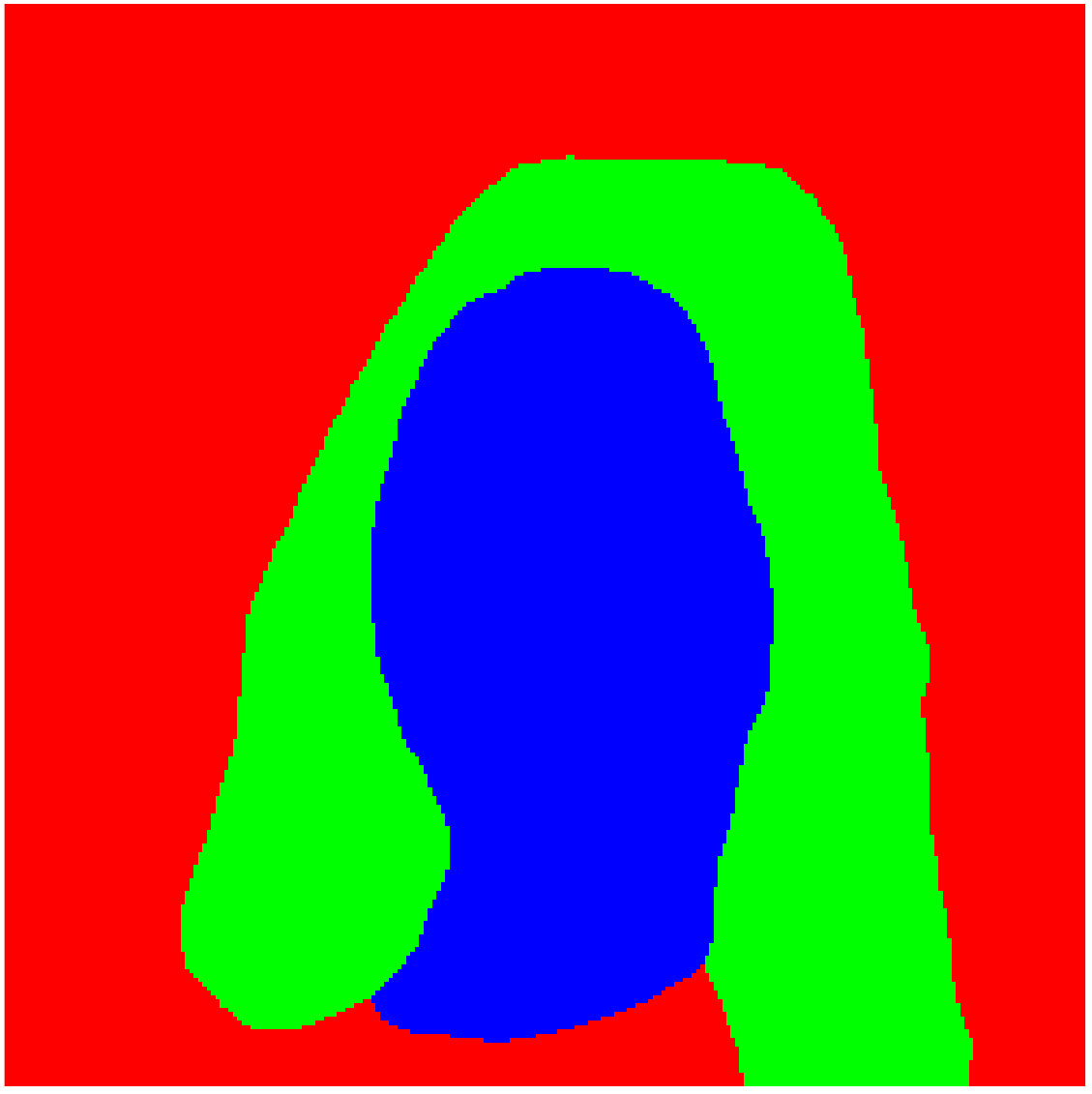}
\includegraphics[width=0.06\textwidth]{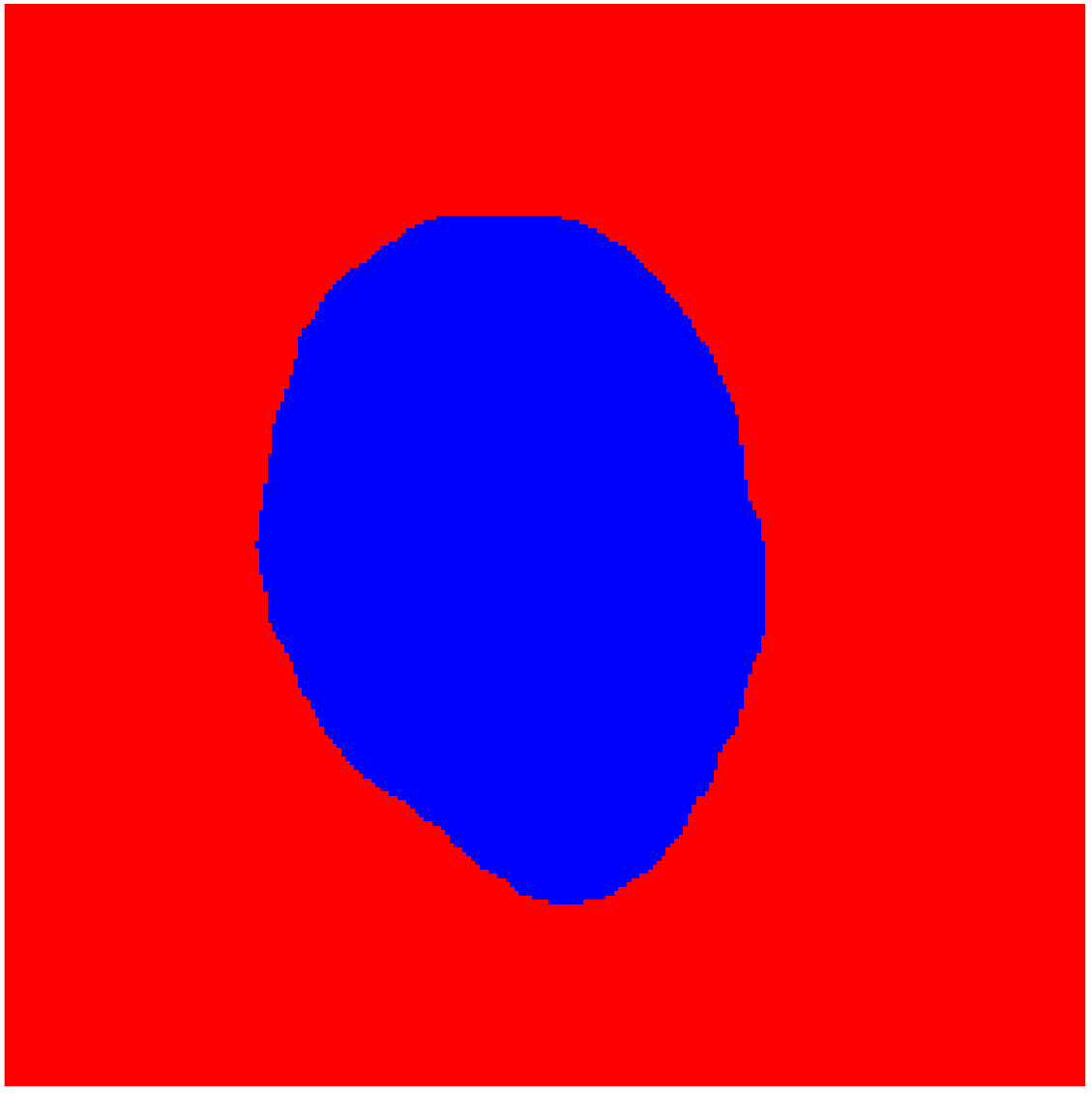}
\includegraphics[width=0.06\textwidth]{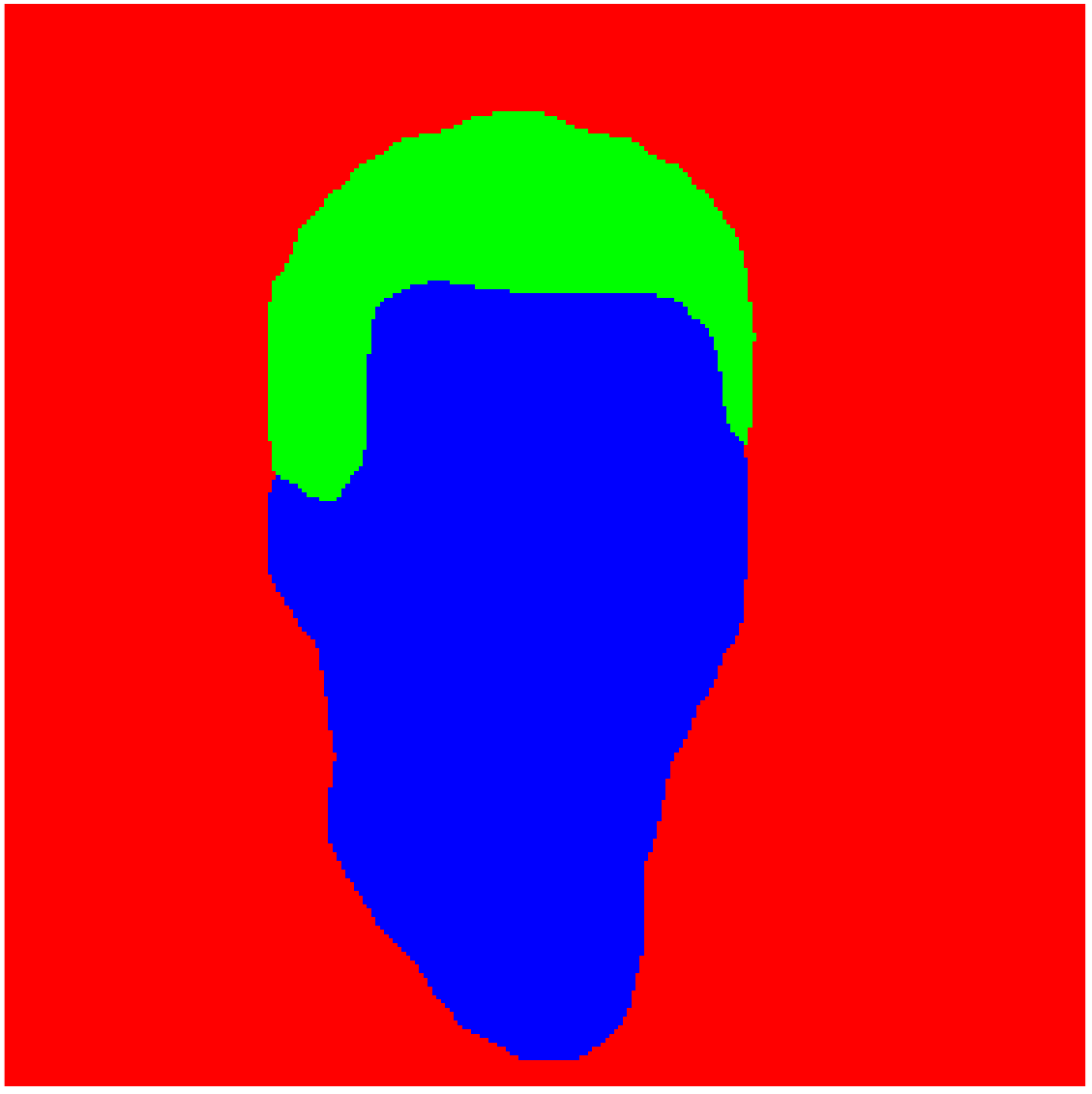}
\includegraphics[width=0.06\textwidth]{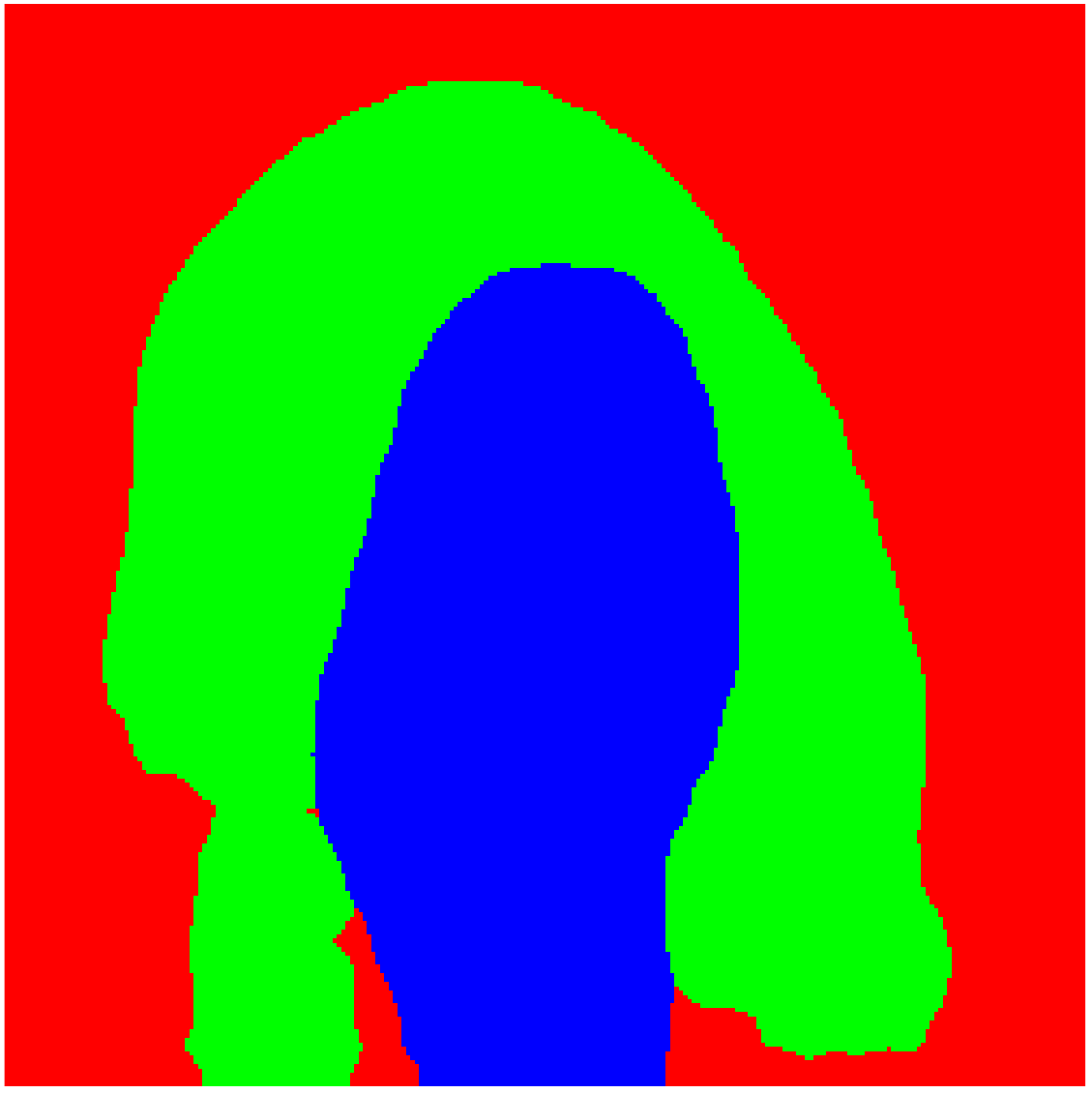}
\includegraphics[width=0.06\textwidth]{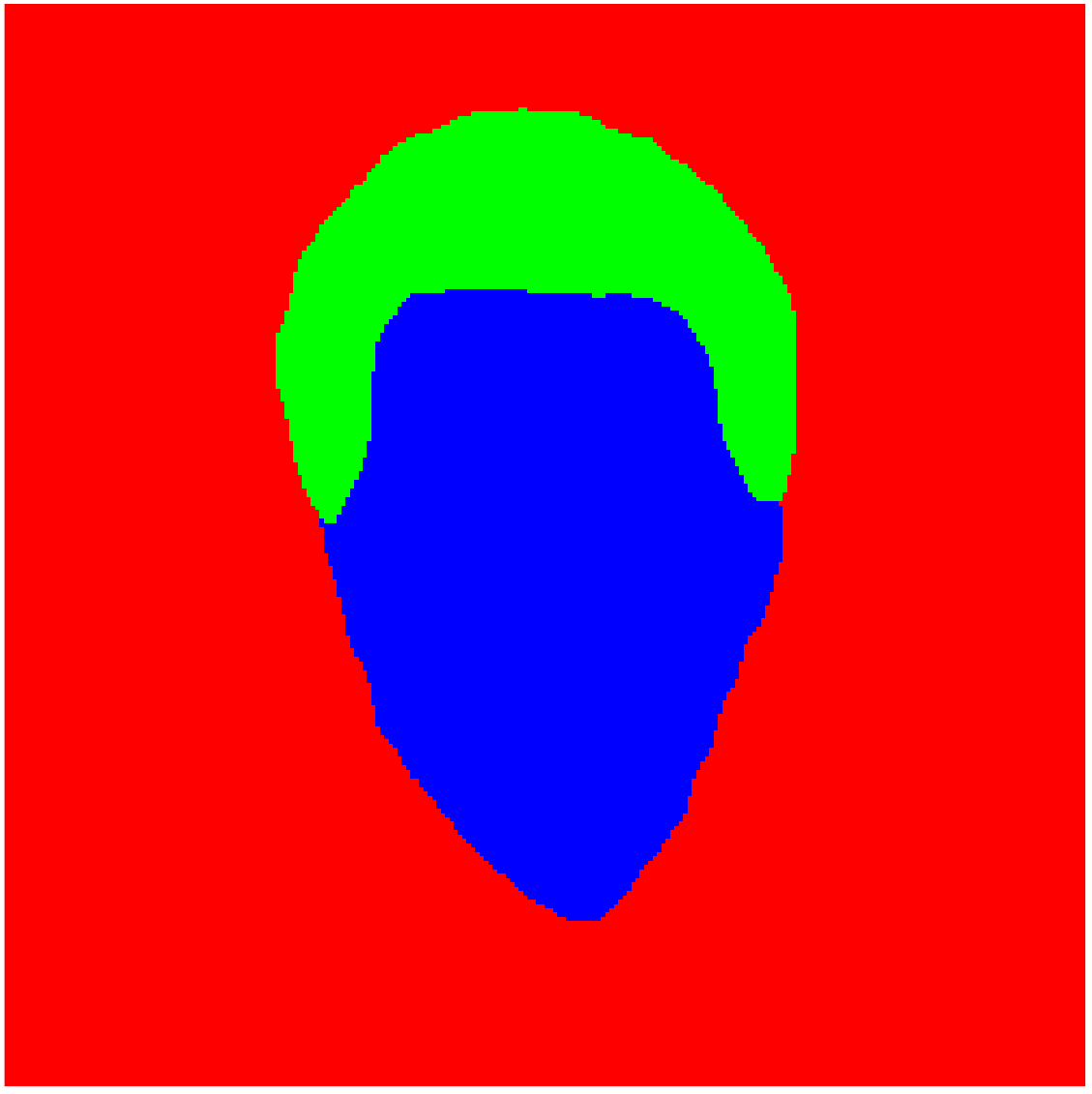}
\includegraphics[width=0.06\textwidth]{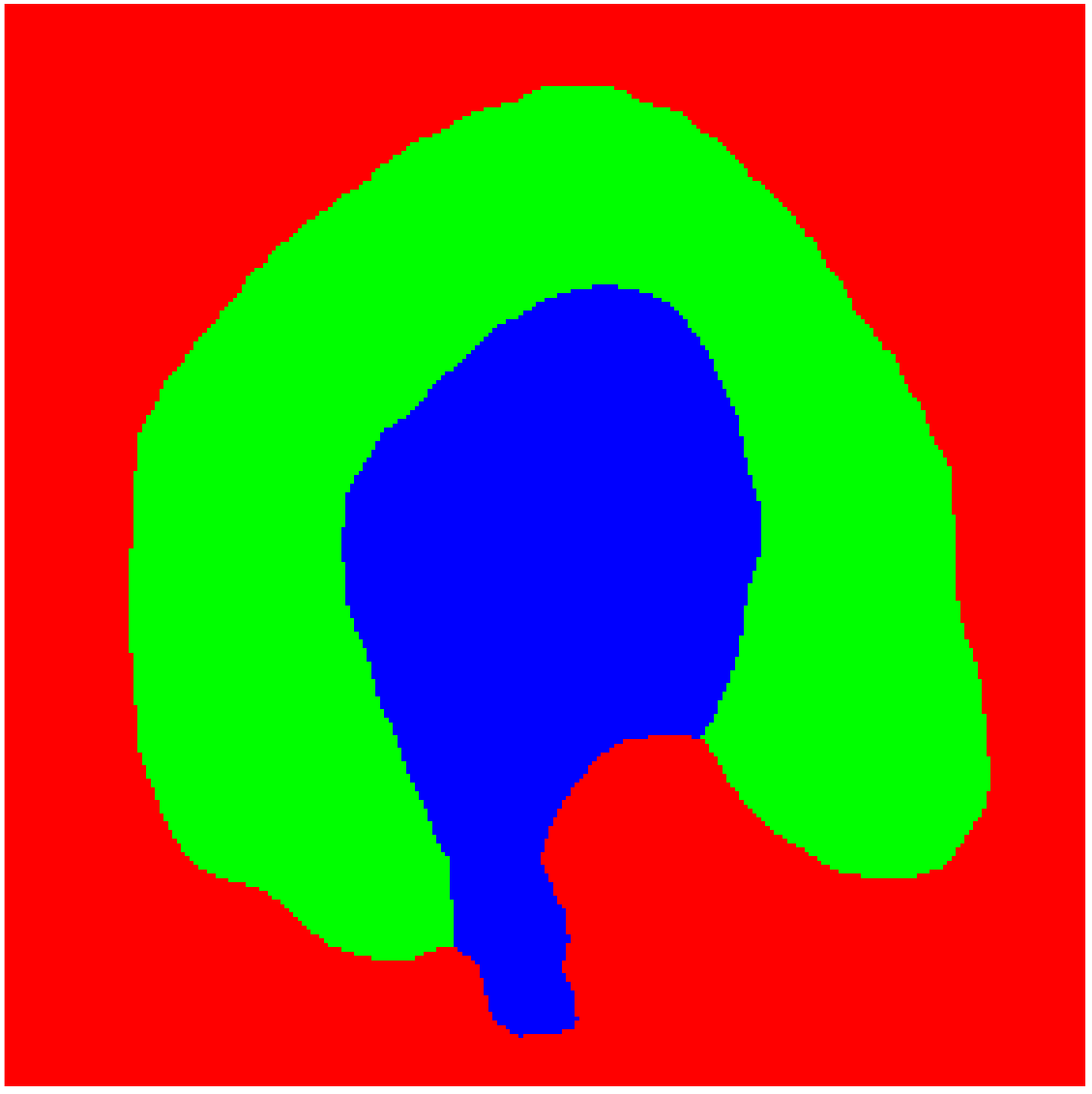}
\includegraphics[width=0.06\textwidth]{Janet_Napolitano_0002_cnn}
\includegraphics[width=0.06\textwidth]{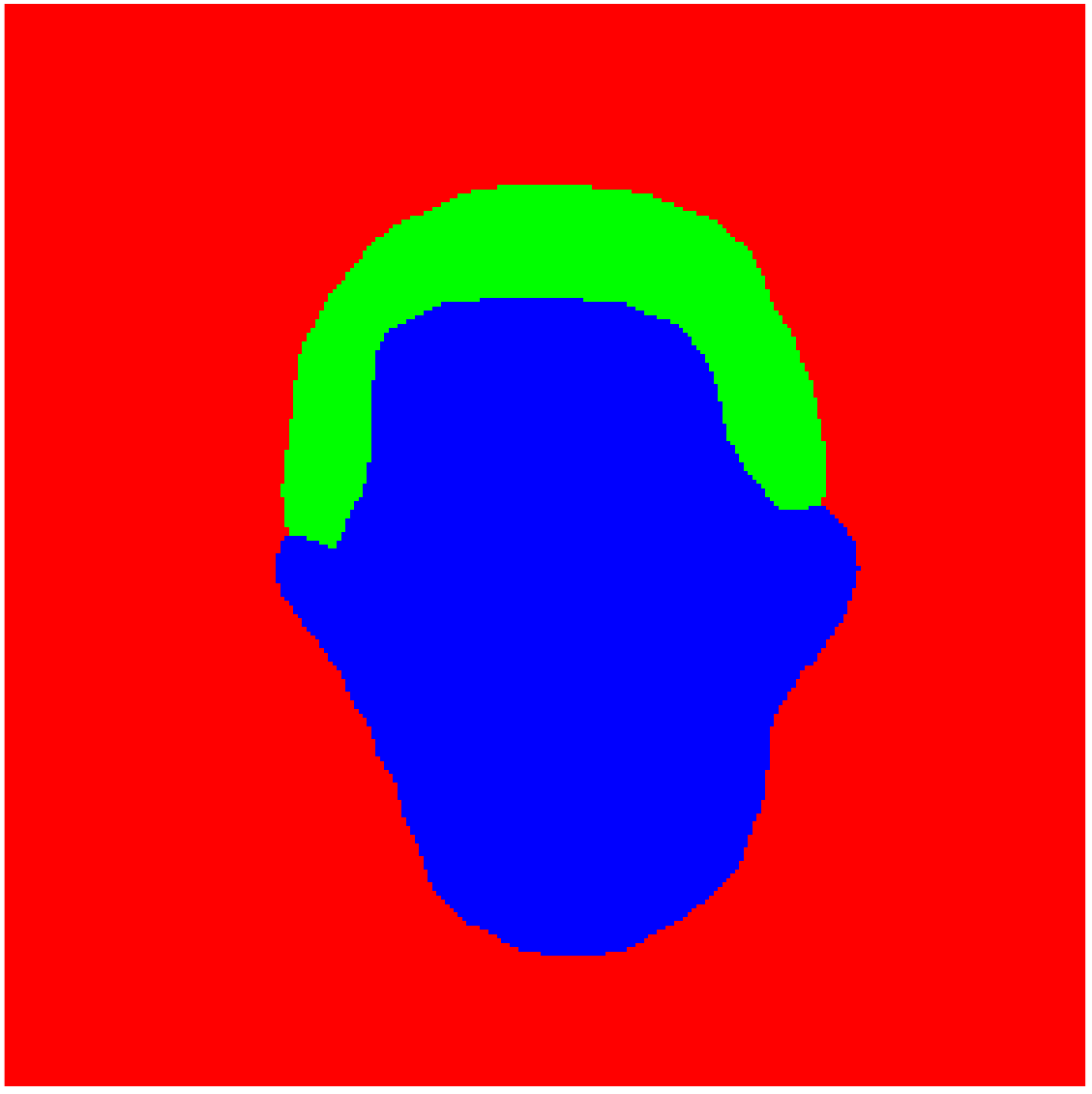}
\includegraphics[width=0.06\textwidth]{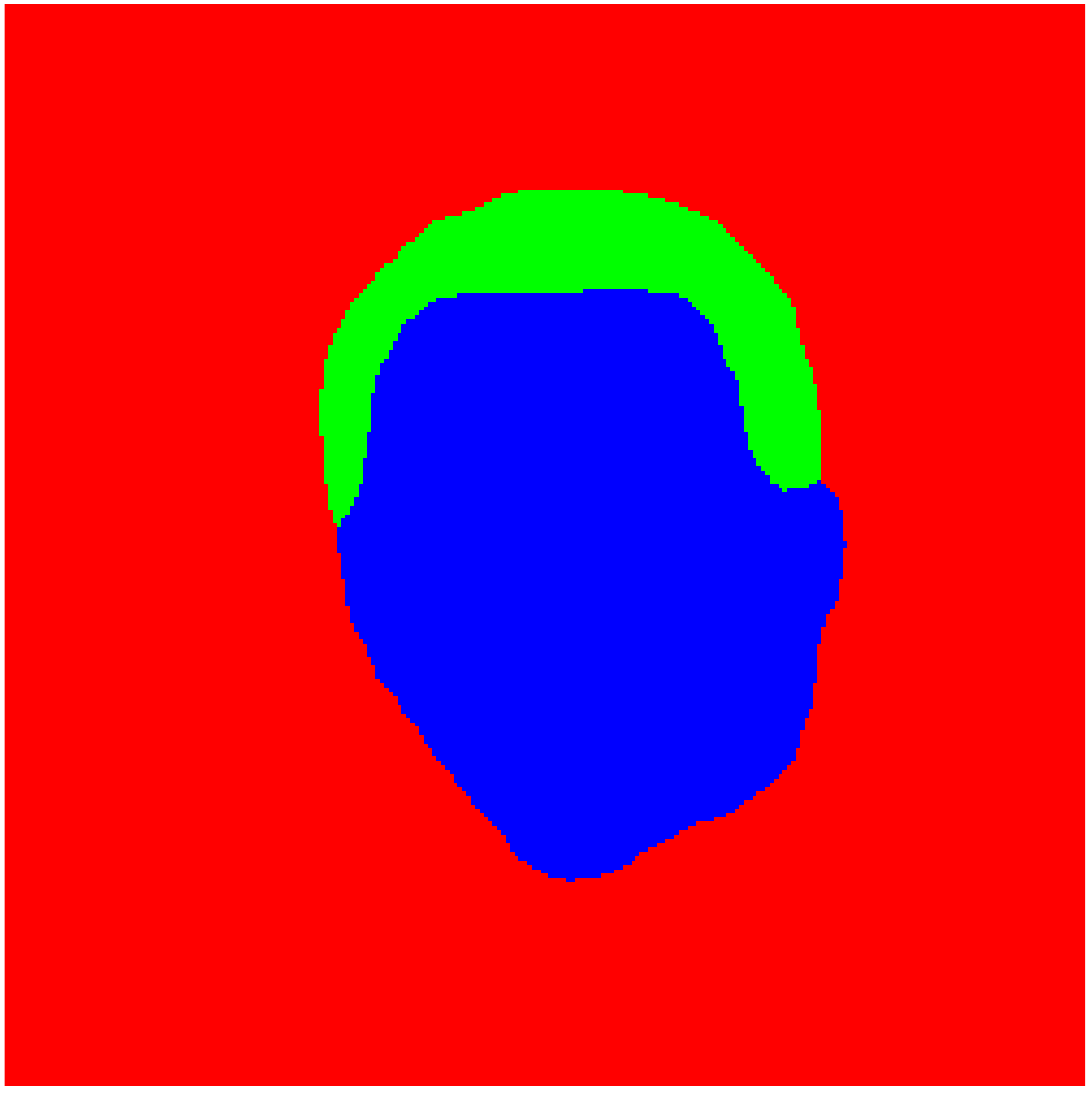}
\includegraphics[width=0.06\textwidth]{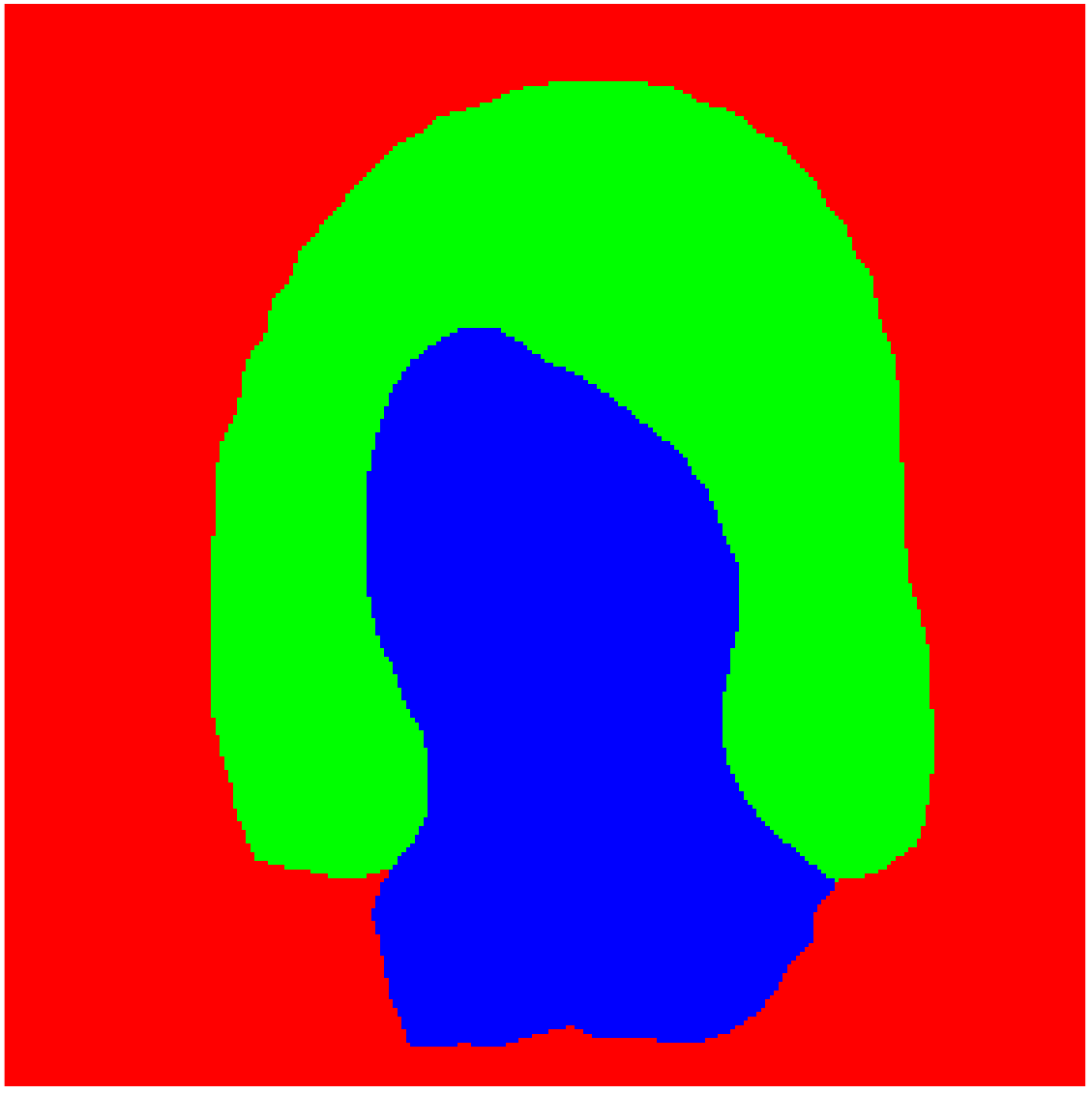}
\includegraphics[width=0.06\textwidth]{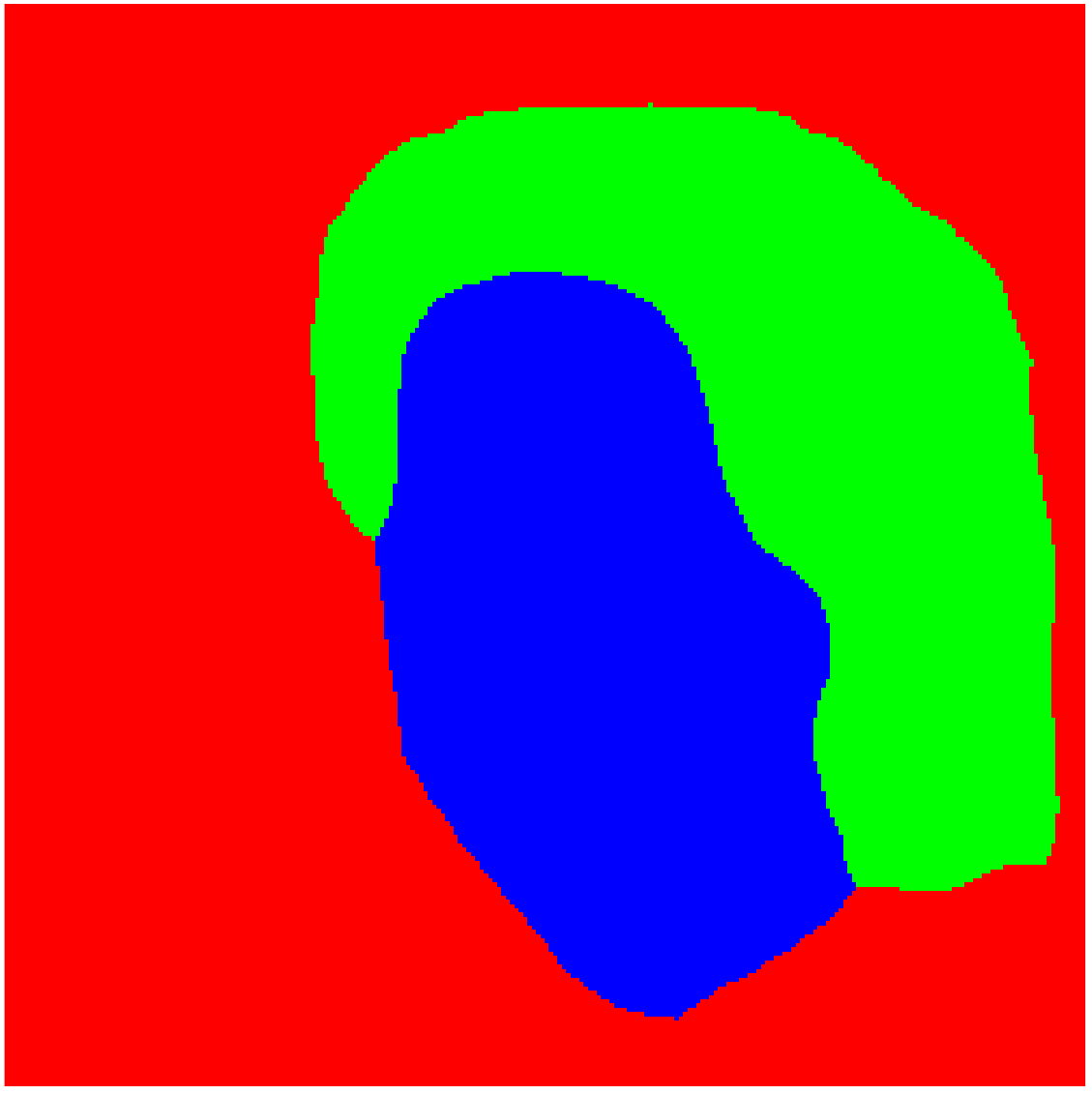}

\includegraphics[width=0.06\textwidth]{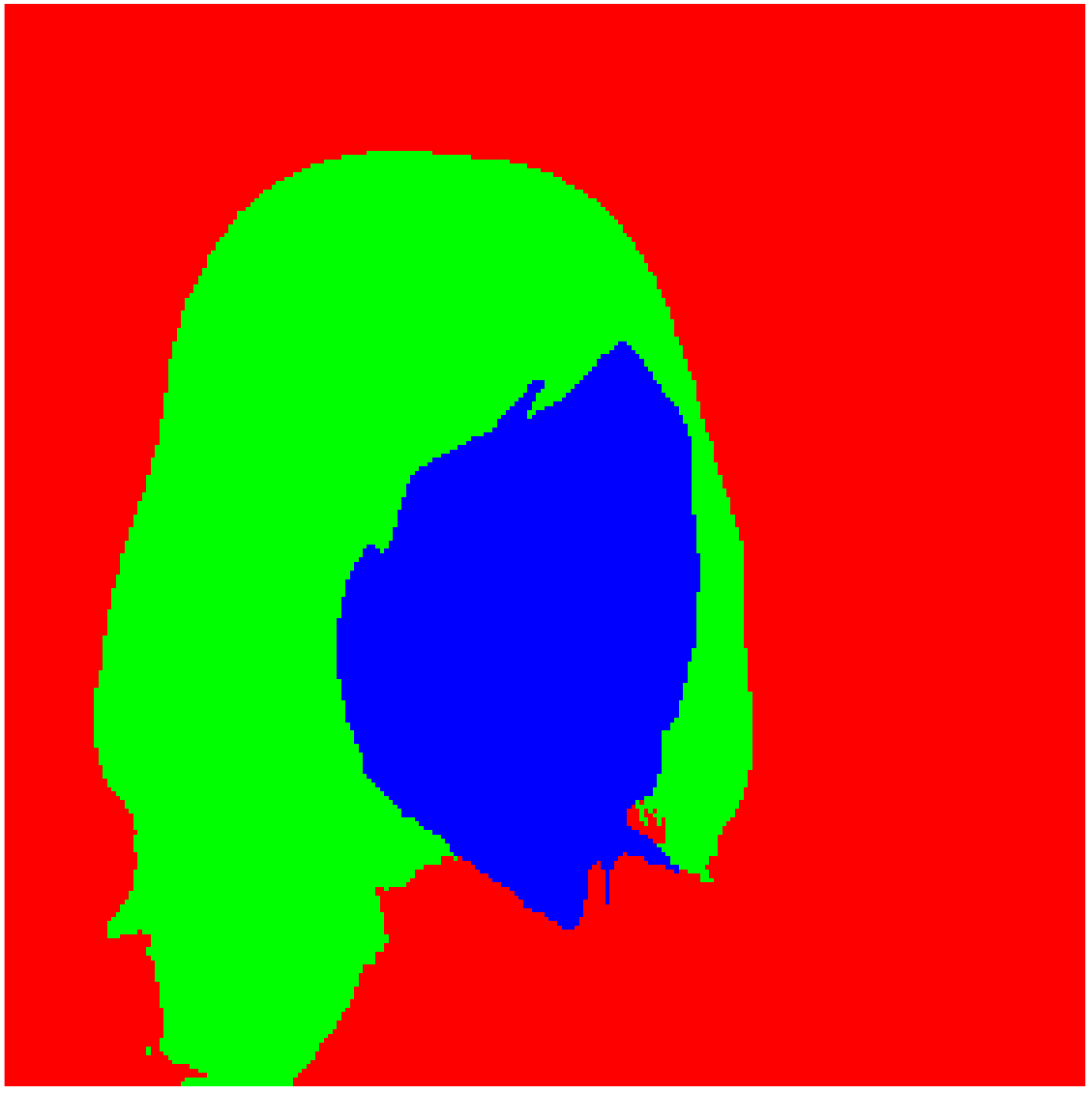}
\includegraphics[width=0.06\textwidth]{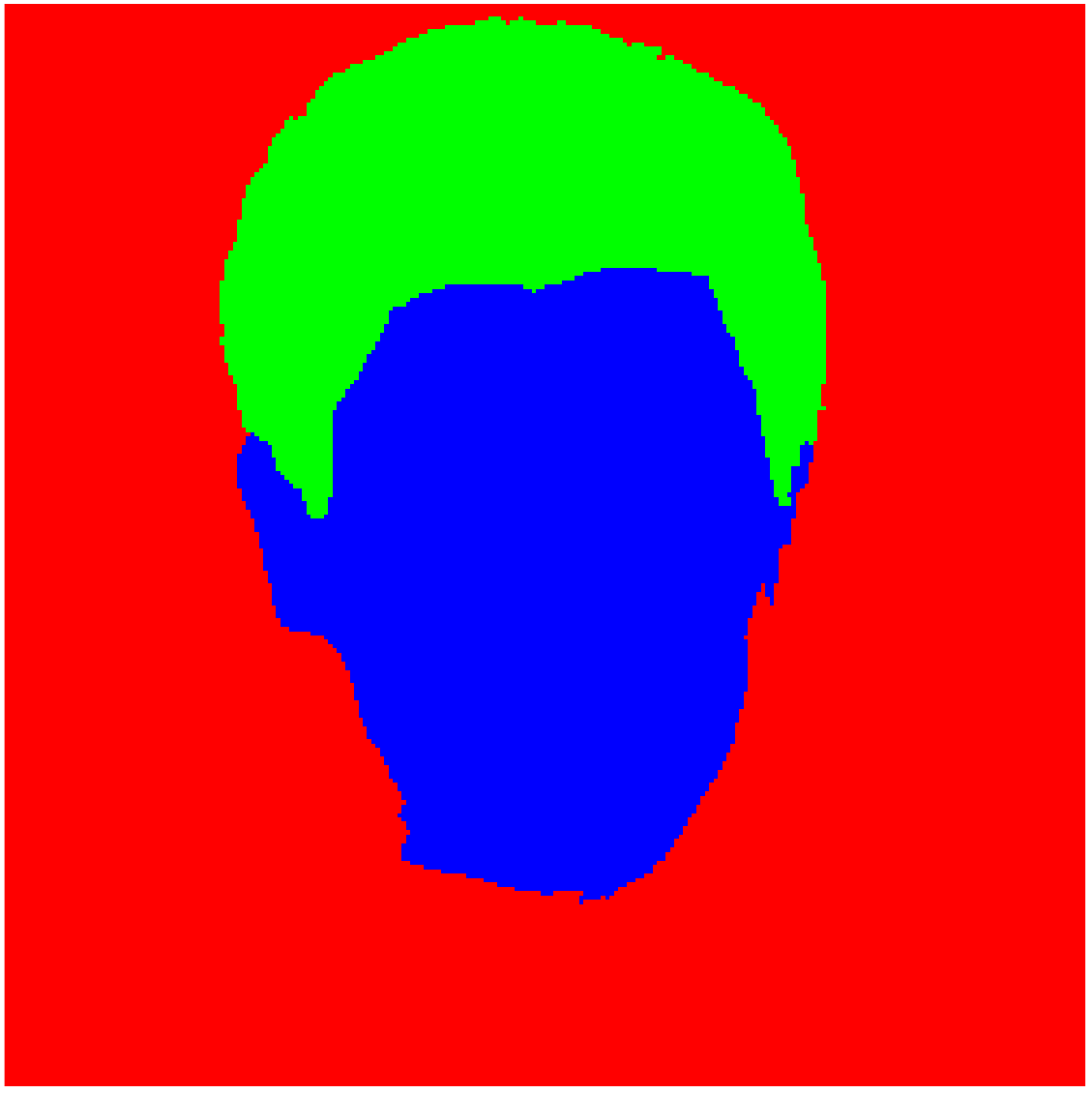}
\includegraphics[width=0.06\textwidth]{Jane_Pauley_0001_crf}
\includegraphics[width=0.06\textwidth]{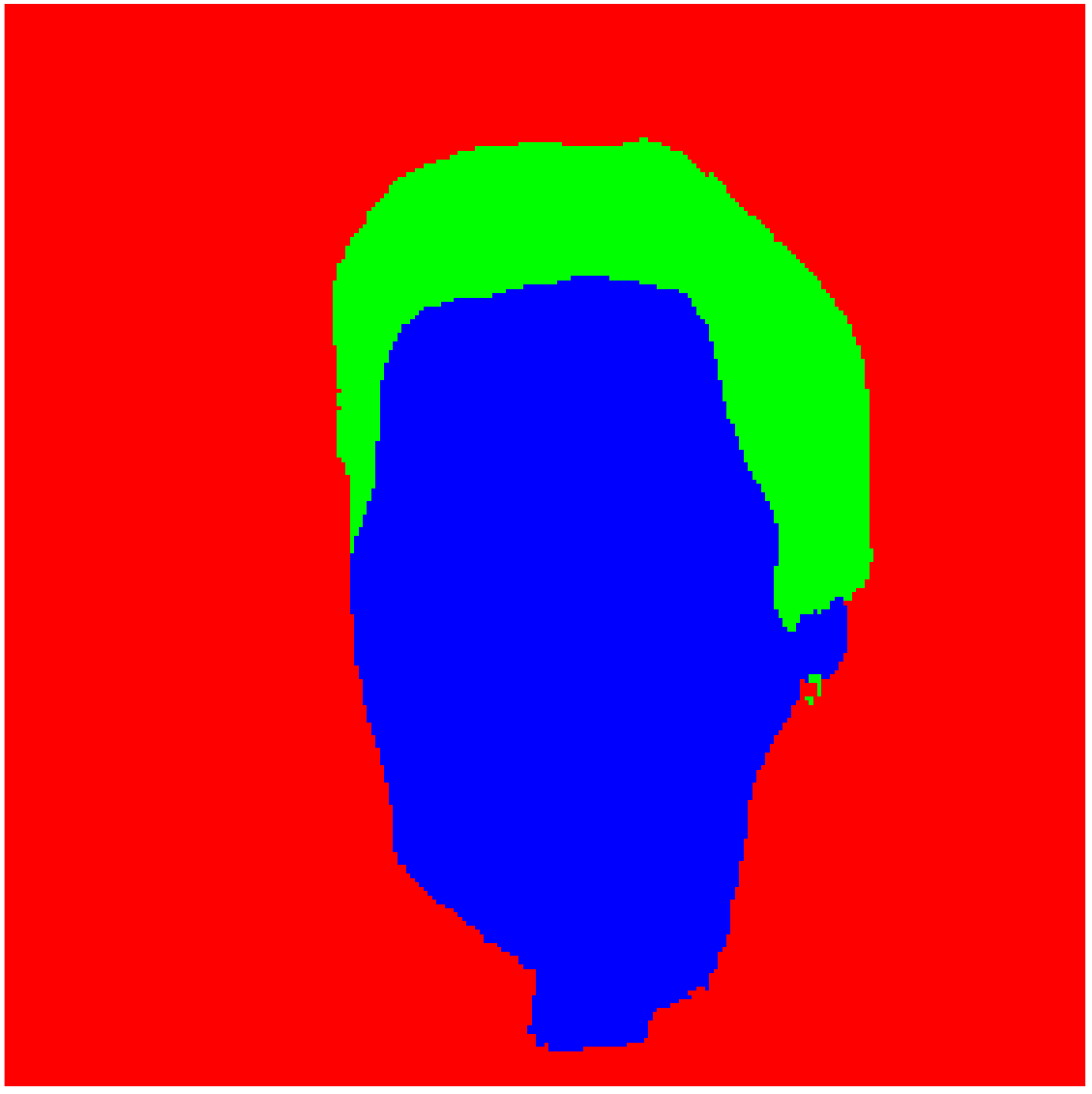}
\includegraphics[width=0.06\textwidth]{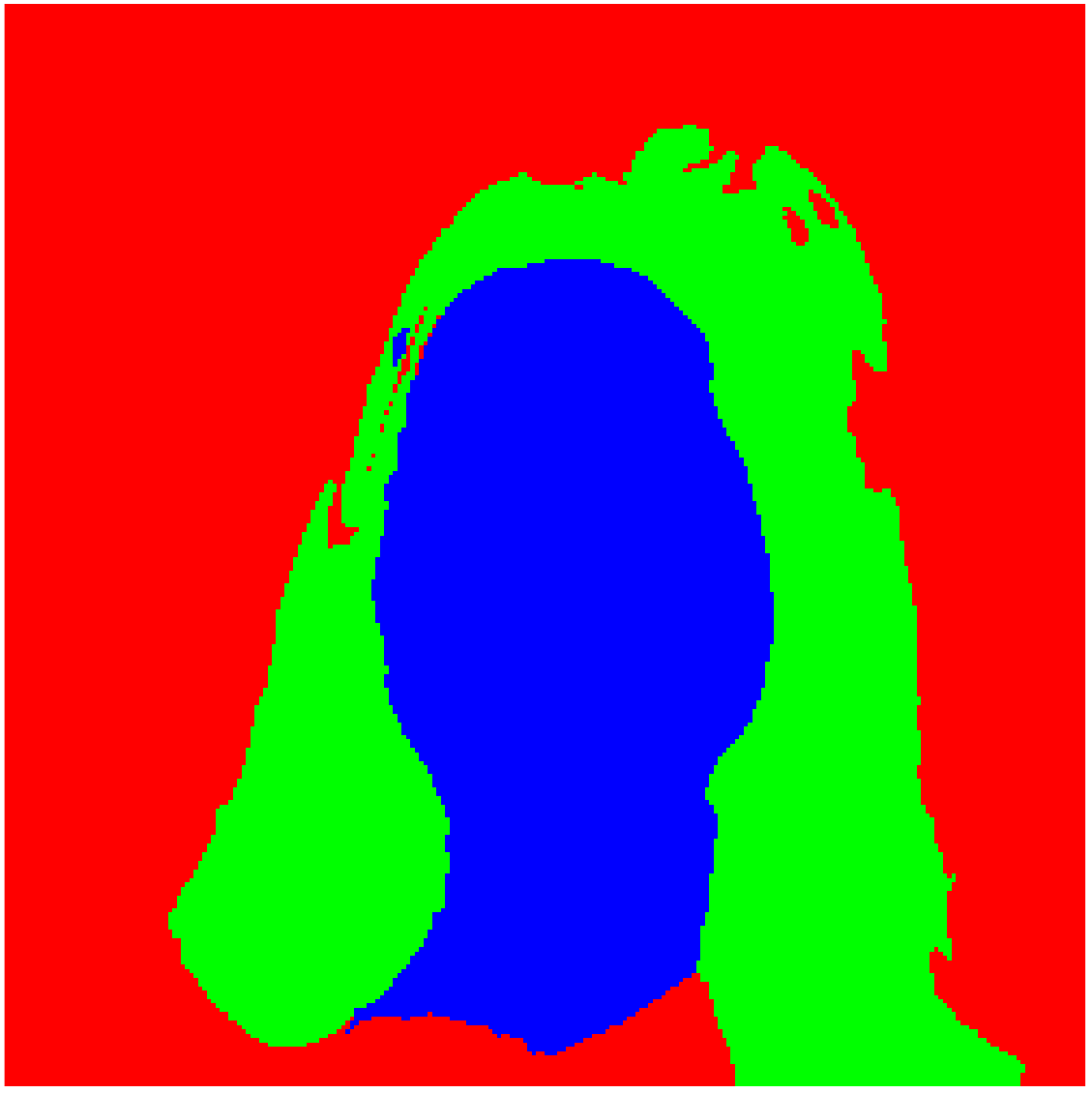}
\includegraphics[width=0.06\textwidth]{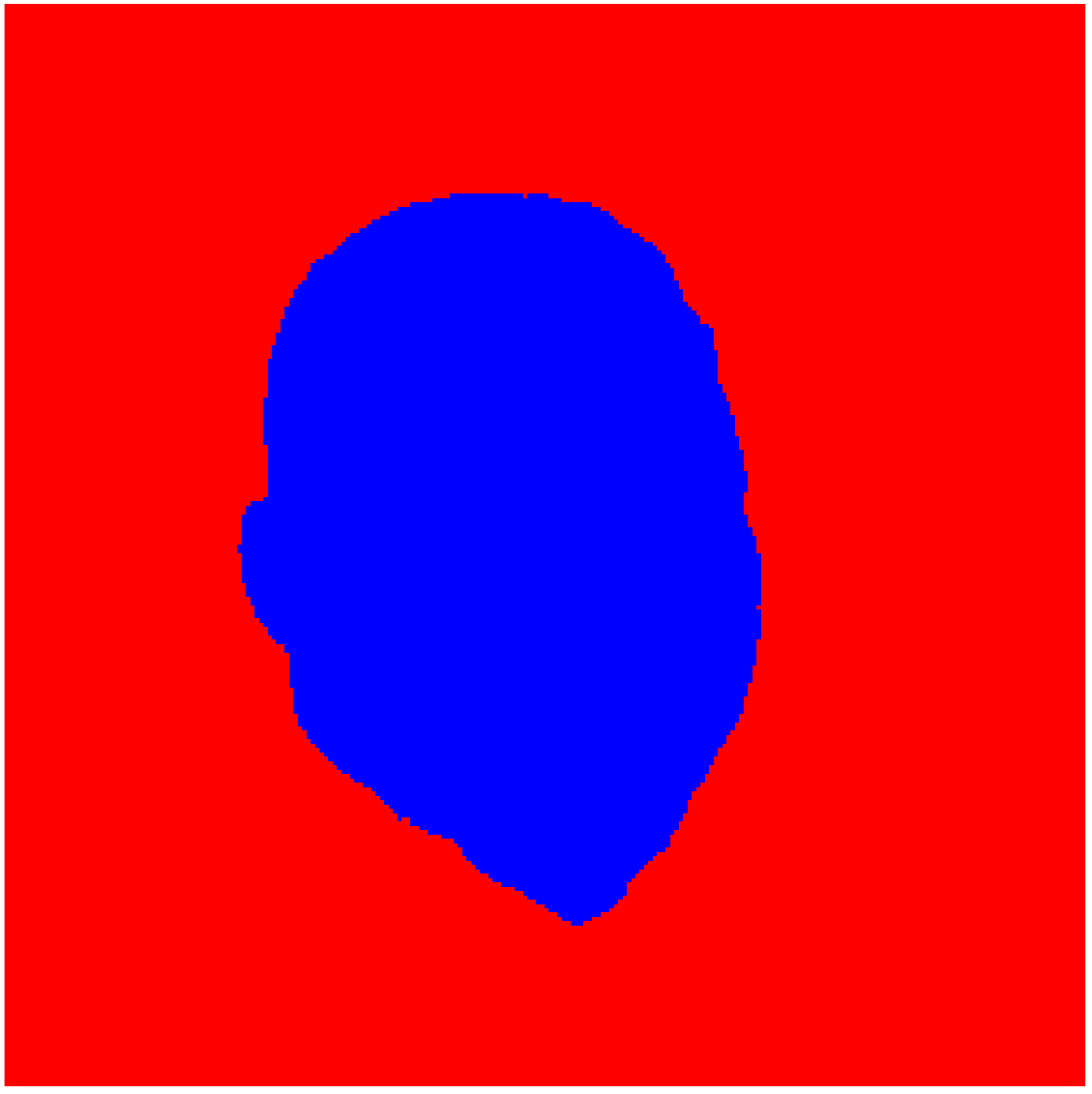}
\includegraphics[width=0.06\textwidth]{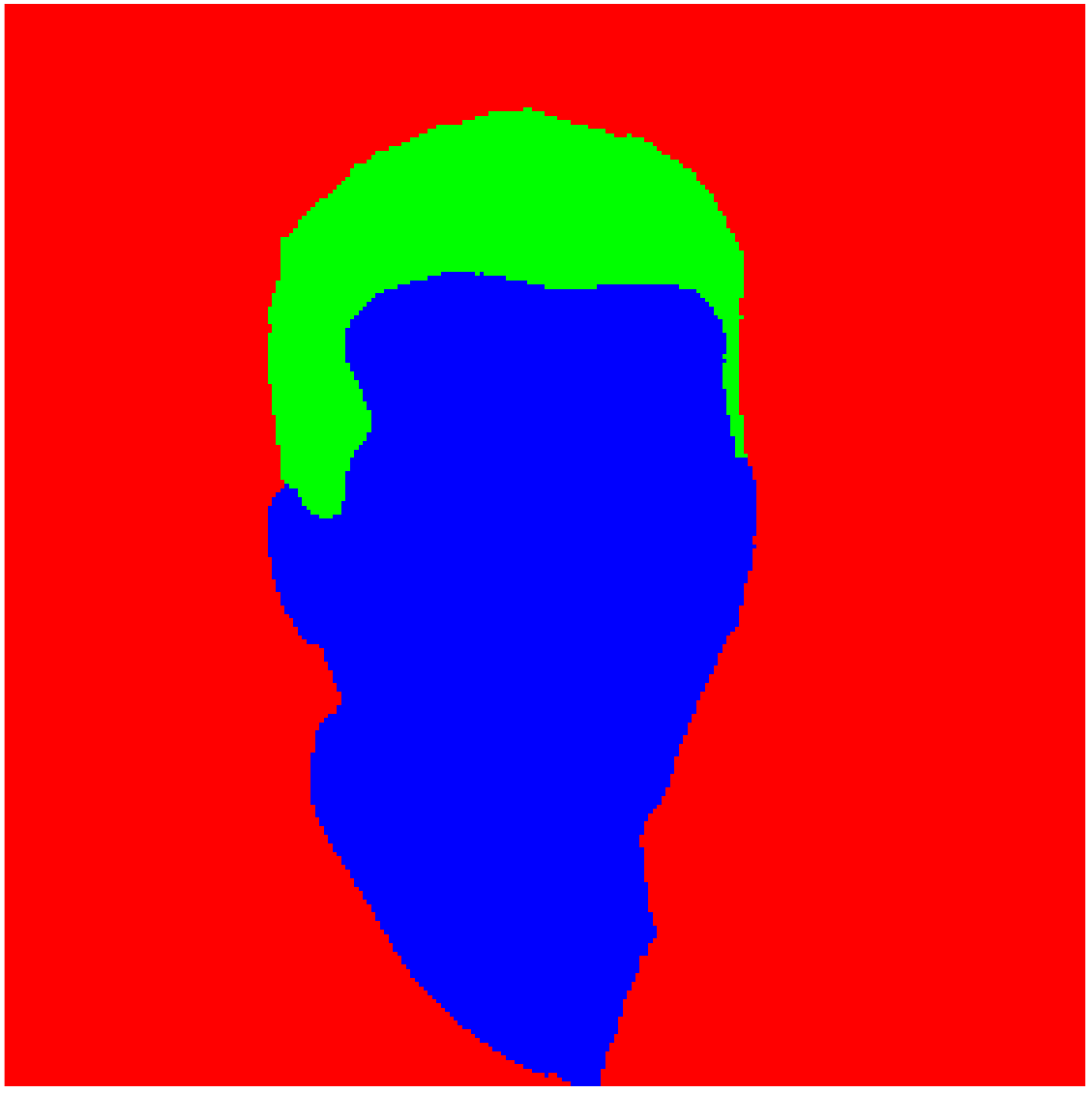}
\includegraphics[width=0.06\textwidth]{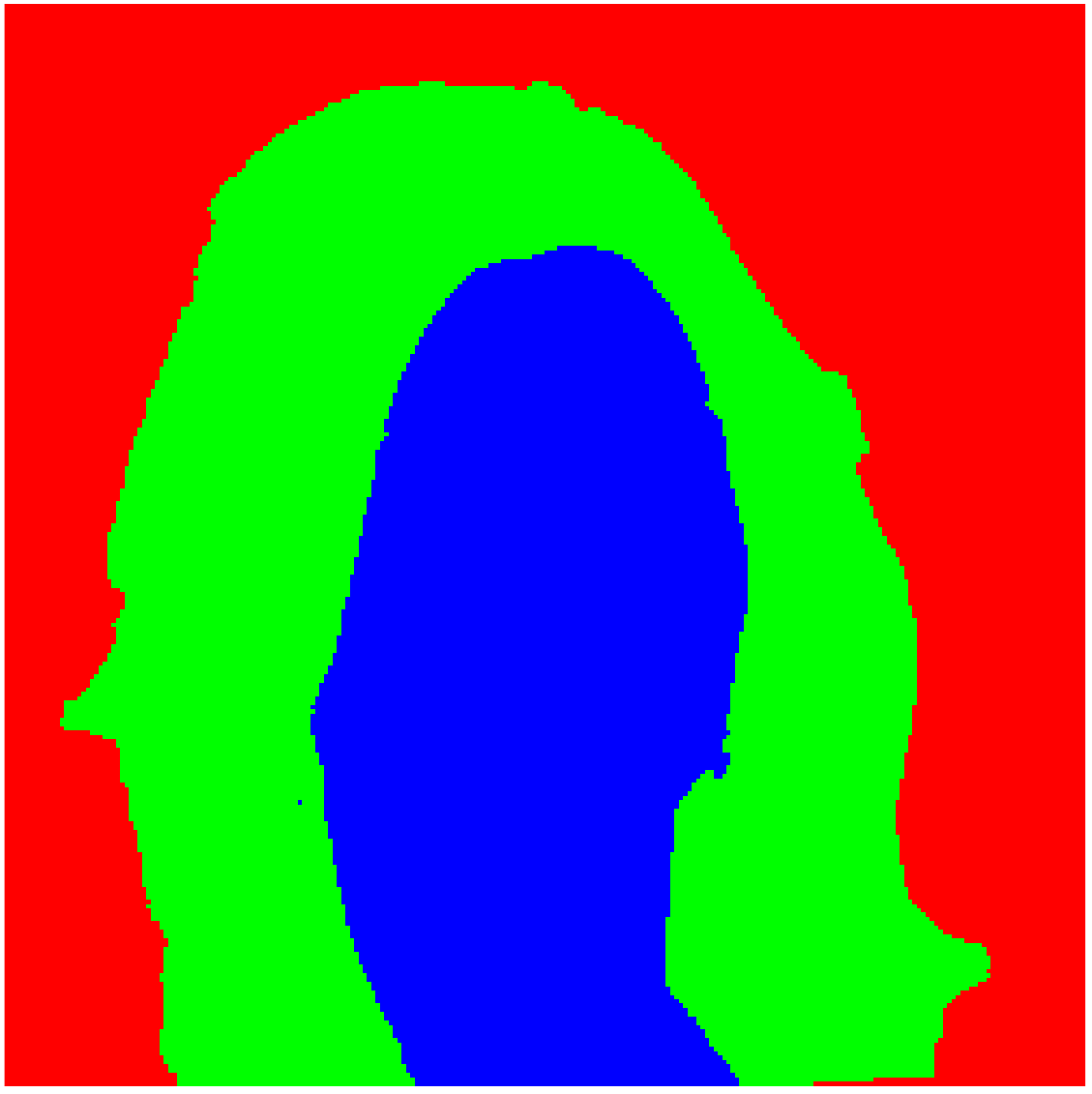}
\includegraphics[width=0.06\textwidth]{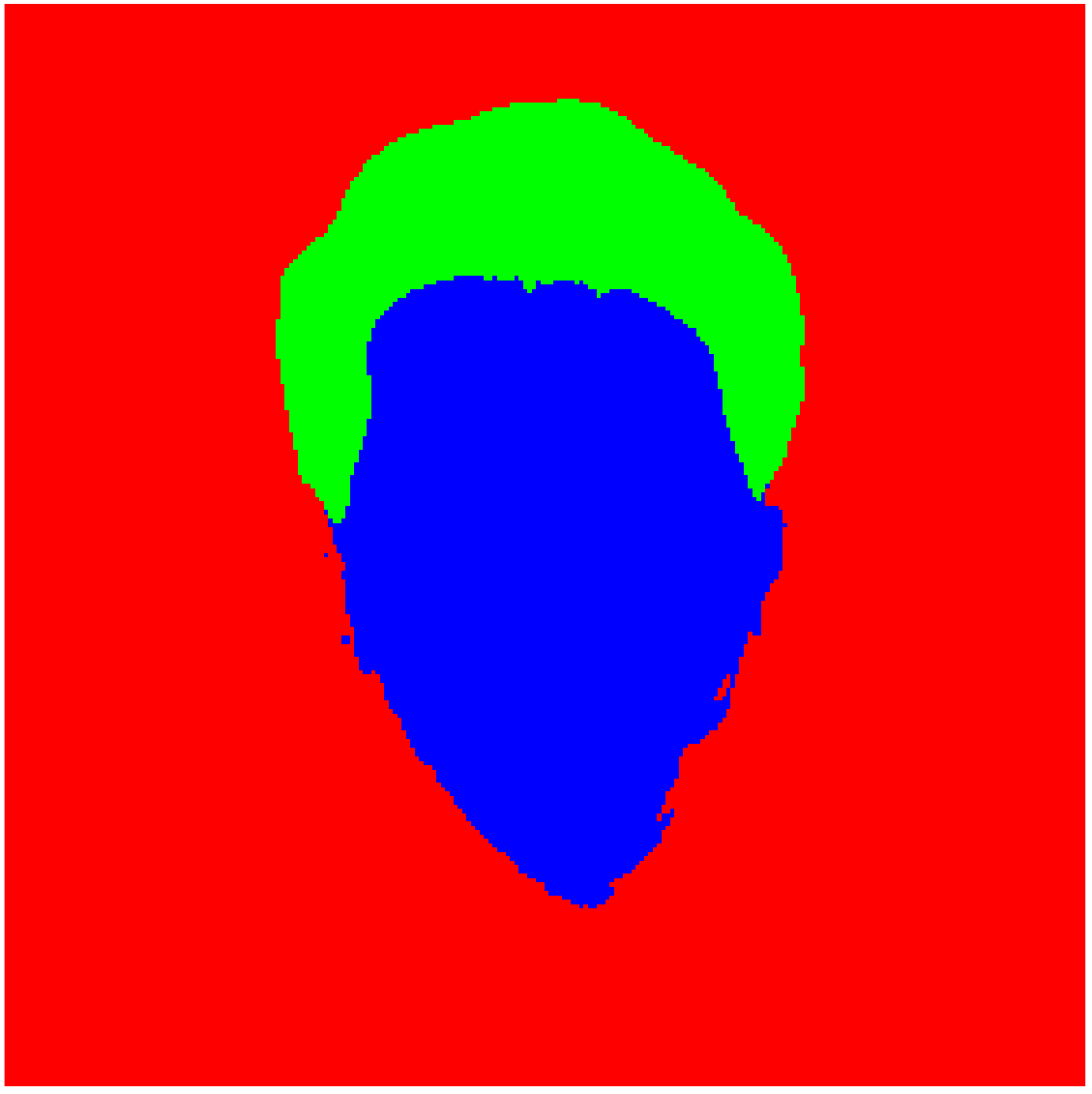}
\includegraphics[width=0.06\textwidth]{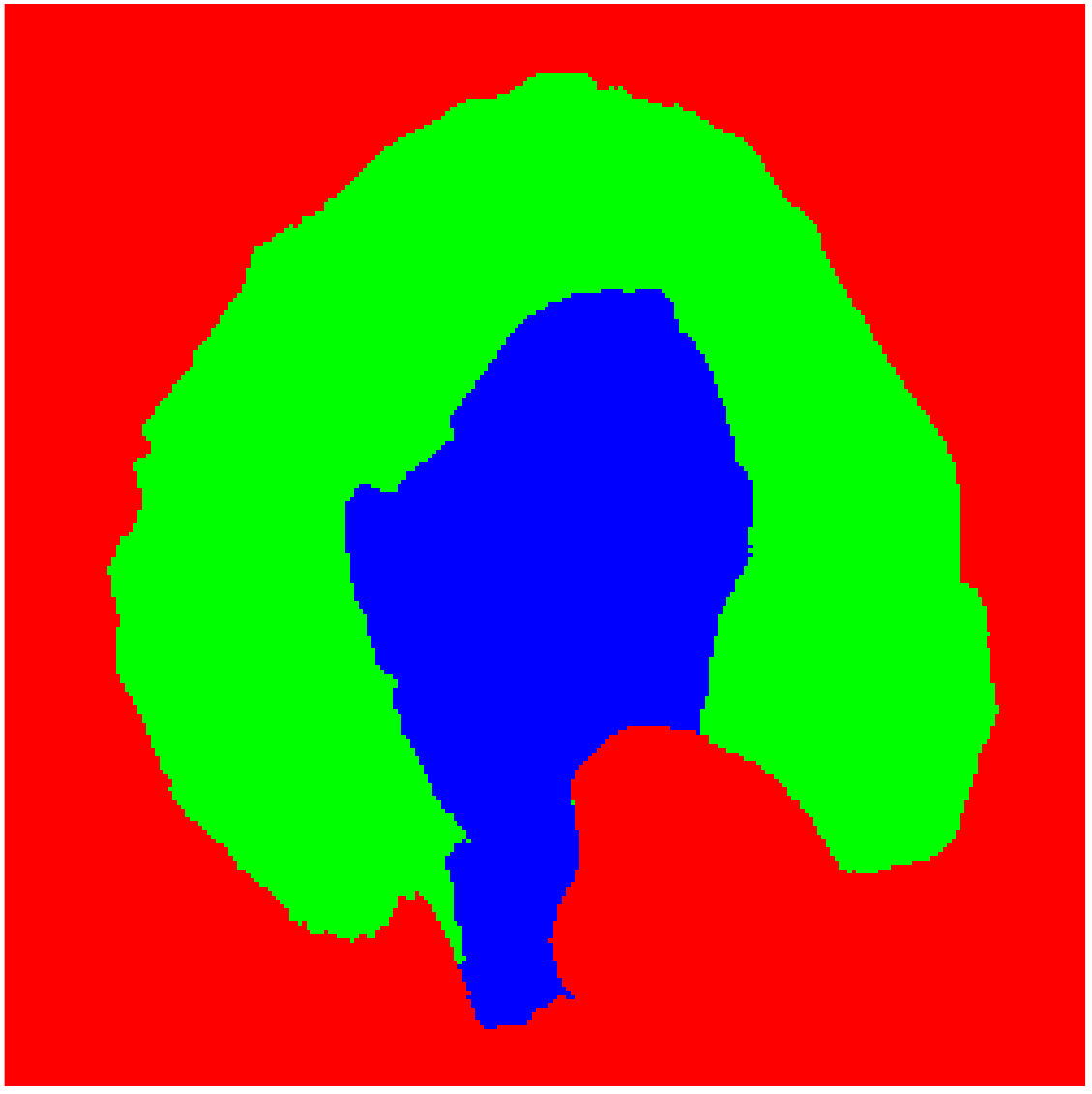}
\includegraphics[width=0.06\textwidth]{Janet_Napolitano_0002_crf}
\includegraphics[width=0.06\textwidth]{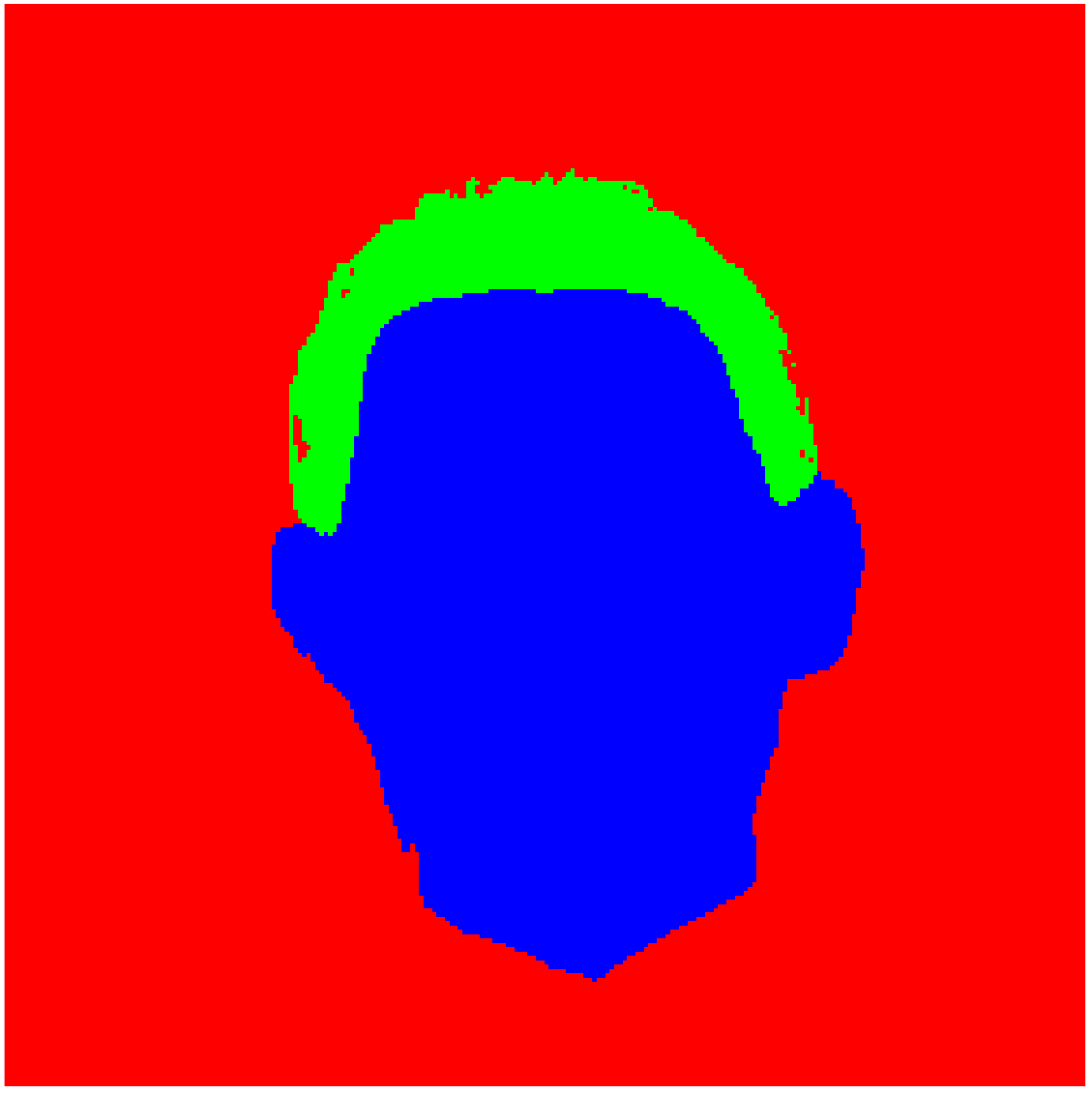}
\includegraphics[width=0.06\textwidth]{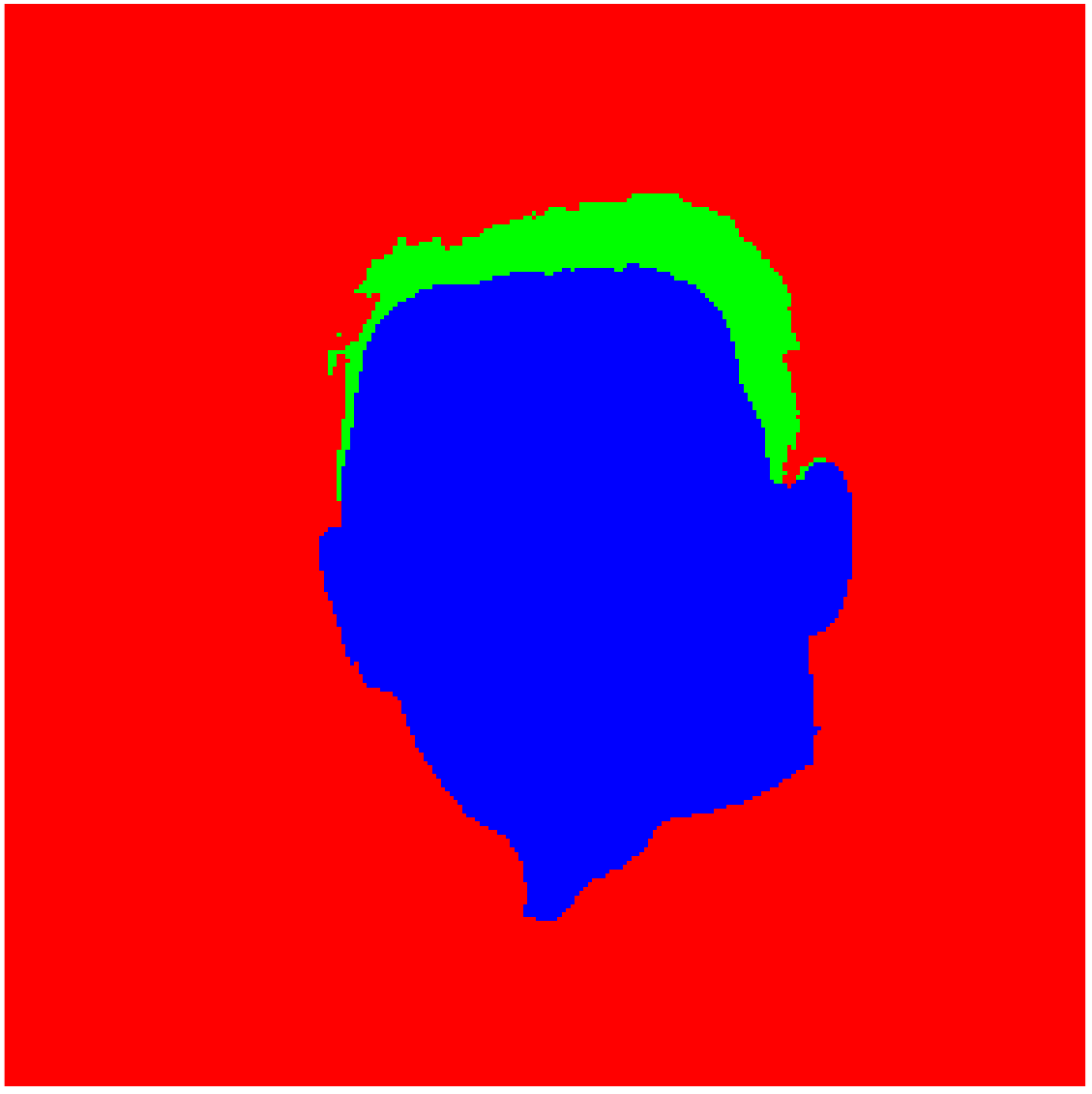}
\includegraphics[width=0.06\textwidth]{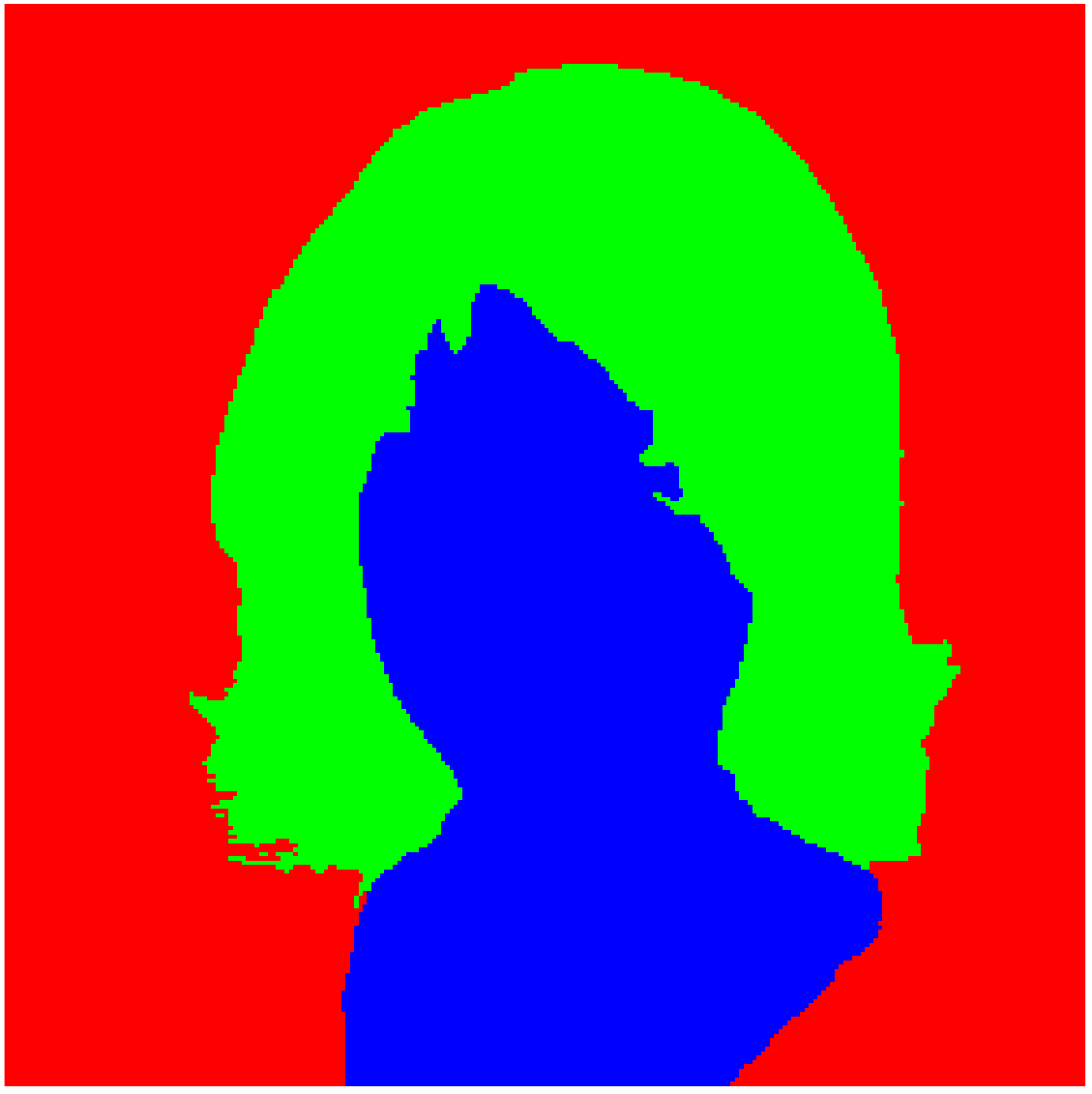}
\includegraphics[width=0.06\textwidth]{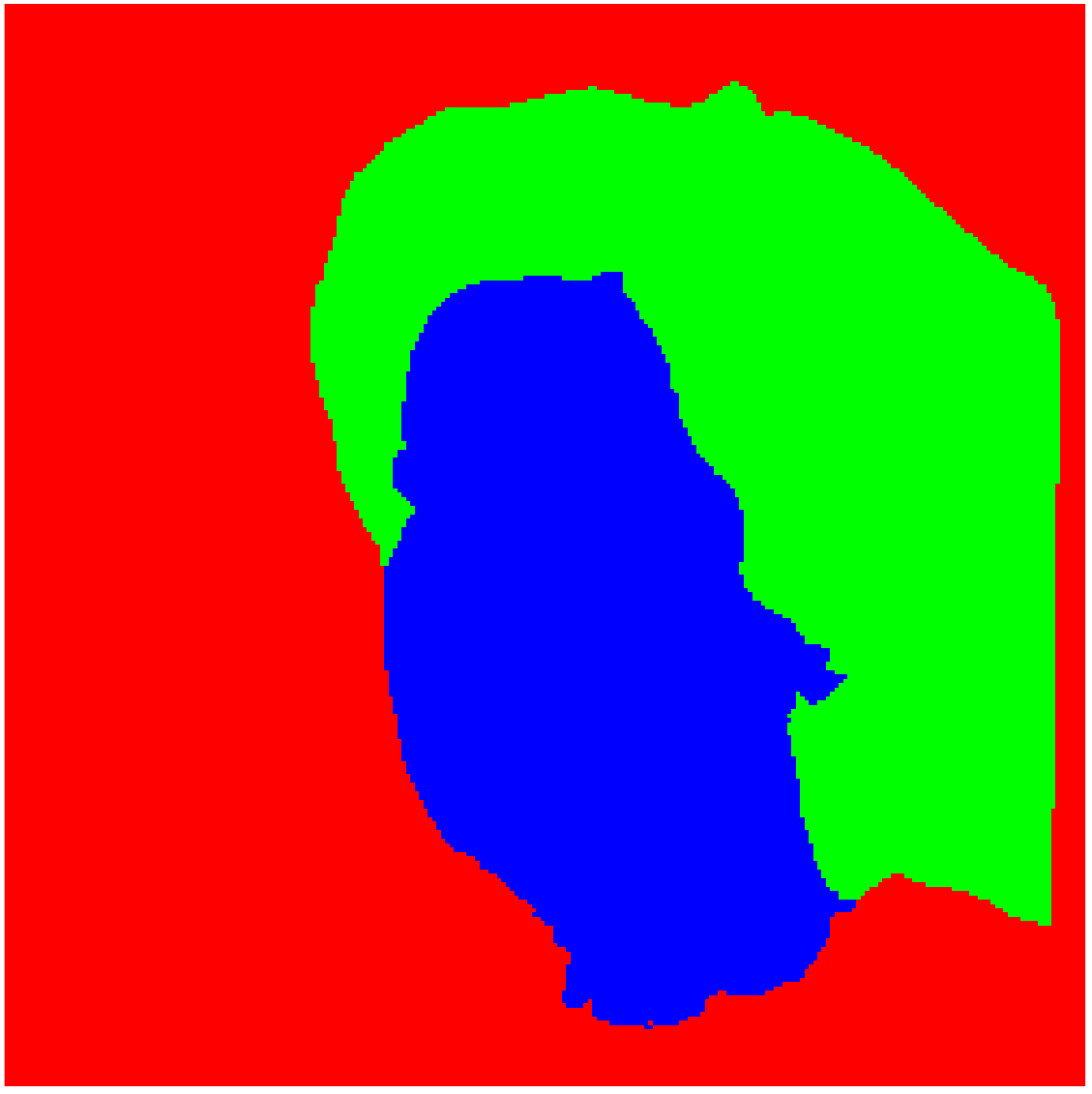}

\includegraphics[width=0.06\textwidth]{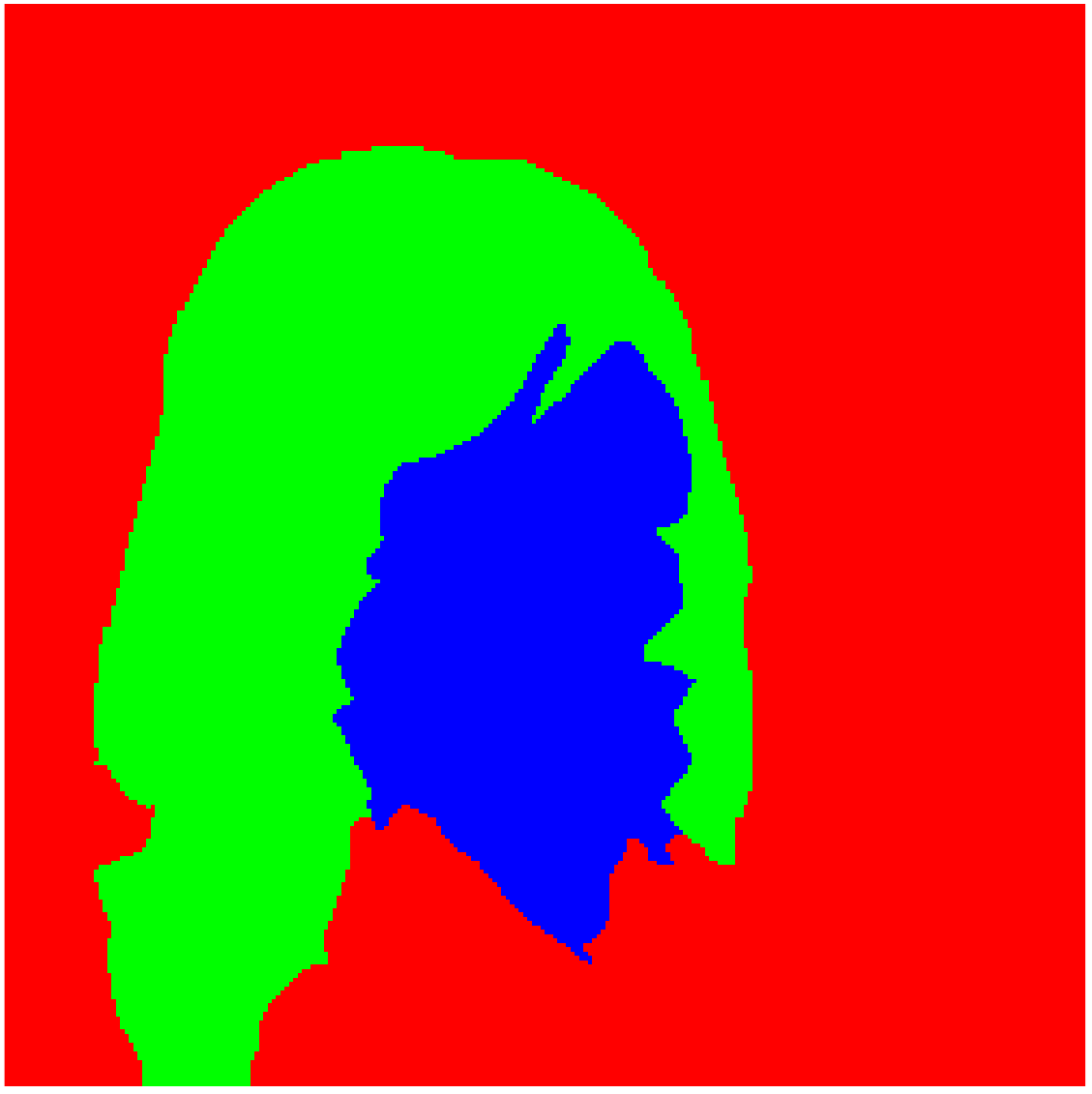}
\includegraphics[width=0.06\textwidth]{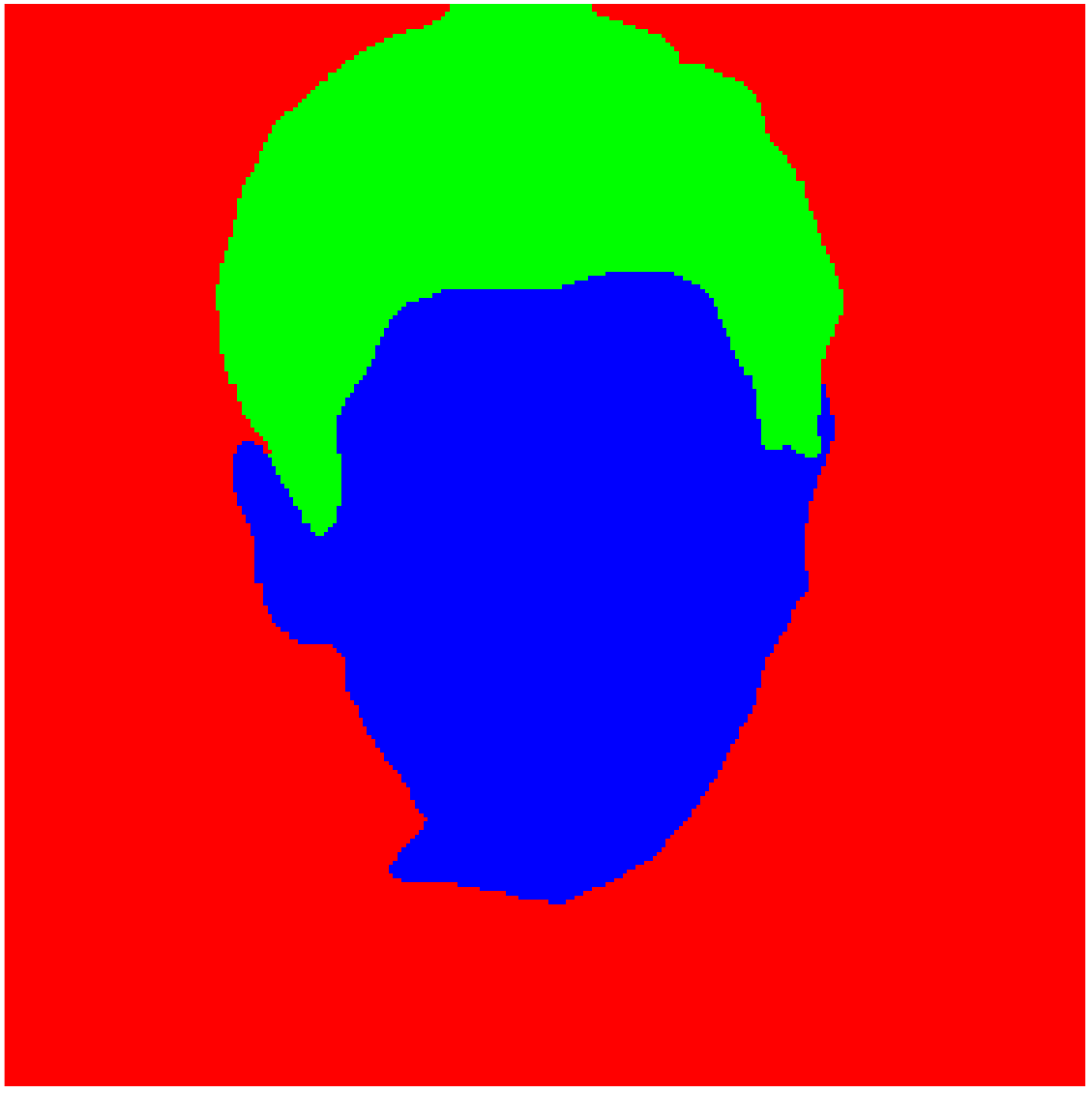}
\includegraphics[width=0.06\textwidth]{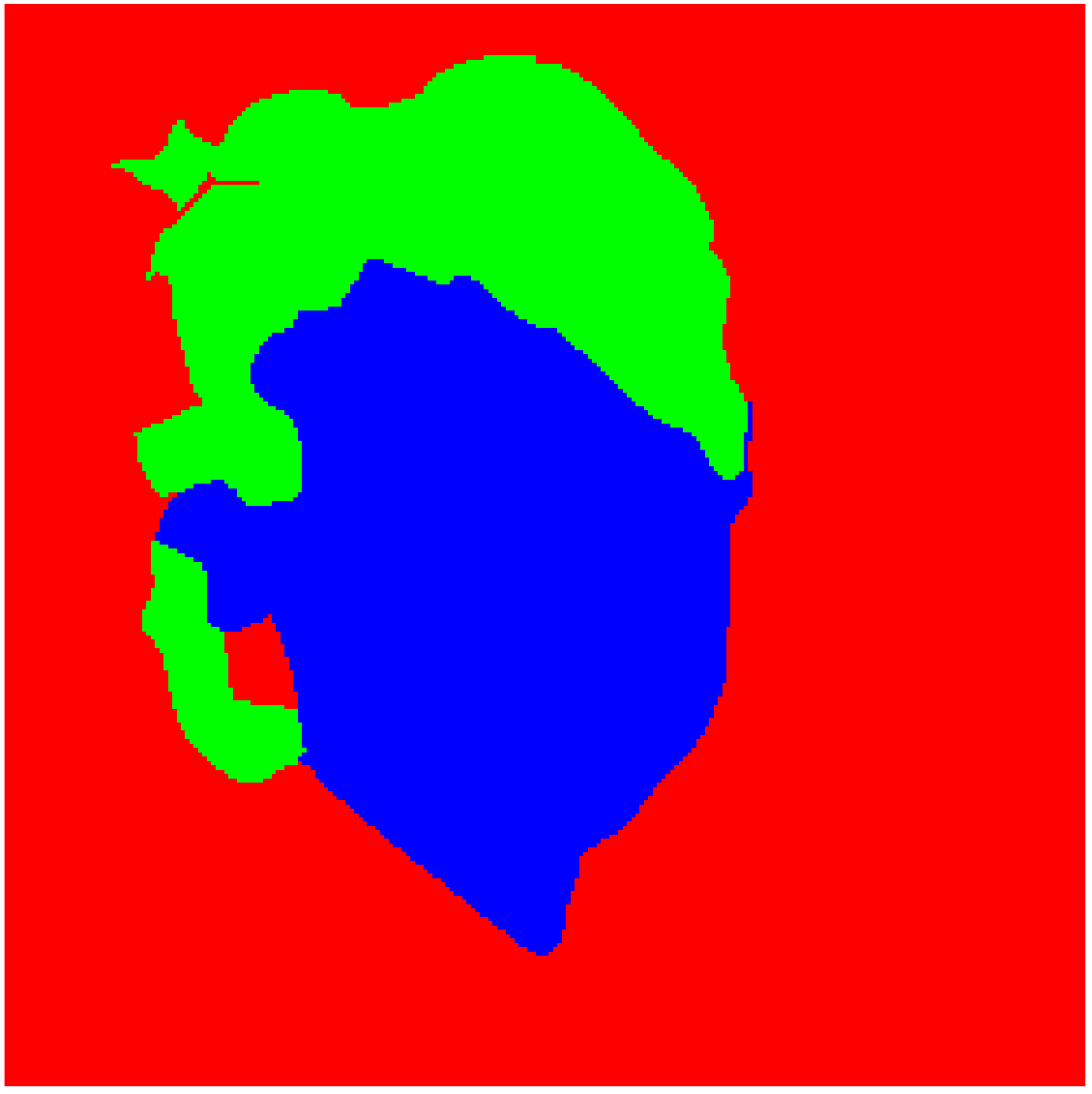}
\includegraphics[width=0.06\textwidth]{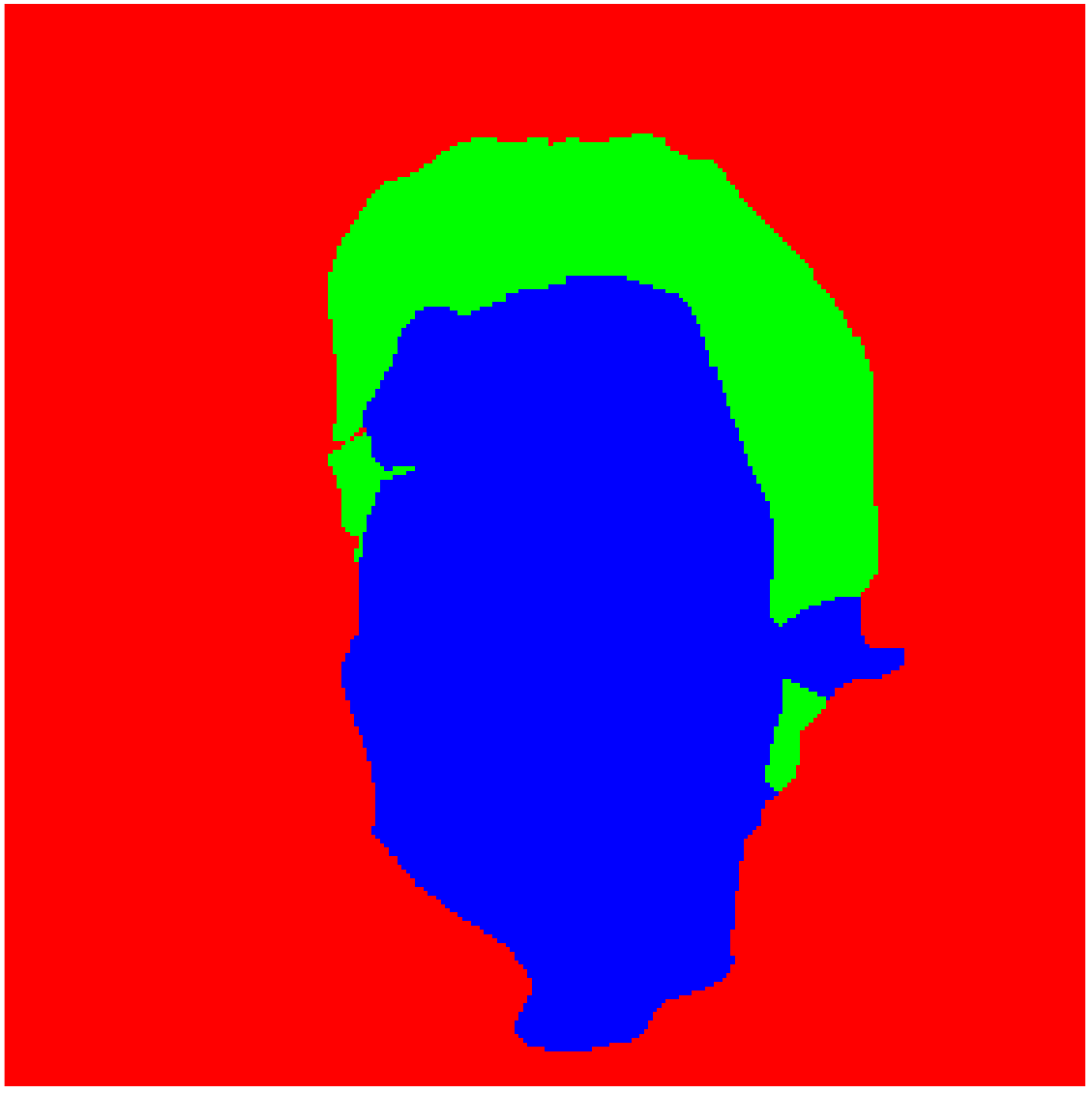}
\includegraphics[width=0.06\textwidth]{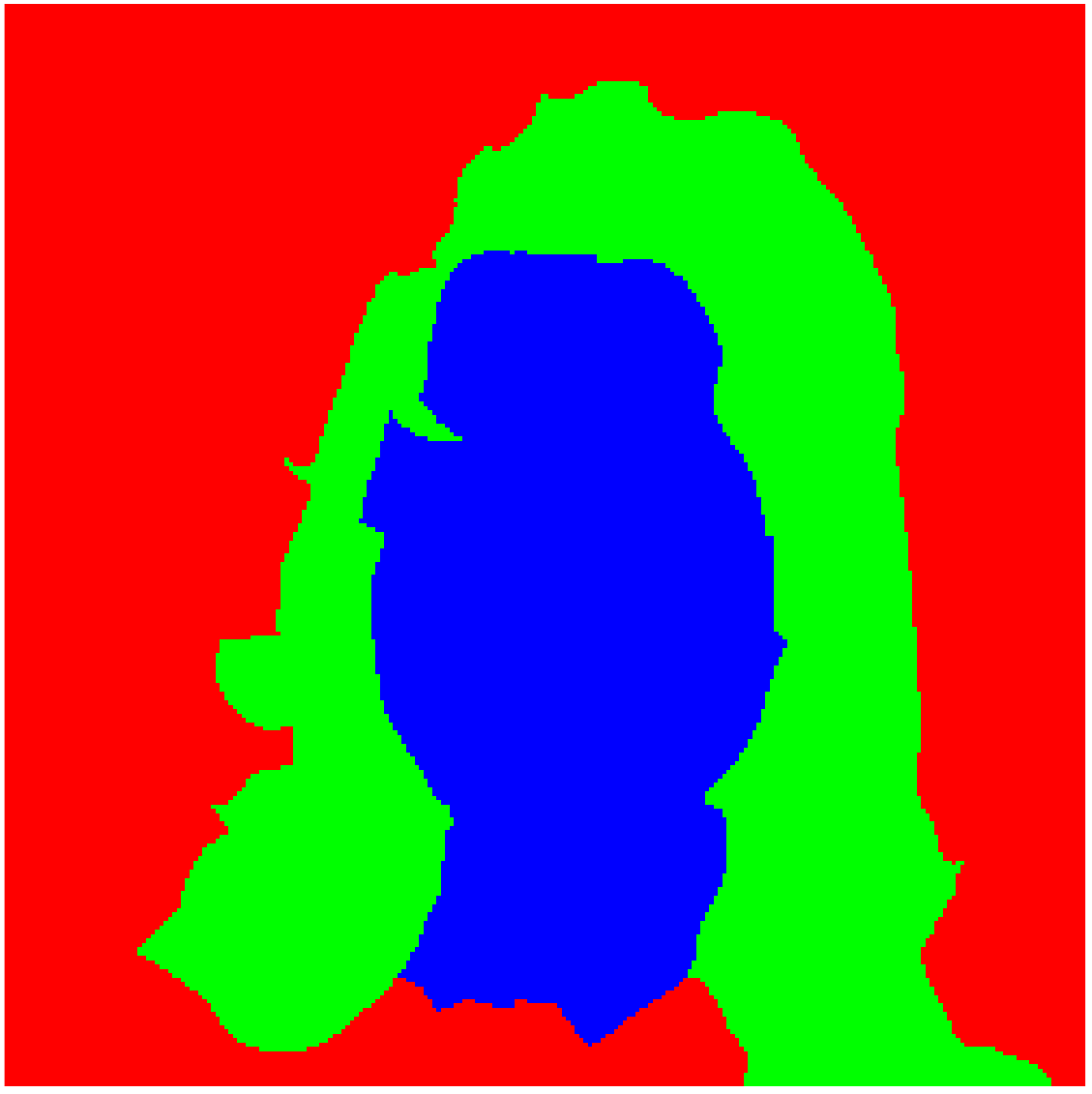}
\includegraphics[width=0.06\textwidth]{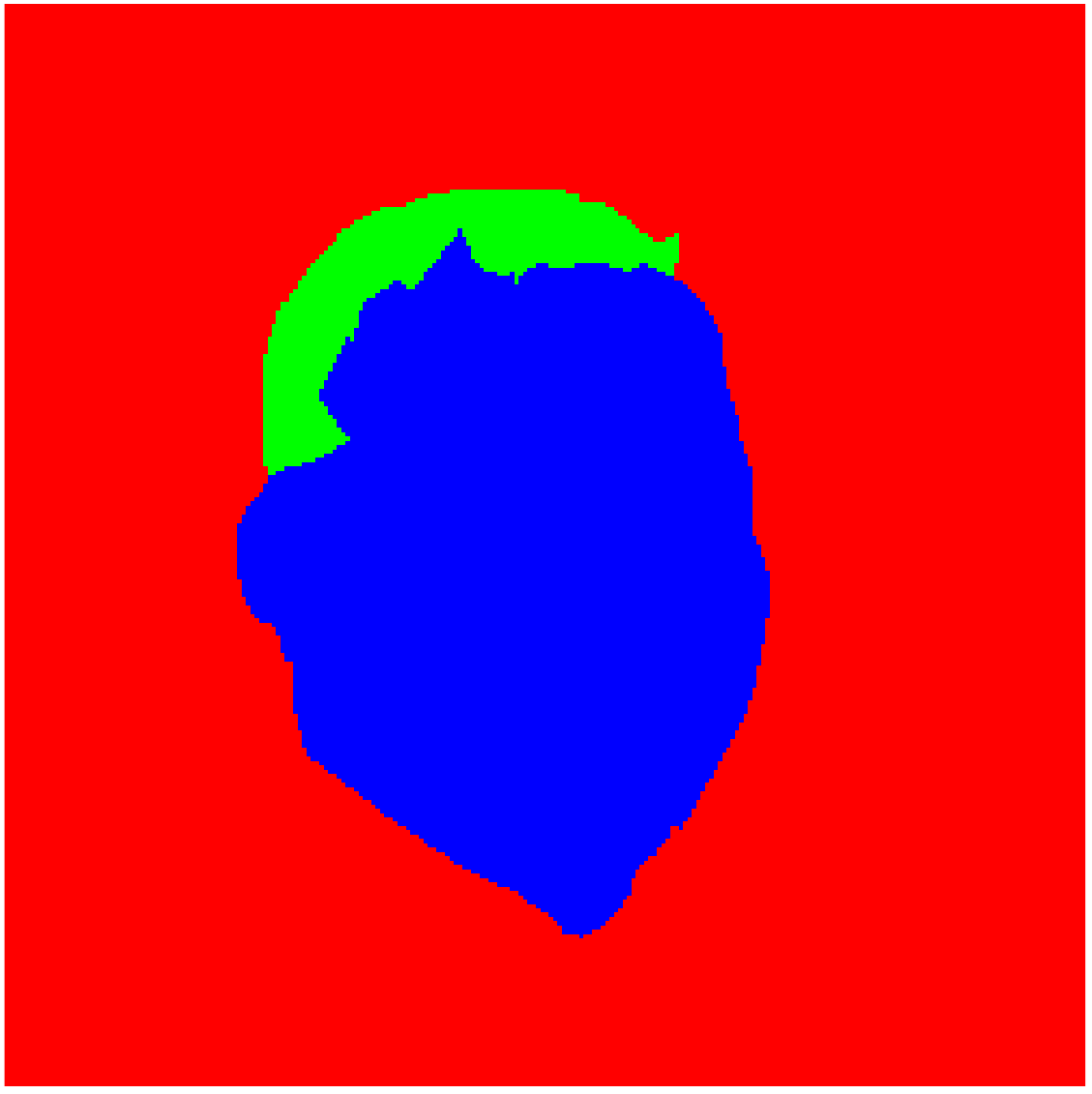}
\includegraphics[width=0.06\textwidth]{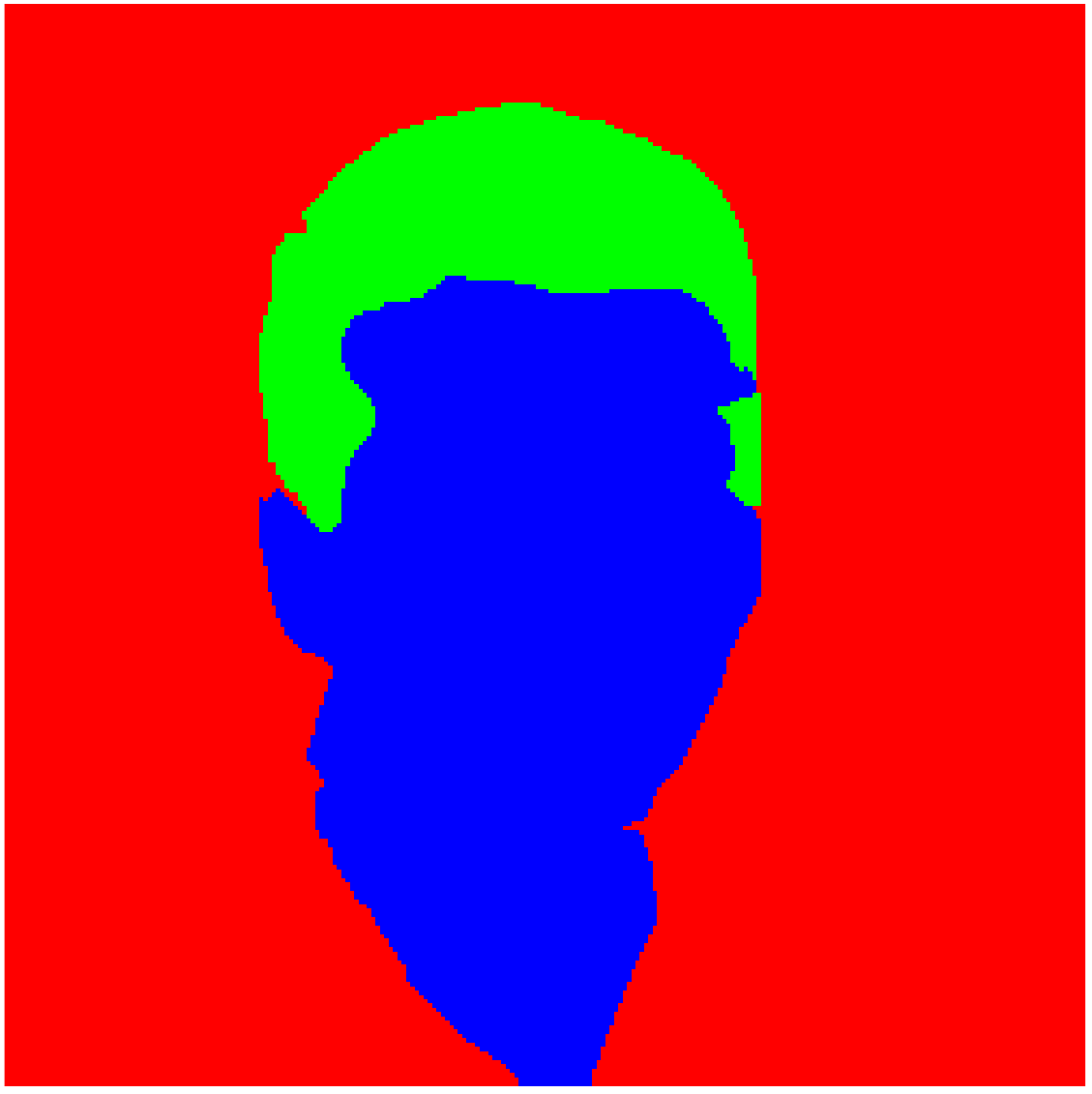}
\includegraphics[width=0.06\textwidth]{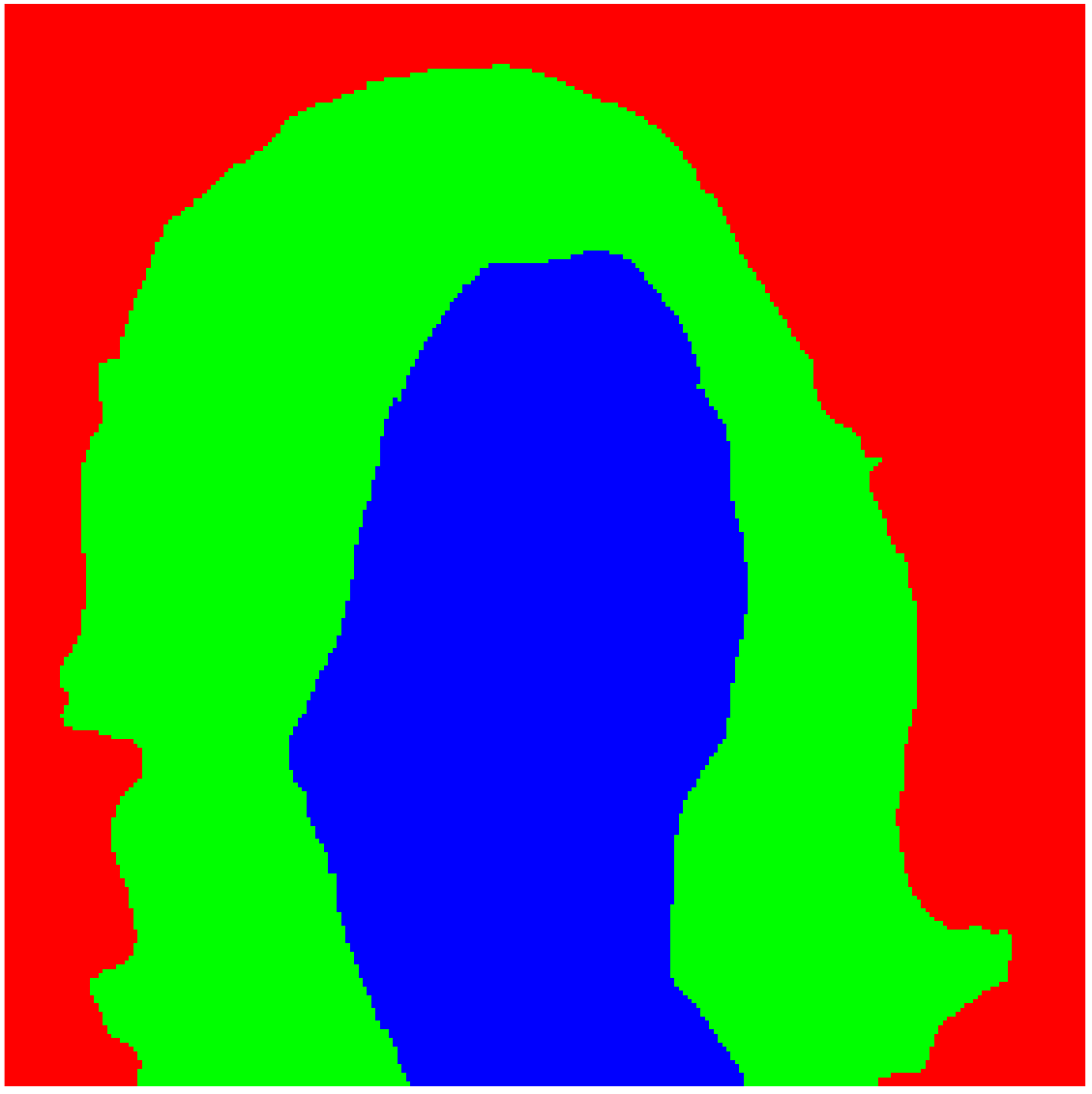}
\includegraphics[width=0.06\textwidth]{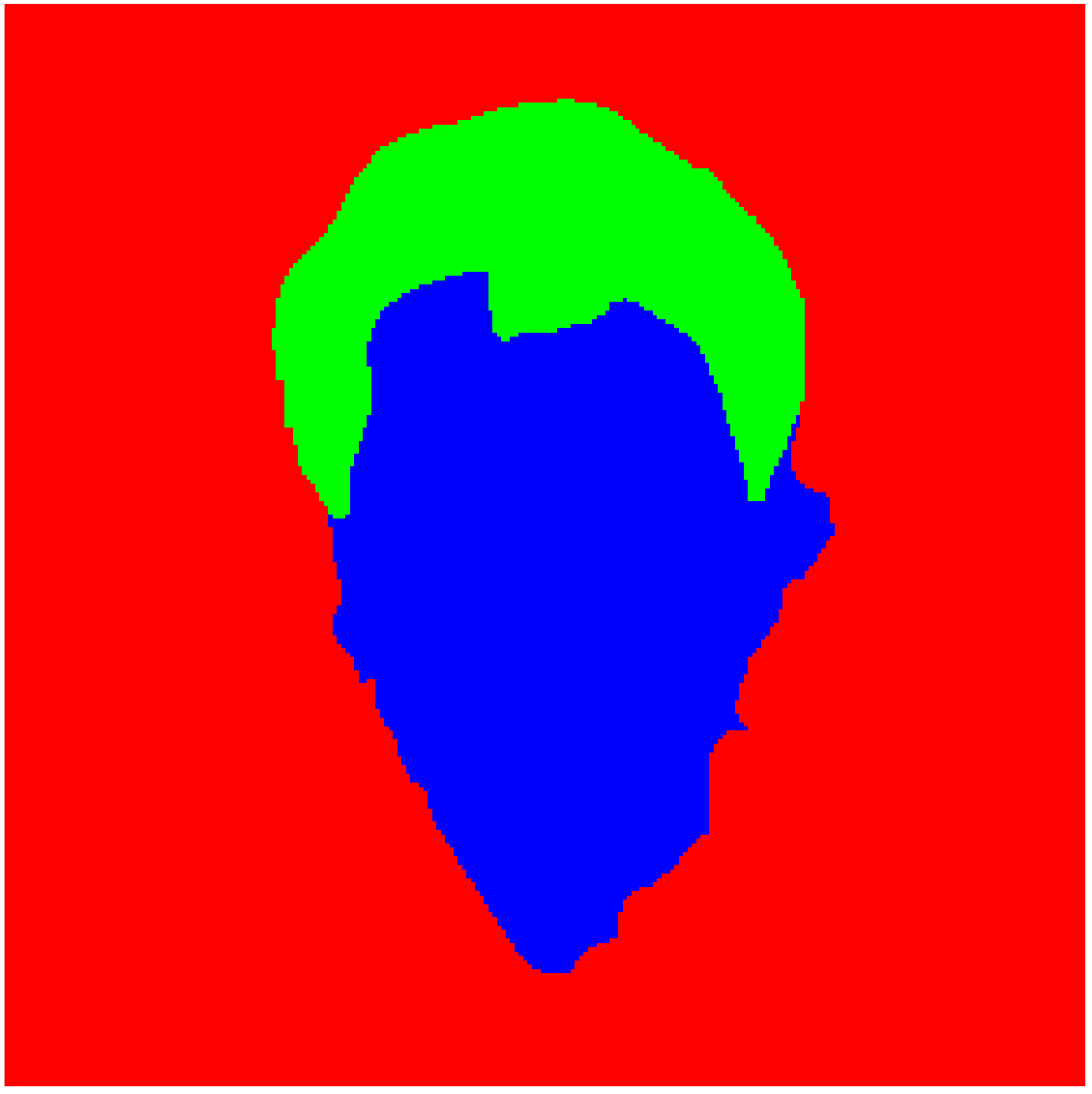}
\includegraphics[width=0.06\textwidth]{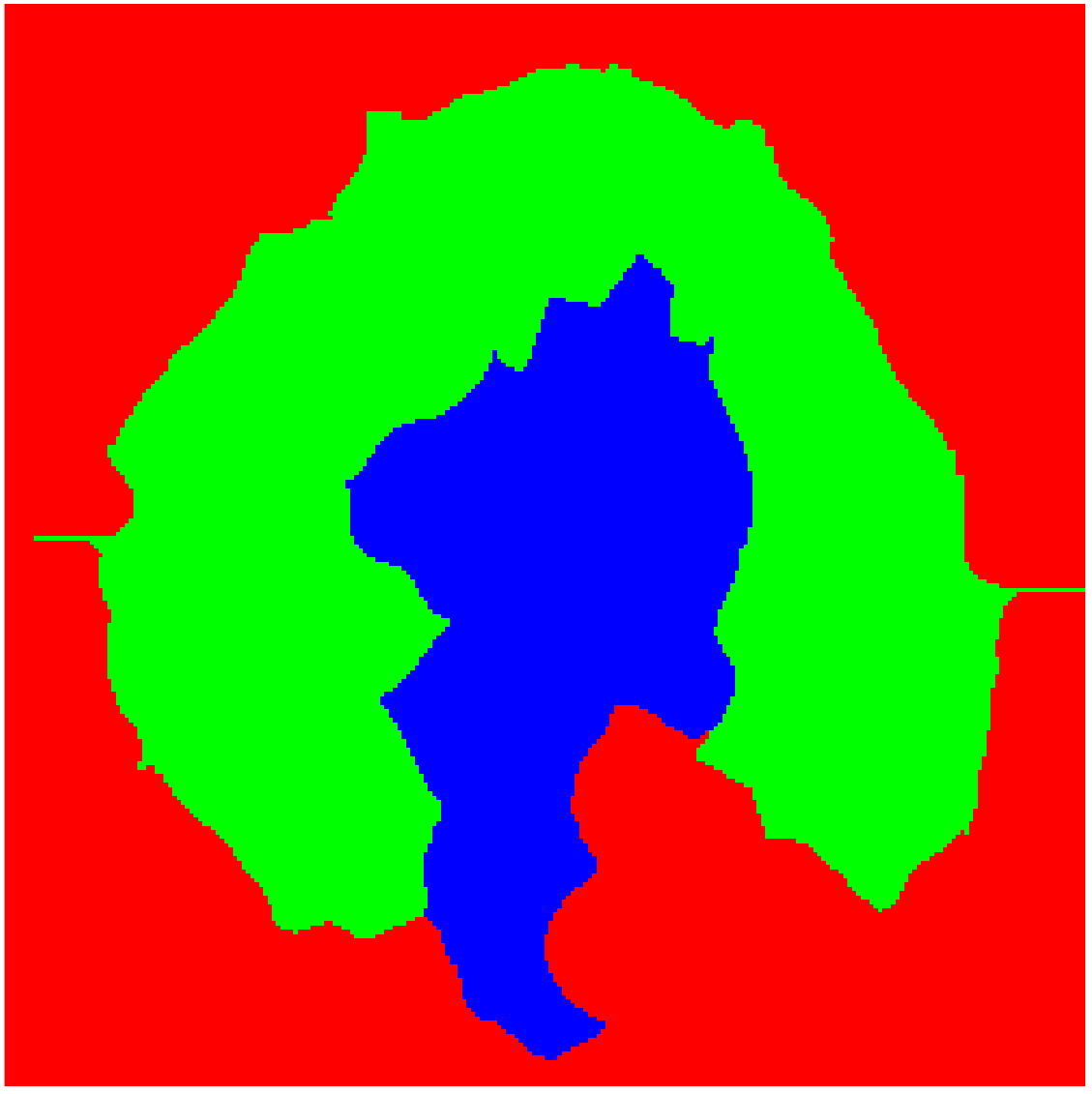}
\includegraphics[width=0.06\textwidth]{Janet_Napolitano_0002_gt}
\includegraphics[width=0.06\textwidth]{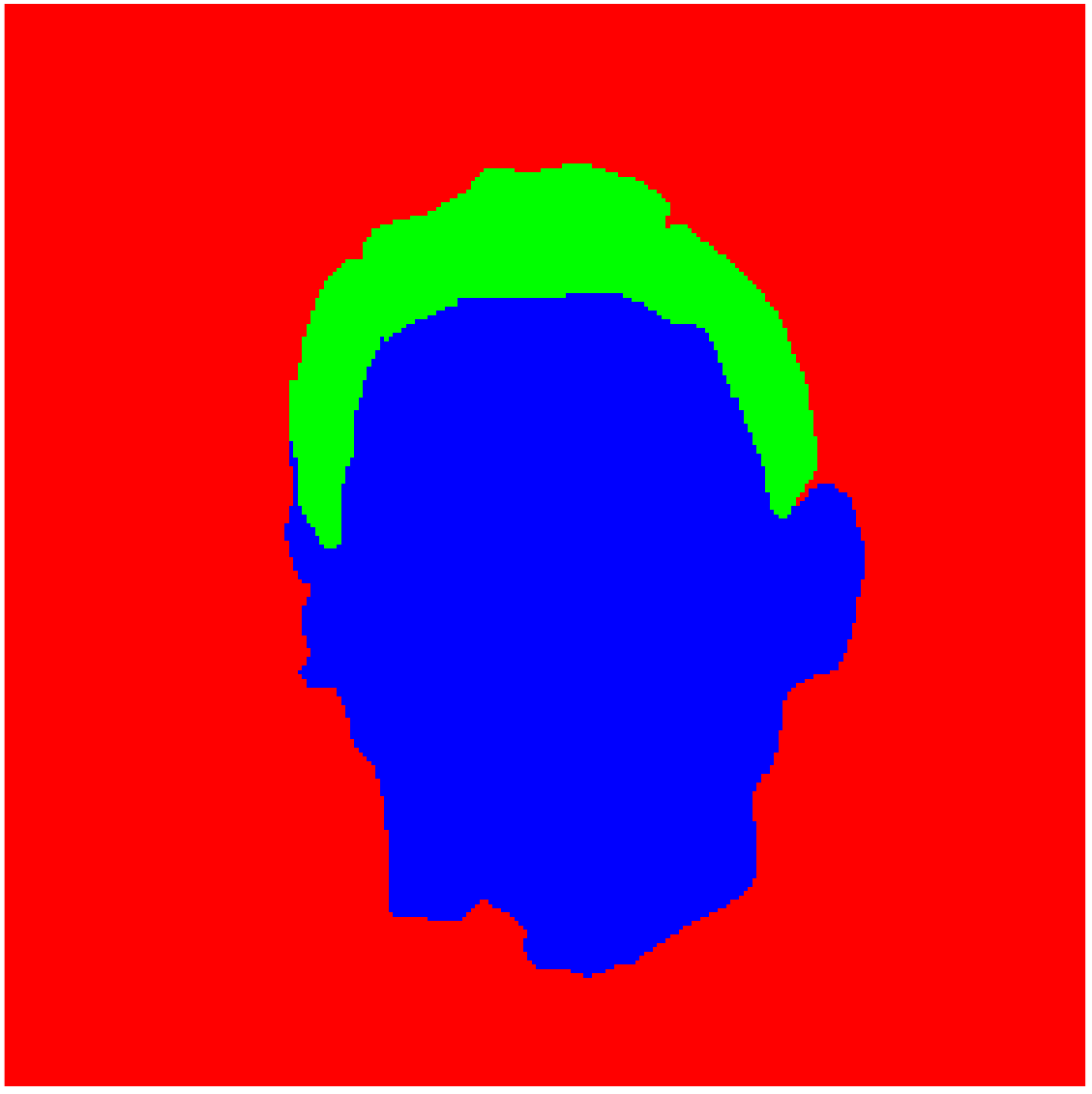}
\includegraphics[width=0.06\textwidth]{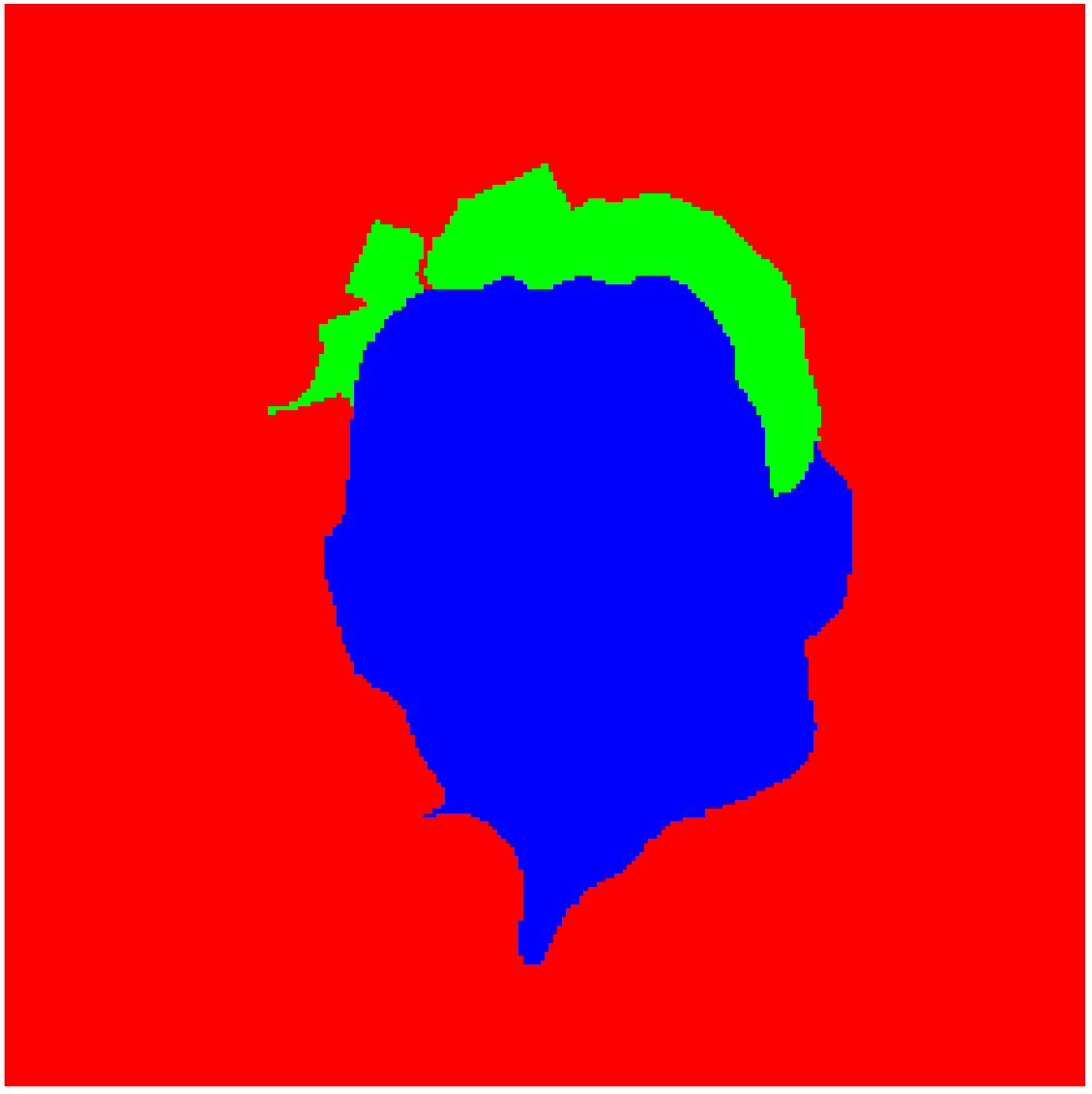}
\includegraphics[width=0.06\textwidth]{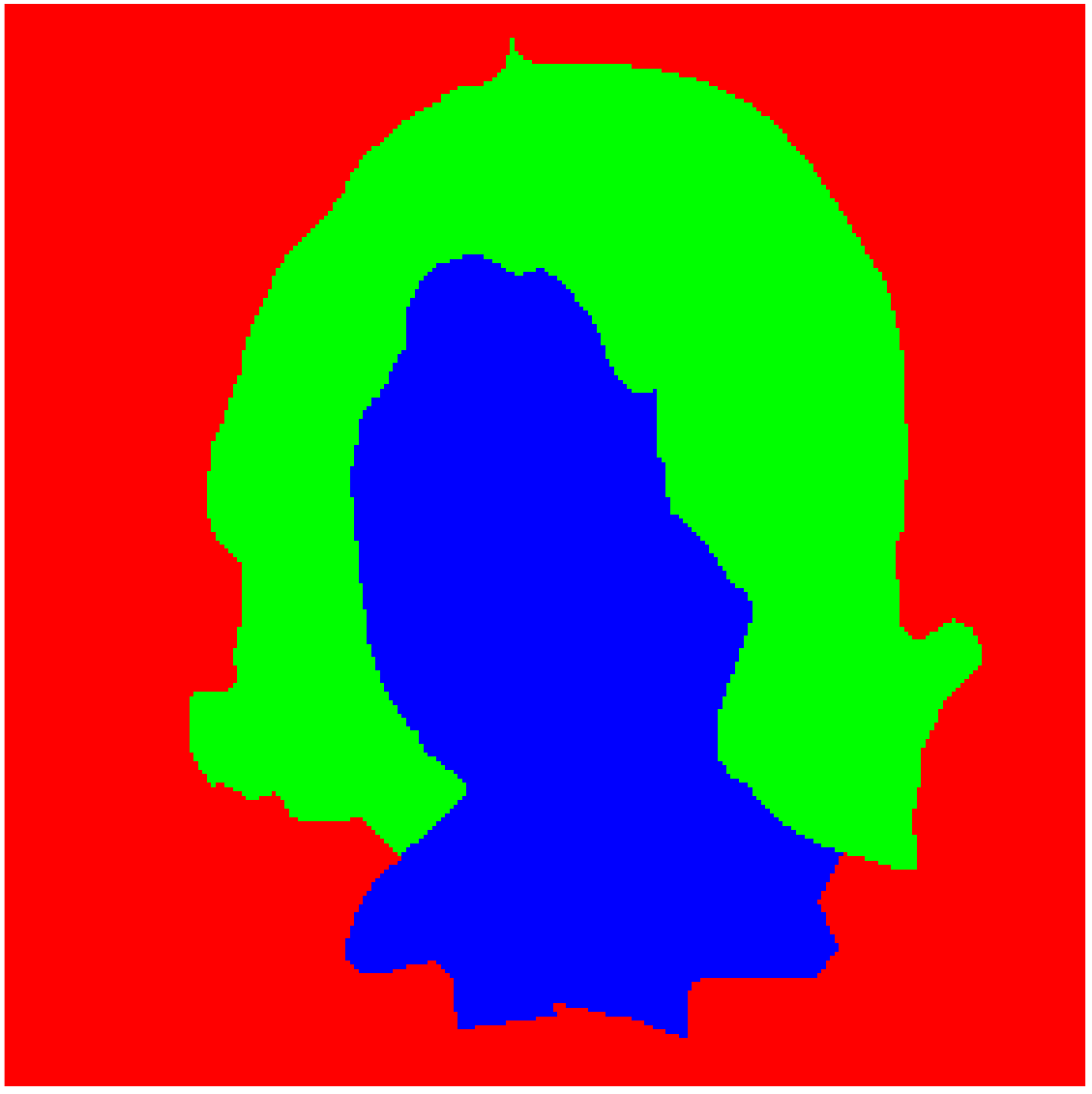}
\includegraphics[width=0.06\textwidth]{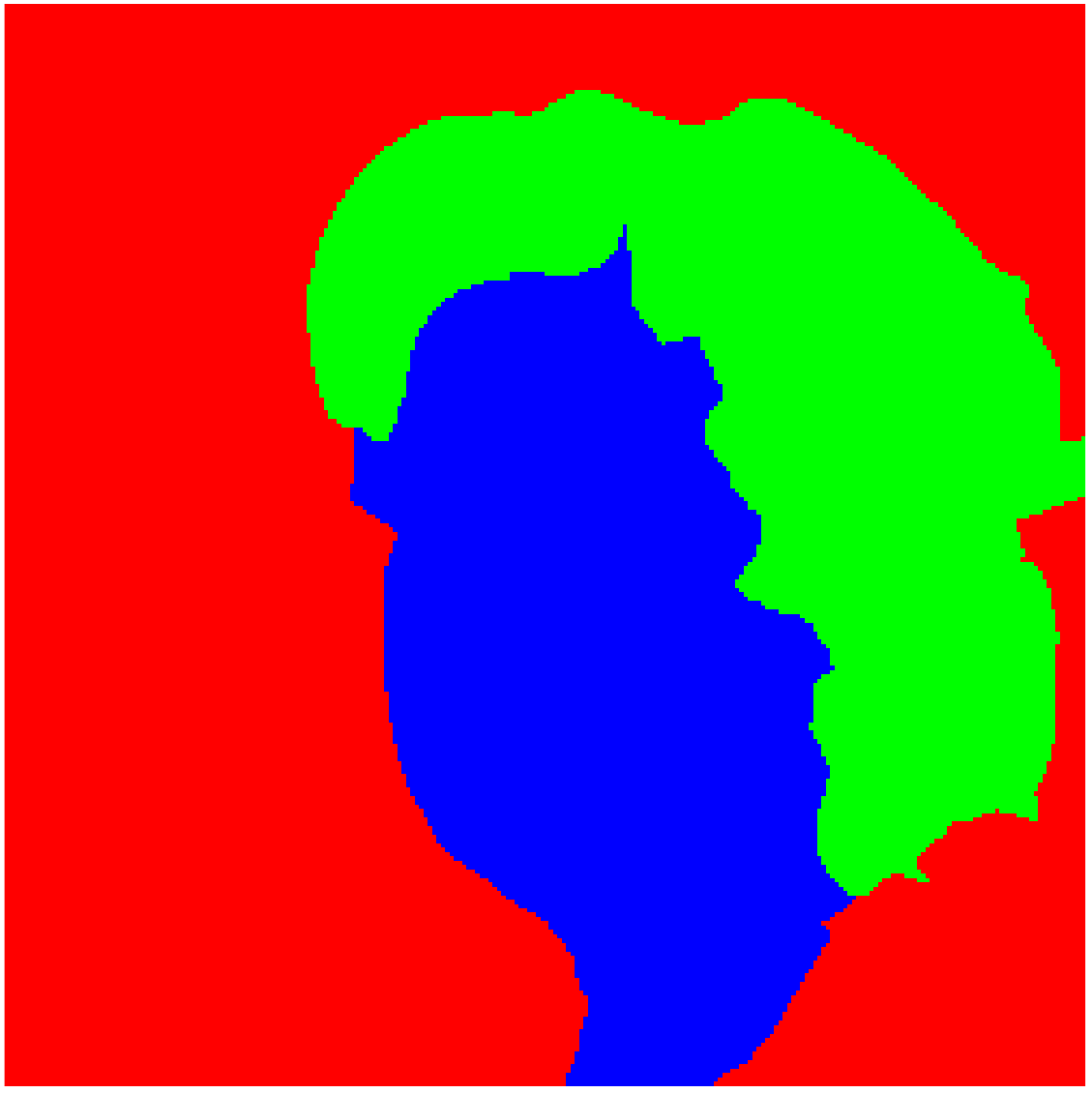}
\caption{Face parsing results on LFW. From top to bottom: a) Input image, b) Masks from raw CNN scores, c) CNN+CRF, d) Groundtruth. CRF sharpens boundaries, especially in the case of long hair. In the 6th image we see a failure case: our system failed to distinguish the man's very short hair from the similar-color head skin.}

\label{fig:faces}
\end{figure*}

\subsubsection*{Caltech-UCSD Birds-200-2011}
\label{sec:cub200}

CUB-200-2011~\cite{WahCUB_200_2011} is a dataset for fine-grained recognition that contains over 11000 images of various types of birds. CUB-200-2011 does not contain segmentation masks for parts, however Zhang \etal~\cite{zhang2014part}  provide bounding boxes for the whole bird, as well as for its head and body. In that work, the authors describe a system for detecting object parts under two different settings: 1) When the object bounding box is given, and 2) when the location of the object is unknown.

We can assess the performance of our system by converting segmentation masks of bird parts to their (unique) corresponding bounding boxes. 
We train our system on the trainval bird part annotations in the Pascal Parts dataset and use the CUB-200-2011 test set (5793 images) for evaluation. We consider five parts: head, body, wings and legs and compare with~\cite{zhang2014part}. We only focus on the case where the bounding box is considered to be known. Given the bird's bounding box, we compute the segmentation masks of four parts using our network: head, body, wings and legs. We then use the label masks for head and body to construct bounding boxes. Since our final goal is to convert a segmentation mask to a bounding box, sharp boundaries are not mandatory and we only utilize the coarse CNN scores. 
We measure accuracy in terms of PCP (Percentage of Correctly Localized Parts). Our simple approach proves effective and outperforms~\cite{zhang2014part} by ~2\% in detecting bounding boxes for the bird's body, raising performance from $79.82\%$ to $81.79\%$ . Part R-CNN is ahead by 4\% in localizing birds' heads ($68.19\%$ \vs $64.41\%$) but this is a system that was specifically trained for this task. Furthermore, R-CNN capitalizes on the large number of region proposals (typically more than 1000) returned by the Selective Search algorithm~\cite{uijlings2013selective}. These bottom-up proposals can potentially be a bottleneck when trying to localize small parts, or when higher IOU is required~\cite{zhang2014part}. In contrast, our approach yields a single, high-quality segmentation proposal for each object part, removing the need to score ``partness'' of hundreds or thousands of  individual regions.

\subsubsection*{Pascal Parts dataset}
\label{sec:pascal-parts}
In our last experiment, we evaluate our system on the PASCAL Parts dataset~\cite{chen2014detect}. This dataset includes high quality part annotations for the 20 PASCAL object classes (train and val sets), but was released fairly recently, so there are not many works reporting part segmentation performance. The only work that we know of is by Lu \etal~\cite{lu2014parsing} on car parsing, but the authors do not provide quantitative results in the form of some accuracy percentage, making comparison challenging. 

Nevertheless, we report our own results for~\emph{horse, cow} and \emph{car}, which could serve as a first baseline. 
For each class we train a separate DCNN on the train set annotations (using horizontal flipping to augment the training dataset), and test on the validation set. 
Our quantitative results are compiled in \reftab{tab:pascal}, while in \reffig{fig:pedestrian} we show qualitative results.

\begin{table}
\label{tab:pascal}
\subfloat[IOU scores on PASCAL-Parts horse class]{
\resizebox{0.48\columnwidth}{!}{
	\begin{tabular}{|c|c|c|c|c|c|c|c|}
		\hline
		Method 			& head   		& neck   		& torso 		& legs 			& tail		& BG 		& Average  		\\\hline
		CNN				& 55.0   		& \bf{34.2}     & 52.4    		& \bf{46.8} 	& 37.2		& 76.0   	& \bf{50.3}			\\\hline
		CNN+CRF			& \bf{55.4}    	& 31.9   		& \bf{53.6} 	& 43.4			& \bf{37.7} & \bf{77.9}	& 50.0 			\\\hline
	\end{tabular}
	}}
\subfloat[IOU scores on PASCAL-Parts cow class.]
{\resizebox{0.5\columnwidth}{!}{
	\begin{tabular}{|c|c|c|c|c|c|c|}
		\hline
		Method 		  	& head   		& torso 		& legs 			& tail		& BG 		& Average  		\\\hline
		CNN				& 57.6   		& 62.7     		& \bf{38.5}    	& \bf{11.8}	& 69.7   	& 48.03			\\\hline
		CNN+CRF			& \bf{60.0}    	& \bf{64.8}   	& 34.8		 	& 9.9 		& \bf{72.4}	& \bf{48.38} 	\\\hline
	\end{tabular}
	}}\\

\subfloat[IOU scores on PASCAL-Parts car class.]{
	\resizebox{0.48\columnwidth}{!}{
	\begin{tabular}{|c|c|c|c|c|c|c|c|}
		\hline
		Method 			& body   		& plates   		& lights 		& wheels 		& windows	& BG 		& Average  		\\\hline
		CNN				& 73.4   		& \bf{41.7}    	& \bf{42.2}    	& \bf{66.3} 	& 61.0		& 67.4   	& \bf{58.7}		\\\hline
		CNN+CRF			& \bf{75.4}    	& 35.8   		& 36.1 			& 64.3			& \bf{61.8}	& \bf{68.7}	& 57.0 			\\\hline
	\end{tabular}
}}
\subfloat[PASCAL-cow]{
\resizebox{.26\columnwidth}{!}{
	\begin{tabular}{|c|c|c|}
		\hline
		Method 			& Val set   & Val subset 	\\\hline
		CNN				& 77.3   	& 83.7  	\\\hline
		CNN+RBM			& 77.7   	& 84.4   	 	\\\hline
		CNN+CRF			& 79.1    	& 85.1   	 	\\\hline
		CNN+RBM+CRF		& 79.2    	& 84.7   	 	\\\hline
	\end{tabular}
	}\label{tab:pascal-sbm1}}
\subfloat[PASCAL-horse]
{\resizebox{.26\columnwidth}{!}{
	\begin{tabular}{|c|c|c|}
		\hline
		Method 		  	& Val set   	& Val subset 	\\\hline
		CNN				& 76.6   		& 86.4     		\\\hline
		CNN+RBM			& 76.3    		& 86.9   		\\\hline
		CNN+CRF			& 77.6    		& 88.1   		\\\hline
		CNN+RBM+CRF		& 76.7    		& 87.6   		\\\hline
	\end{tabular}
	}\label{tab:pascal-sbm2}}\\

	\caption{IOU scores on PASCAL-Parts.}\label{tab:pascal}
\end{table}


In Tables \ref{tab:lfw}, \ref{tab:pascal-sbm1},\ref{tab:pascal-sbm2} we report on the relative performance of the CNN-based system compared to the CNN-RBM combination, as well as the results we obtain when combined with the CRF system.  For the ``cow'' and ``horse'' categories we also consider a separate subset of images containing poses of only moderate variation, to focus on cases that should be tractable for an RBM-based shape prior.

We observe that while the RBM typically yields a moderate improvement in performance over the CNN, this does not necessarily always carry over  to the combination of these results with the CRF post-processing module. This suggests that also the CRF stage should be trained jointly, potentially along the lines of \cite{kae2013augmenting}, which we leave for future work.

\section{Multi-scale Semantic segmentation in the wild}\label{sec:multi}
In all the experiments described so far we assume that we have a tight bounding box around the object we want to segment. However,  knowing the precise location of an object in an image is a challenging problem in its own right. In this section we investigate possible ways to relax this constraint, by applying our system on the full input image and segmenting object parts ``in the wild''. There are two ways to attack this task. An obvious approach would be to simply run the DCNN on the input image; since the network is fully convolutional, the input can be an image of arbitrary height and width. A complication that arises is that our system has been trained using examples resized at a canonical scale $(321\times 321)$, whereas an image might contain objects at various scales. As a consequence, using a single-scale model will probably fail to capture fine part details of objects deviating from its nominal scale. Another approach is  to utilize an object detector to obtain an estimate of the object's bounding box in the image, resize the cropped box in the canonical dimensions, and segment the object parts as in the previous sections. The obvious drawback is that potential errors in detection -- recovering a misaligned bounding box or missing an object altogether -- hinder the final segmentation result.

We explore a simple way of tackling these issues, by coupling our system with a recent, \stateoftheart{} object detector~\cite{ren2015faster}. We focus on the \emph{person} class from the PASCAL-Parts dataset, training our part segmentation network on the \emph{train} set and using \emph{val} to test our performance. Our approach consists of the following steps:
We start by applying the CNN over the full image domain,  at three different scales (original dimensions + upsampling by a factor of  $1.5, 2$) (we did not use finer resolutions, due to GPU RAM constraints).
We then use~\cite{ren2015faster} to obtain a set of region proposals, along with their respective class scores. Unlike~\cite{hariharan2014hypercolumns}, we keep \emph{all} proposals, omitting any NMS or thresholding steps, and use their bounding boxes as an indicator for scale selection. We associate each bounding box with its ``optimal scale'', namely the scale at which the bounding box dimensions are closer to the nominal dimensions of the network input: $s_o = \argmin{|h_b - h_N| + |w_b - w_N|}$, $h_b, w_b$  being the box's height and width, and $h_N = w_N = 321$ being the nominal scale at which the network was trained. CNN scores at an image location $\vec{x}$ are selected from the optimal scale of the box that contains it; if $\vec{x}$ is contained in multiple boxes, we use the scale and scores supported by  the box  with the highest detector score. 

This approach allows us to synthesize a map of part scores, combining patches from finer resolutions when the object is small, and coarser resolutions when the object is large. 
At test time, we extract all \groundtruth{} bounding boxes for the \emph{person} class in PASCAL-Parts \emph{val} set and calculate pixel accuracy within the boxes.
As a baseline for comparison we use the naive evaluation of the CNN, applied on a single scale (original image dimensions). This simple approach boosts performance from $73.9\%$ to $74.7\%$ \emph{without} training the network with multi-scale data, even though end-to-end training could yield further improvements.
\begin{figure}[ht]
\centering
\subfloat[Input image]{\includegraphics[width=0.245\textwidth]{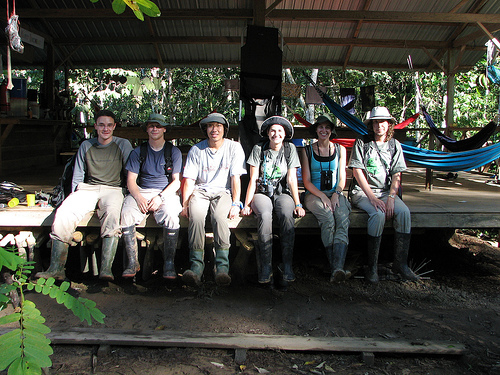}}
\subfloat[Single resolution]{\includegraphics[width=0.245\textwidth]{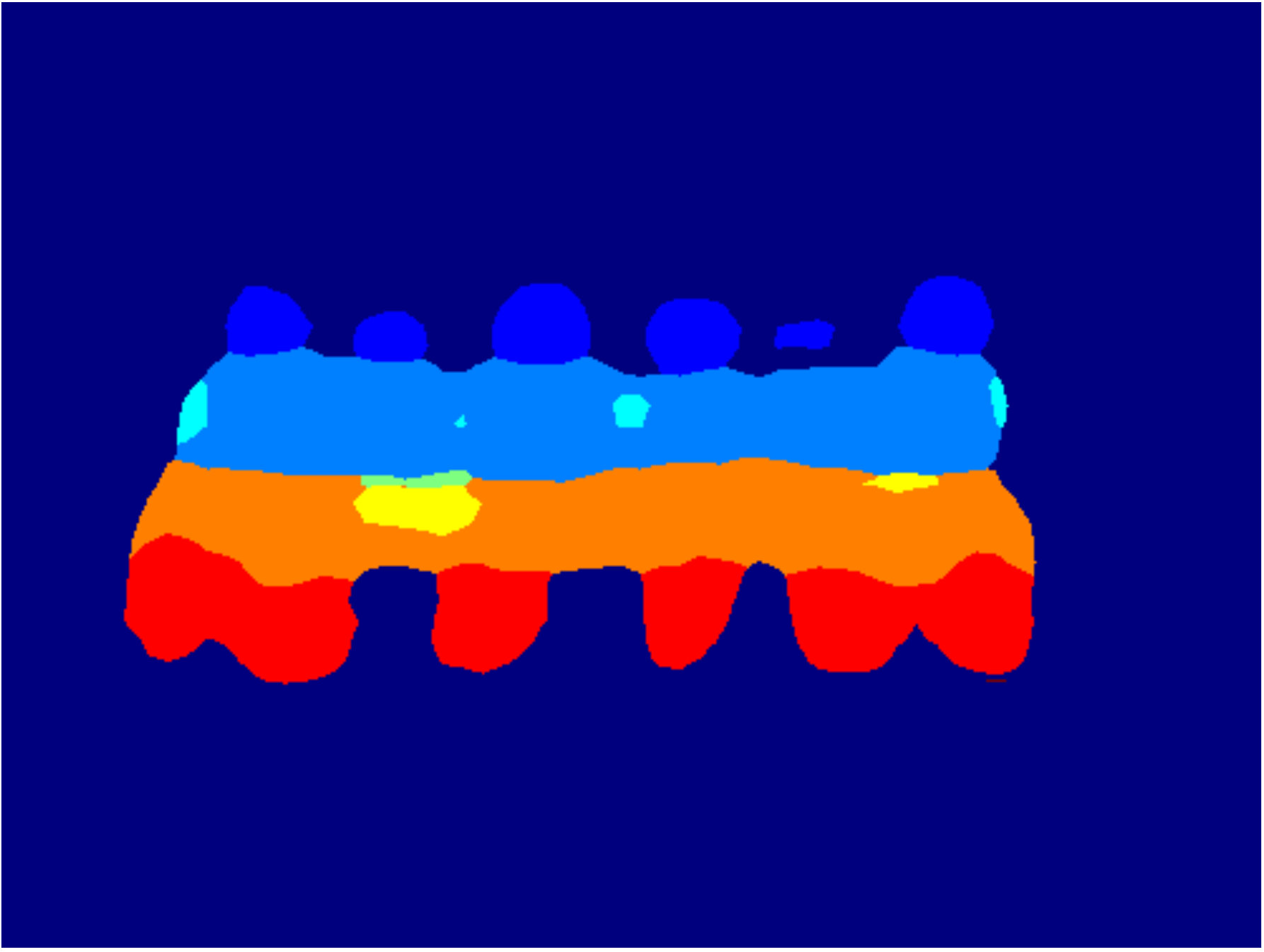}}
\subfloat[Multi-scale scheme]{\includegraphics[width=0.245\textwidth]{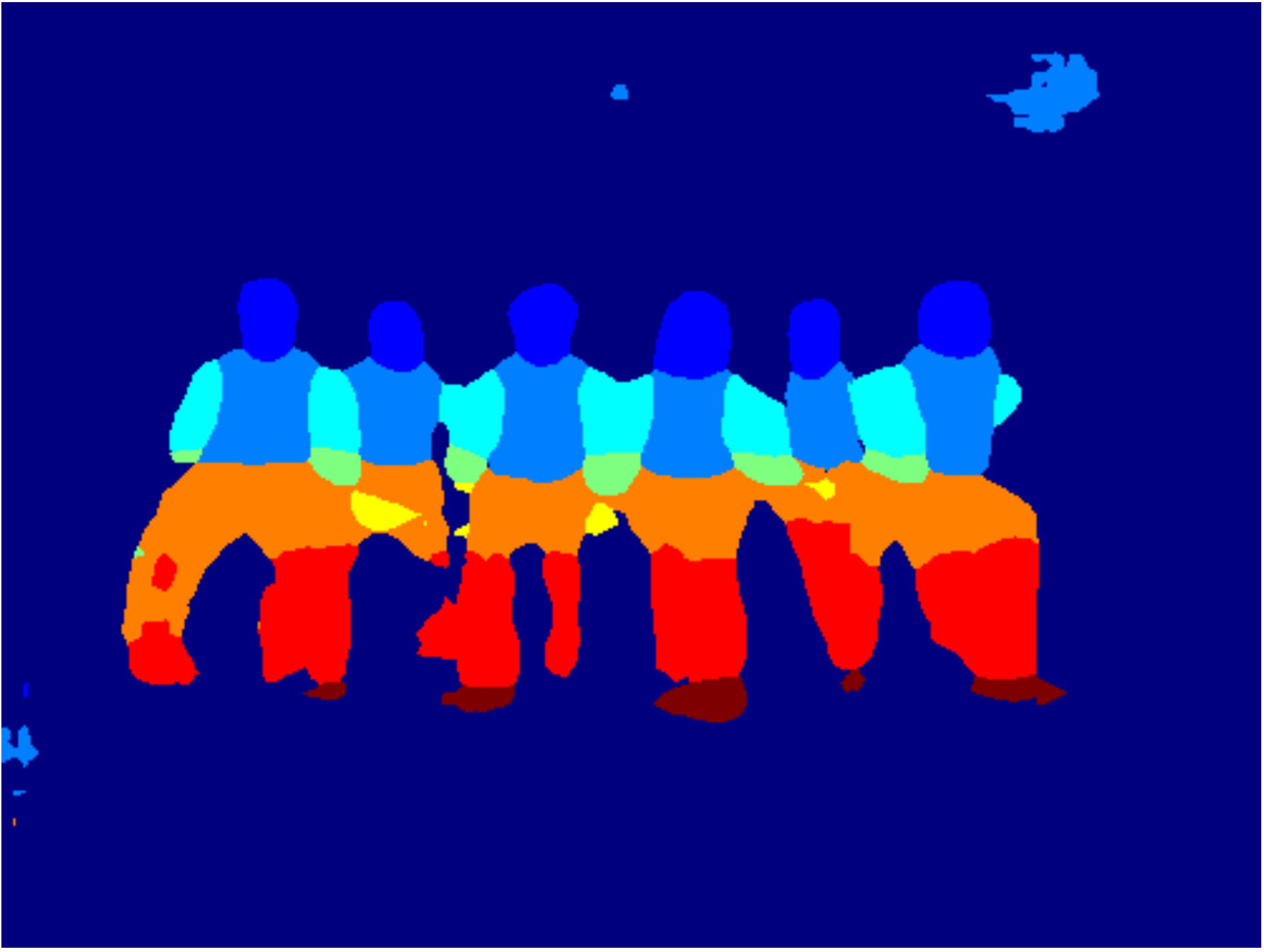}}
\subfloat[Ground truth]{\includegraphics[width=0.245\textwidth]{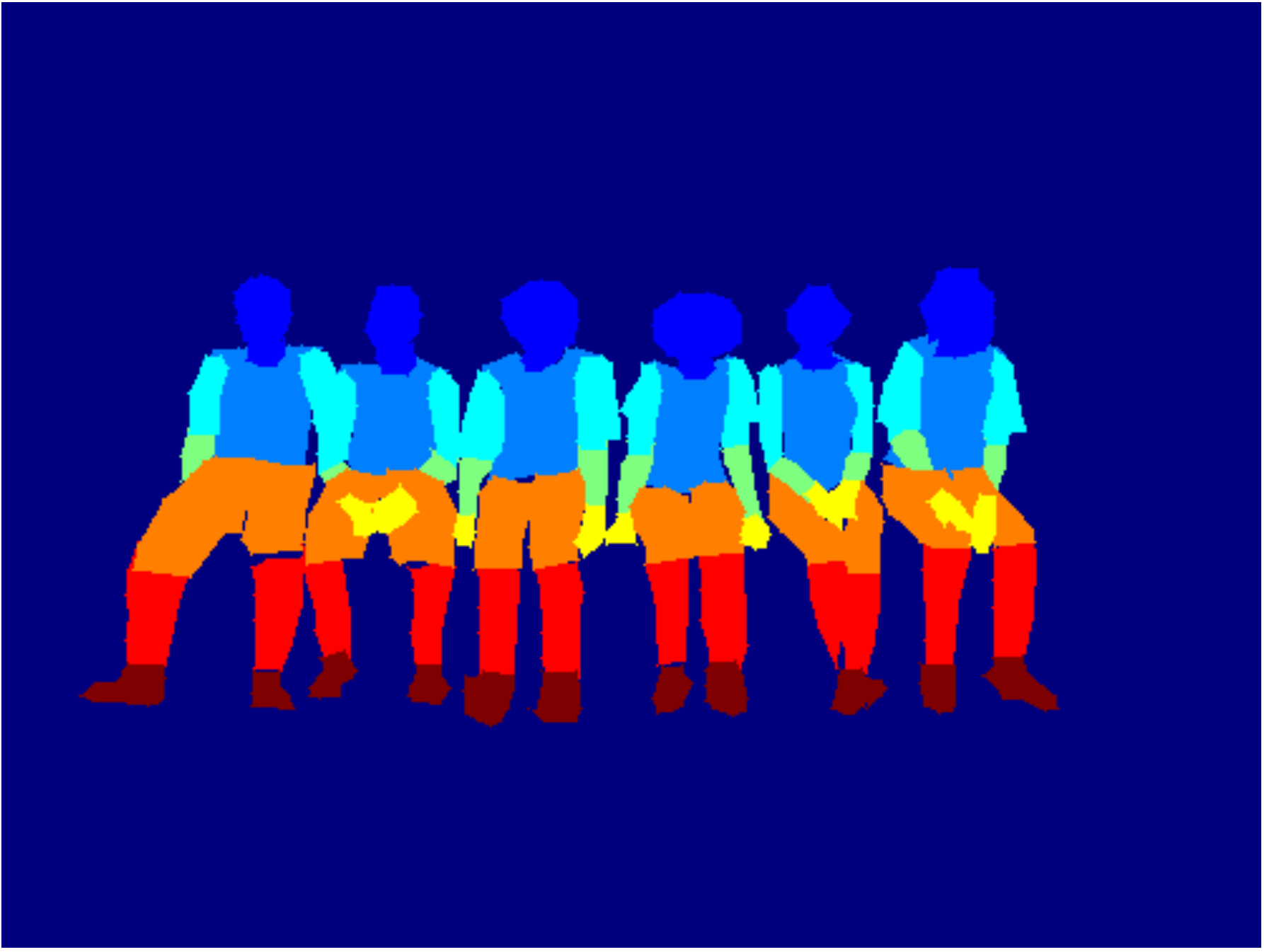}}
\caption{Combining CNN features from multiple scales, we can recover segmentations for finer object parts.}
\label{fig:multi_scale}
\end{figure}

\section{Discussion}\label{sec:discussion}
In this work we have demonstrated that a simple and generic system for semantic segmentation relying on Deep CNNs and Dense CRFs can provide state-of-the-art results in the task of semantic part segmentation.  We have also explored methods of integrating high-level information through a joint discriminative training of the network with a statistical, category-specific shape prior, showing that these can  act in a complementary manner to the bottom-up information provided by DCNNs. 
We also proposed a simple, yet effective multi-scale scheme for segmentation ``in the wild'', guided by a fast object detector, that is used both to propose possible object boxes, and select the appropriate scale for segmentation in a pyramid of CNN scores.
 
In future work we aim at exploring the joint training of DCNNs with high-level shape priors in an end-to-end manner, as well as to further explore the practical applications of semantic part segmentation in  detection and fine-grained recognition. 

\bibliography{iclr2016_conference}
\bibliographystyle{iclr2016_conference}

\end{document}